\documentclass[final]{cvpr}

\usepackage{times}
\usepackage{graphicx}
\usepackage{amsmath, amsthm, amssymb, amsfonts}

\usepackage[utf8]{inputenc} %
\usepackage[T1]{fontenc}    %
\usepackage{booktabs}       %
\usepackage{nicefrac}       %
\usepackage{microtype}      %
\usepackage{float}
\usepackage{multirow}
\usepackage{adjustbox}
\usepackage{subfigure}
\usepackage{xcolor}

\usepackage{mathtools}
\usepackage{physics}

\usepackage[english]{babel}
\usepackage{bm}

\usepackage[pagebackref=true,breaklinks=true,colorlinks,bookmarks=false]{hyperref}

\def\confYear{CVPR 2021}

\newcommand{\titlepre}{-0.25em}
\newcommand{\titlepost}{-0.25em}
\newtheorem{theorem}{Theorem}
\newtheorem{proposition}[theorem]{Proposition}
\newtheorem{lemma}[theorem]{Lemma}
\newtheorem{corollary}[theorem]{Corollary}

\newtheorem{remark}[theorem]{Remark}

\theoremstyle{definition}
\newtheorem{definition}[theorem]{Definition}

\newcommand{\RR}{\mathbb{R}}

\newcommand{\SpS}{\mathbb{S}}

\newcommand{\Radon}[1]{\mathcal{R} \{ #1 \}}
\newcommand{\RadonT}[1]{\mathcal{R}^* \{ #1 \}}

\begin{document}

\title{Neural Splines: Fitting 3D Surfaces\\ with Infinitely-Wide Neural Networks}

\author{
    Francis Williams\textsuperscript{1}
    \and Matthew Trager\textsuperscript{1, 2}\thanks{Work done prior to joining Amazon}
    \and Joan Bruna\textsuperscript{1}
    \and Denis Zorin\textsuperscript{1}\vspace{0.25em}
    \and \textsuperscript{1}New York University, \textsuperscript{2}Amazon
    \and \texttt{\small francis.williams@nyu.edu, mtrager@cims.nyu.edu, bruna@cims.nyu.edu, dzorin@cs.nyu.edu}
}

\maketitle

\begin{abstract}
We present Neural Splines, a technique for 3D surface reconstruction that is based on random feature kernels arising from infinitely-wide shallow ReLU networks. Our method achieves state-of-the-art results, outperforming recent neural network-based techniques and widely used Poisson Surface Reconstruction (which, as we demonstrate, can also be viewed as a type of kernel method). Because our approach is based on a simple kernel formulation, it is easy to analyze and can be accelerated by general techniques designed for kernel-based learning. We provide explicit analytical expressions for our kernel and argue that our formulation can be seen as a generalization of cubic spline interpolation to higher dimensions. In particular, the RKHS norm associated with Neural Splines biases toward smooth interpolants.

\end{abstract}
\vspace{\titlepost}
\section{Introduction}
\vspace{\titlepre}

\begin{figure}
    \begin{center}
    \minipage[b]{0.2385\textwidth}
    \centering
    \includegraphics[width=0.99\textwidth]{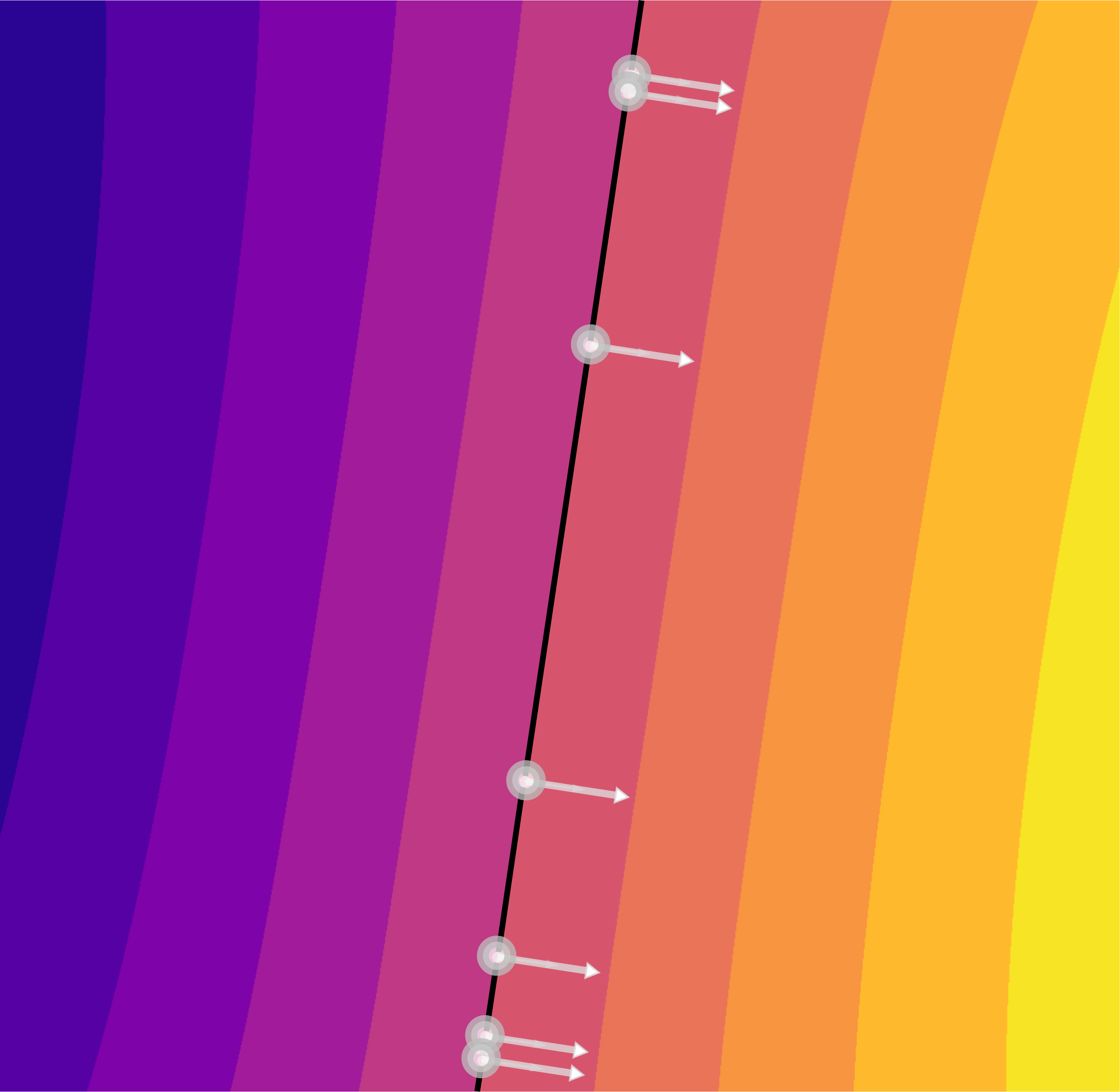}
    \endminipage
    \minipage[b]{0.2385\textwidth}
    \centering
    \includegraphics[width=0.99\textwidth]{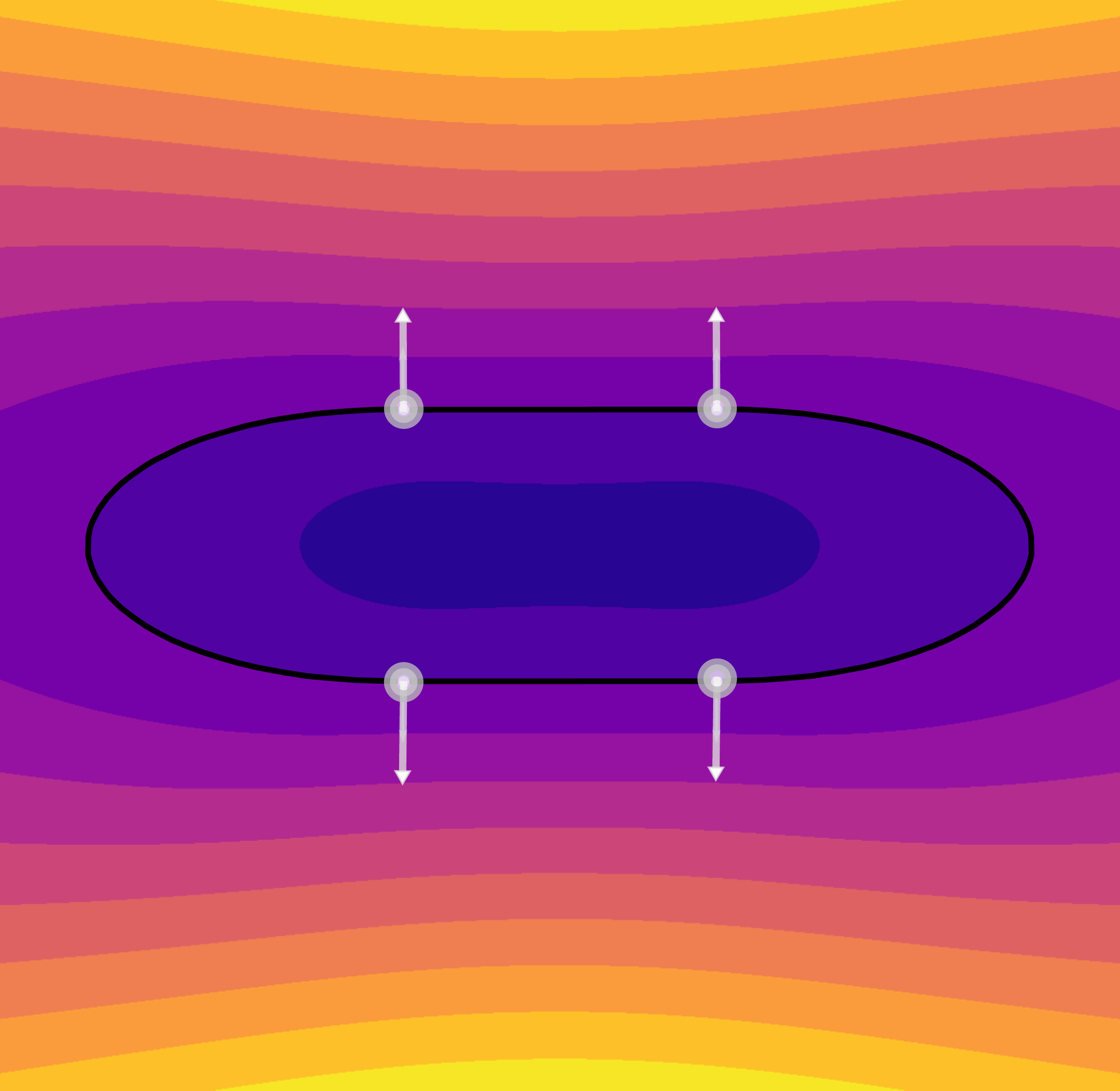}
    \endminipage\hfill
    \minipage[b]{0.2385\textwidth}
    \centering
    \includegraphics[width=0.99\textwidth]{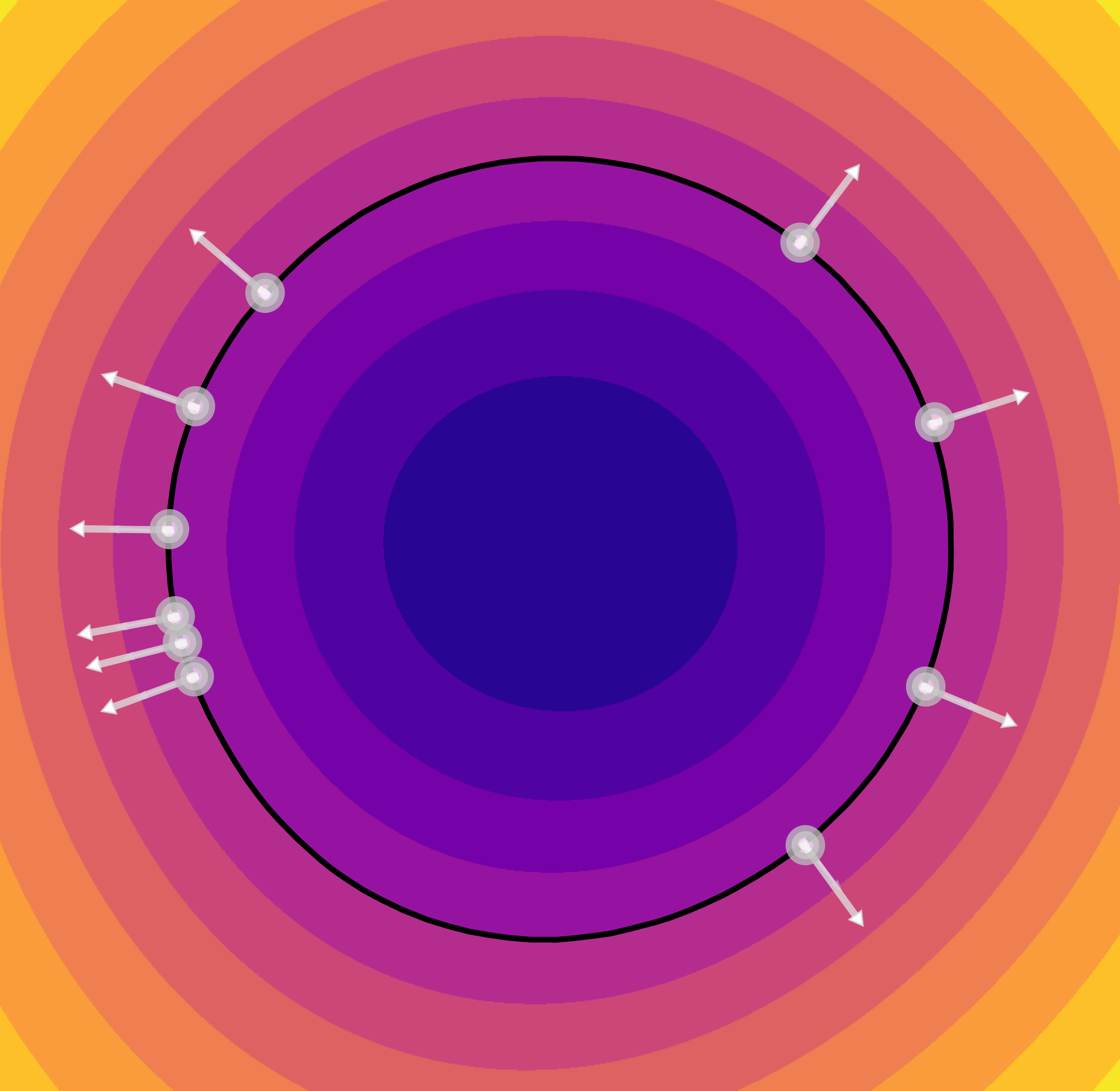}
    \endminipage
    \minipage[b]{0.2385\textwidth}
    \centering
    \includegraphics[width=0.99\textwidth]{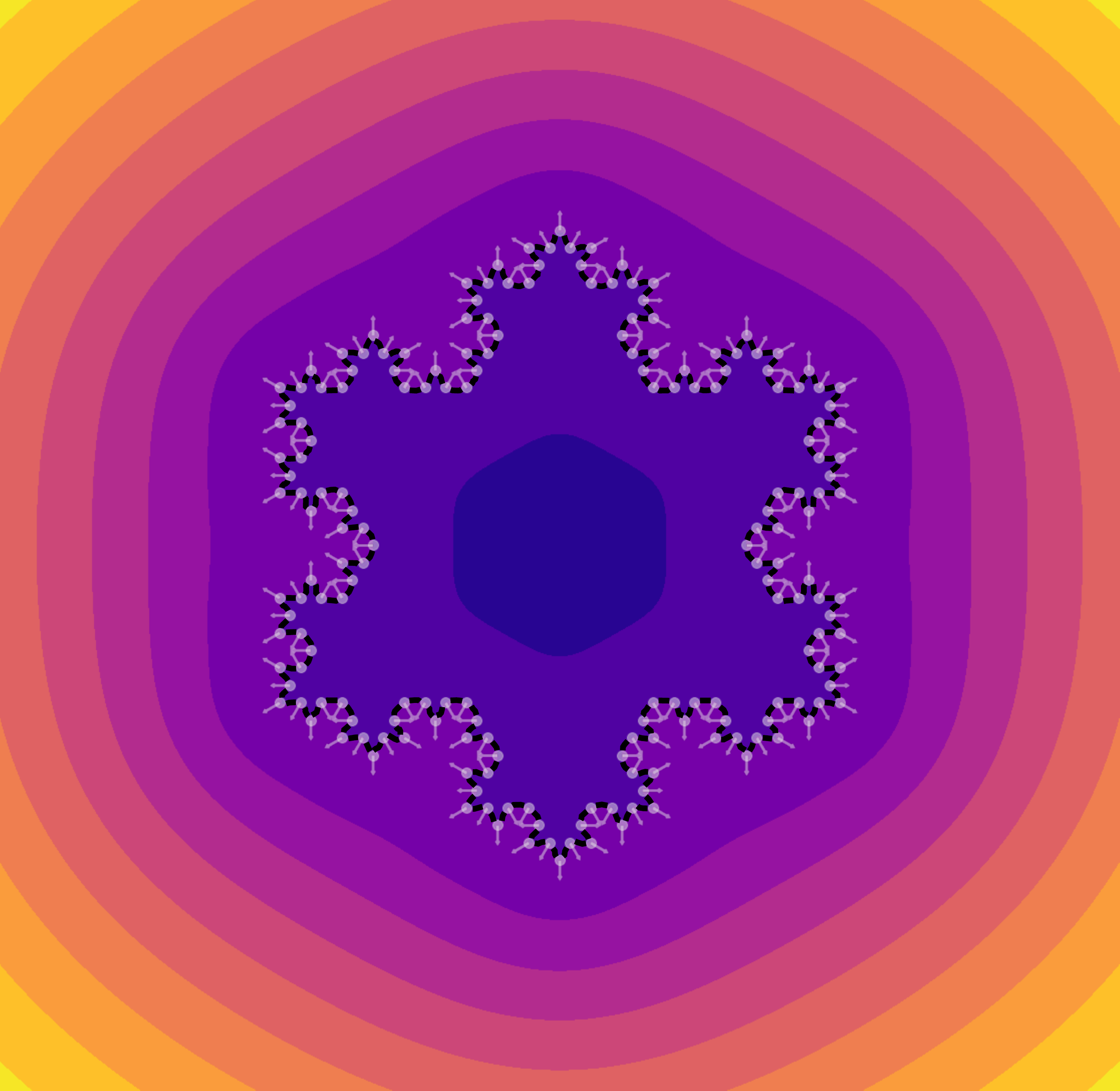}
    \endminipage\hfill
    \end{center}
    \caption{Neural Splines use points (the white dots) and normals (the white arrows) as input and estimate a scalar function whose zero level set (the black lines) corresponds to the reconstructed surface and whose gradient agrees with the normals.}
    \label{fig:2d_examples}
\end{figure}

Estimating a 3D surface from a scattered point cloud is a classical and important problem in computer vision and computer graphics. In this task, the input is a set of 3D points sampled from an unknown surface and, possibly, normals of the surface at those points. The goal is to estimate a representation of the complete surface from which the input samples were obtained, for example, a polygonal mesh or an implicit function. This problem is challenging in practice: It is inherently ill-posed, since an infinite number of surfaces can interpolate the data. Furthermore, the input 3D points are often incomplete and noisy, as they are acquired from range sensors such as LIDAR, structured light, and laser scans. 
Ideally, the recovered surface should not interpolate noise but preserve key features and surface details. %

Many early surface reconstruction techniques consider a kernel formulation of the surface reconstruction problem, using translation-invariant kernels such as 
biharmonic RBFs \cite{carr2001reconstruction}, Gaussian kernels \cite{NIPS2004_2724}, or compactly supported RBFs  \cite{ohtake2006sparse}. 
Currently, the most widely used method for surface reconstruction is \emph{Screened Poisson Surface Reconstruction}~\cite{kazhdan2013screened}, which solves a variant of the Poisson equation to find an implicit function whose zero-level set produces an approximating surface. 
We show in this paper that this method can also be viewed as a kernel method for a particular choice of kernel. 

More recently, many papers have used neural networks to represent an implicit function or a local chart in a manifold atlas  as a means of reconstructing a surface \cite{Williams_2019, gadelha2020deep, hanocka2020point2mesh, atzmon2019sal, gropp2020implicit, sitzmann2020implicit, tancik2020fourier}. These methods can be integrated into a data-driven learning pipeline or directly applied in the so called “overfitting” regime, where a massively overparameterized (\ie, more parameters than input points) neural network is fitted to a single input point cloud as a functional representation for a surface.  Empirical evidence has shown that 
these methods 
enjoy some form of ``implicit regularization'' that biases the recovered surface towards smoothness. Moreover, employing early stopping in gradient descent can prevent these approaches from interpolating noise. 

Under certain parameter initializations, infinitely-wide neural networks \emph{de facto} behave as kernel machines \cite{jacot2018neural, chizat2019lazy}, defining Reproducing Kernel Hilbert Spaces (RKHS), whose kernel is obtained by linearizing the neural network mapping around its initialization. 
While the kernel regime simplifies non-linear neural network learning into a convex program and provides a simple explanation for the successful optimization of overparameterized models, it cannot explain the good generalization properties observed for high-dimensional problems due to the inability of the RKHS to approximate non-smooth functions \cite{bach2017}.
However, the situation for low-dimensional problems such as surface reconstruction is entirely different, and RKHS can provide powerful characterizations of regularity. In this context, \cite{williams2019gradient} shows that in the univariate case the RKHS norm associated with a wide ReLU network is a weighted curvature, and leads to cubic spline interpolants. In higher dimensions, similar (albeit more complex) characterizations of the RKHS norm exist \cite{ongie2019function} (see also Proposition~\ref{prop:ongie}). In order to assess the benefits of neural networks on such low-dimensional problems, it is thus important to first understand their linearized counterparts, given by their associated kernel machines. 

In this work, we demonstrate that in fact kernels arising from shallow ReLU networks are extremely competitive for 3D surface reconstruction, achieving state-of-the-art results: outperforming classical methods as well as non-linear methods based on far more complex neural network optimization.

Kernels provide many advantages over neural networks in our context: \textit{(i)}~They are  well understood theoretically. \textit{(ii)}~Kernel regression boils down to solving a linear system, and avoids gradient descent that suffers from slow convergence. \textit{(iii)}~Kernel-based interpolants are represented using a number of parameters that is linear in the size of the input, whereas overparameterized neural networks require many more parameters than points. \textit{(iv)}~The inductive bias of kernel methods can be characterized explicitly via the RKHS norm (Section~\ref{sec:implicit_bias}). \textit{(v)}~Kernel methods lend to scalable and efficient implementations (Section~\ref{sec:fast_implementation}).
We provide explicit expressions for two kinds of infinite-width ReLU kernels, their derivatives, and their corresponding RKHS norms. We further argue that these kernels can be viewed as a multidimensional generalization of cubic spline interpolation in 1D. Moreover, we show that Poisson Surface Reconstruction can itself be viewed as a kernel method and give an expression for its RKHS norm, suggesting that kernels are a broad framework which enable the rigorous understanding both traditional and modern surface reconstruction techniques.

\vspace{-0.5em}
\vspace{\titlepre}
\subsection{Additional Related Work}
\vspace{\titlepost}

Methods for 3D surface reconstruction can mostly be divided by their choice of surface representation: These are \textit{(i)} the zero level set of a volumetric scalar function \cite{kazhdan2013screened, atzmon2019sal, gropp2020implicit, Mescheder_2019, Park_2019, chabra2020deep, sitzmann2020implicit, tancik2020fourier}, \textit{(ii)} the fixed point of a projection operator onto locally fitted patches \cite{adamson2003approximating, guennebaud2007algebraic, huang2013edge}, \textit{(iii)} a mesh connecting the input points \cite{hoppe1992surface}, \textit{(iv)} a collection of local parametric maps \cite{Williams_2019, hanocka2020point2mesh, deprelle2019learning, gadelha2020deep}, or \textit{(v)} the union of parametric shapes \cite{genova2019learning, genova2020local, deng2020cvxnet, chen2020bspnet, Williams_2020_CVPR_Workshops}. For a recent up-to-date survey of surface reconstruction techniques, we refer the reader to \cite{berger2017survey}.

The ``random feature'' kernels used this paper arise from training the top-layer weights in two-layer networks of infinite width and were described in~\cite{cho2009kernel,le2007continuous}. More recently, a lot of work has focused on a different kernel, known as the ``neural tangent kernel''~\cite{jacot2018neural,chizat2019lazy}, that linearly approximates the training of \emph{both} layers of a neural network. 
We chose to use random feature kernels since, when the input dimension is small, typical initializations in neural networks lead to mainly training the top layer weights. 
More broadly, the function spaces associated with shallow neural networks were studied in \cite{bach2017,bietti2019,ongie2019function,savarese2019infinite,williams2019gradient}.

\vspace{\titlepre}
\section{Neural Spline Formulation}
\vspace{\titlepost}
We formulate the problem of surface reconstruction as the task of finding a scalar function $f: \RR^3 \rightarrow \RR$ whose zero level set $S = \{p  \colon f(p) = 0\} \subset \RR^3$ is the estimated surface (see Figure~\ref{fig:2d_examples}). In its most general form in arbitrary dimensions, we write our problem as follows.

We assume we are given a set of $s$ input points $\mathcal X = \{x_j\}_{j=1}^s\subset \RR^d$, function values $\mathcal Y = \{y_j\}_{j=1}^s \subset \RR$, and normals $\mathcal N = \{n_j\}_{j=1}^s\subset \RR^d$. Our goal is to optimize an objective function of the form:
\begin{small}
\begin{equation}\label{eq:objective}
\begin{split}
\begin{gathered}
    \underset{\theta \in \RR^{d_\theta}}{\text{min }} L(\theta), \mbox{ with } \\
    \quad L(\theta) =  \frac{1}{2}\sum_{j=1}^{s} |f(x_j; \theta) - y_j|^2 + \|\nabla_x f(x_j; \theta) - n_j\|^2.
\end{gathered}
\end{split}
\end{equation}
\end{small}
Here $f(x;\theta)$ is a family of functions $\RR^d \rightarrow \RR$ parameterized by $\theta \in \RR^{d_\theta}$.
For surface reconstruction, we have $d = 3$ and $y_j = 0$, since the reconstructed surface is given by the zero level set.

We begin by introducing the Neural Spline family of functions through the lens of finite-width shallow neural networks in Section \ref{sec:finite_width_kernel}, and turn to the infinite-width limit formulation in Section \ref{sec:infinite_width_kernel}. While our approach is simple and our derivations straightforward, we differ slightly from a standard kernel regression setting since we employ a multi-dimensional kernel that includes the gradient of the fitted function. This formulation allows us to fit points and normals simultaneously.

\vspace{\titlepre}
\subsection{Finite-Width Kernel}\label{sec:finite_width_kernel}
\vspace{\titlepost}
We first assume that the model $f(x;\theta)$ is a two-layer ReLU neural network with $m$ neurons, but we keep
the bottom layer weights \emph{fixed} from initialization:
\begin{equation}\label{eq:finite_model}
f(x; \theta) = \frac{1}{\sqrt m}\sum_{i=1}^m c_i [a_i^T x + b_i]_+ , \,\, \theta = (c_1,\ldots,c_m) \in \RR^{m}
\end{equation}
Here we write $[z]_+ = \max(z,0)$ for the ReLU function. The bottom-layer weights $(a_{i1},\ldots,a_{id},b_i) \in \RR^{d+1}$ are {fixed} randomly according to some specified distribution.

\begin{remark} In typical initialization schemes for neural networks (\eg, Kaiming He initalization \cite{He_2015}), the weights of each layer are initialized with a variance that is inversely proportional to the number of inputs of that layer. Our choice of fixing $(a_i, b_i)$ is motivated by the fact that using a standard two-layer network, for a given target model and as $m \rightarrow \infty$, only the top layer weights tend to vary throughout training \cite{williams2019gradient}. %
\end{remark}

Our model~\eqref{eq:finite_model} is linear in the parameters $\theta$, so our objective function $L(\theta)$ is \emph{convex}. However, since we assume $m \gg s$, we expect to have an infinite set of global minimizers (an affine subspace in $\RR^{d_\theta})$. Our goal is to find the minimizer with smallest parameter norm:
\[
\theta^* = {\arg\!\min} \{\|\theta\|^2 \colon \theta \in \RR^{d_\theta} \,\, L(\theta) = 0\}.
\]
This minimizer is given explicitly by $\theta^* = W^\dagger \delta$ where $W^\dagger$ denotes the Moore-Penrose pseudo-inverse of $W$ and
\begin{small} 
\begin{equation}
W = \frac{1}{\sqrt m} \begin{bmatrix}
 [a_j^T x_i + b_j]_+ \\
{\bf 1}[a_j^T x_i + b_j]a_j\\
\end{bmatrix}_{\substack{i =1,\ldots,s\\j=1,\ldots,m}} \quad 
\delta = \begin{bmatrix} y_i\\n_i\end{bmatrix}_{i=1,\ldots,s}
\end{equation}
\end{small}
so that $W \in \RR^{(s + ds) \times m}$ and $\delta \in \RR^{s +ds}$. If $W$ has full rank, then $W^\dagger = W^T(WW^T)^{-1}$ and we can equivalently look for $z^* \in \RR^{(s + ds)}$ such that $K_{\mathcal X} z^* = W\theta^* = \delta$ where $K_{\mathcal X}=WW^T \in \RR^{(s+ds) \times (s+ds)}$. Note that $K_{\mathcal X}$ can be viewed as a Gram matrix associated with the multi-dimensional kernel $K(x,x') \in \RR^{(d+1) \times (d+1)}$ described by
\begin{small}
\begingroup
\setlength\arraycolsep{2pt}
\begin{equation}\label{eq:kernel_finite}
\frac{1}{m}\sum_{i=1}^m \begin{bmatrix}
 \varphi_{w_i}(x)\varphi_{w_i}(x')& 
 \varphi_{w_i}(x)\nabla_{x'} \varphi_{w_i}(x')^T\\[.2cm]
 \nabla_x \varphi_{w_i}(x)\varphi_{w_i}(x') & \nabla_x \varphi_{w_i}(x)\nabla_{x'}\varphi_{w_i}(x')^T
\end{bmatrix},
\end{equation}
\endgroup
\end{small}
where for compactness we used $w_i = (a_i,b_i)$, $\varphi_{w_i}(x) = [a_i x + b_i]_+$ and $\nabla_x \varphi_{w_i}(x) = {\bf 1}[a_i x + b_i]_+ a_i$.

\vspace{\titlepre}
\subsection{Infinite-Width Kernel}\label{sec:infinite_width_kernel}
\vspace{\titlepost}
As the number of neurons $m$ tends to infinity, the kernel~\eqref{eq:kernel_finite} converges to $K_{\infty}(x,x')$ defined by

\begin{small}
\begin{equation}\label{eq:kernel_infinite}
\mathbb{E}_{w \sim \mathcal D} \begin{bmatrix}
 \varphi_{w}(x)\varphi_{w}(x')& 
 \varphi_{w}(x)\nabla_{x'} \varphi_{w}(x')^T\\[.2cm]
 \nabla_x \varphi_{w}(x)\varphi_{w}(x') & \nabla_x \varphi_{w}(x)\nabla_{x'}\varphi_{w}(x')^T
\end{bmatrix},
\end{equation}
\end{small}
where $\mathcal D$ is the chosen distribution of bottom layer weights $w = (a,b)$.
We use this kernel to characterize the interpolant in the infinite-width limit. For this, we simply replace $K$ in ~\eqref{eq:kernel_finite} with $K_{\infty}$ and solve the linear system the described in the previous section. More concretely (and rearranging terms), our goal is to recover $\alpha_1^*,\ldots,\alpha_s^* \in \RR^{d+1}$ such that
\begin{equation}\label{eq:linear_eqs_kernel}
    \sum_{i=1}^s K_\infty(x_j,x_i) \alpha^*_i = \begin{bmatrix}
    y_j\\
    n_j
    \end{bmatrix} \in \RR^{d+1}, \quad j=1,\ldots,s.
\end{equation}
Our interpolant $f^*$ is such that $\begin{bmatrix}f^*(x)\\
\nabla f^*(x)\end{bmatrix}= K_{\infty}(x,x_i)\alpha_i^*$ for all $x \in \RR^d$. It can also be viewed as the solution to
\begin{equation}\label{eq:optimization}
\begin{aligned}
&\underset{f \in \mathcal{H} }{\text{minimize }} \|f\|_\mathcal{H}, \\
&\text{subject to } f(x_i) = 0 \text{ and } \nabla f(x_i) = n_i
\end{aligned}
\end{equation}
where $\mathcal{H}$ is the RKHS corresponding to the one-dimensional kernel $\mathbb E_{w} \, \varphi_w(x)\varphi_w(x') = \mathbb E_{(a,b)} [a^T x + b]_+[a^T x' + b]_+$.
 In Section~\ref{sec:implicit_bias} we give an expression for the norm $\|\cdot\|_\mathcal{H}$ that defines the inductive bias of solutions. In Appendix~\ref{sec:analytic_expressions}, we provide analytic expressions for the kernel \eqref{eq:kernel_infinite} for two natural distributions over the weights $(a,b)$: 
 \begingroup
\setlength\arraycolsep{-.5pt}
\begin{align}
&\mbox{Uniform:} &&a \sim \mathcal{U}[\SpS^{d-1}], \, b \sim \mathcal{U}[-k, k],\label{eq:uniform_initialization}\\[.2cm]
&\mbox{Gaussian:} && (a,b) \sim \mathcal{N}(0,Id_{d}).\label{eq:gaussian_initialization}
\end{align}
\endgroup
 The uniform distribution~\eqref{eq:uniform_initialization} corresponds to the default initialization of linear layers in PyTorch and, as we argue in Section~\ref{sec:cubic_splines}, it leads to a direct generalization of cubic spline interpolation. However, the Gaussian initialization actually leads to simpler analytical expressions for $K_{\infty}$ and produces almost the same results. We refer to Appendix~\ref{sec:analytic_expressions} for a discussion and comparison of the two distributions.

\section{Discussion}

\subsection{Connection to Cubic Splines in 1D}\label{sec:cubic_splines}
When $d=1$, the uniform initialization~\eqref{eq:uniform_initialization} is such that $a \sim \mathcal U(\mathbb S^0) = \mathcal U(\{-1,1\})$ and, assuming $-k \le x \le x' \le k$, the top-left element in $K_{\infty}(x,x')$ is given by
\begin{footnotesize}
\begin{equation}\label{eq:1D_spline}
\begin{aligned}
&K_{\rm spline}(x,x') =  \\
&=\frac{1}{2}\int_{-k}^k [x+b]_+ [x' + b]_+ db + \frac{1}{2}\int_{-k}^k [-x+b]_+ [-x' + b]_+ db \\
&=\frac{1}{12} \, {\left(3\,x' - x + 2\,k \right)} {\left(x + k\right)}^{2} - \frac{1}{12} \, {\left(3 \, x - x' - 2 \, k\right)} {\left(x' - k\right)}^{2}.
\end{aligned}
\end{equation}
\end{footnotesize}
The expression assuming $-k \le x' < x \le k$ is obtained by swapping $x$ and $x'$. 
For fixed $x$, the map $K_{\rm spline}(x,\cdot)$ is piecewise cubic and twice continuously differentiable ($C^2$). This implies that kernel regression with~\eqref{eq:1D_spline} yields \emph{cubic spline interpolation}. Applying the Neural Spline objective~\eqref{eq:objective} with derivative constraints at the samples in $d=1$ also yields a piecewice cubic interpolant, although this curve is in general only $C^1$. For other distributions of $a$ and $b$, the kernel is no longer cubic, but the norm in the RKHS is a weighed norm of curvature (see \cite{williams2019gradient} for details). In this sense, our approach with the initialization \eqref{eq:uniform_initialization} can be viewed as a multi-dimensional version of spline interpolation.

\vspace{\titlepre}
\subsection{Regularization and Robustness to Noise}
\begin{figure}
    \minipage{0.33\linewidth}
    \centering
    \includegraphics[width=0.99\linewidth]{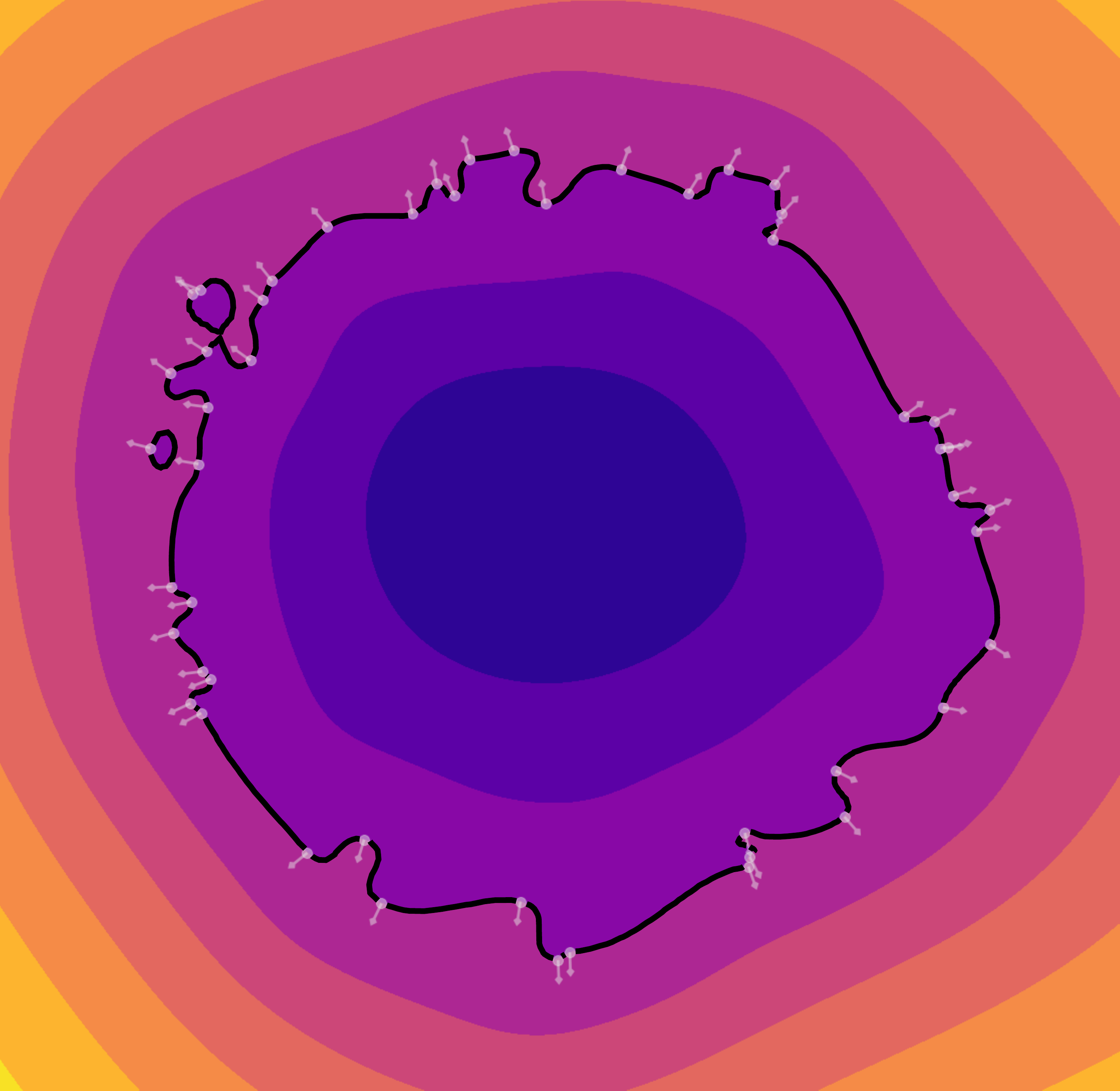}
    \endminipage
    \minipage{0.33\linewidth}
    \centering
    \includegraphics[width=0.99\linewidth]{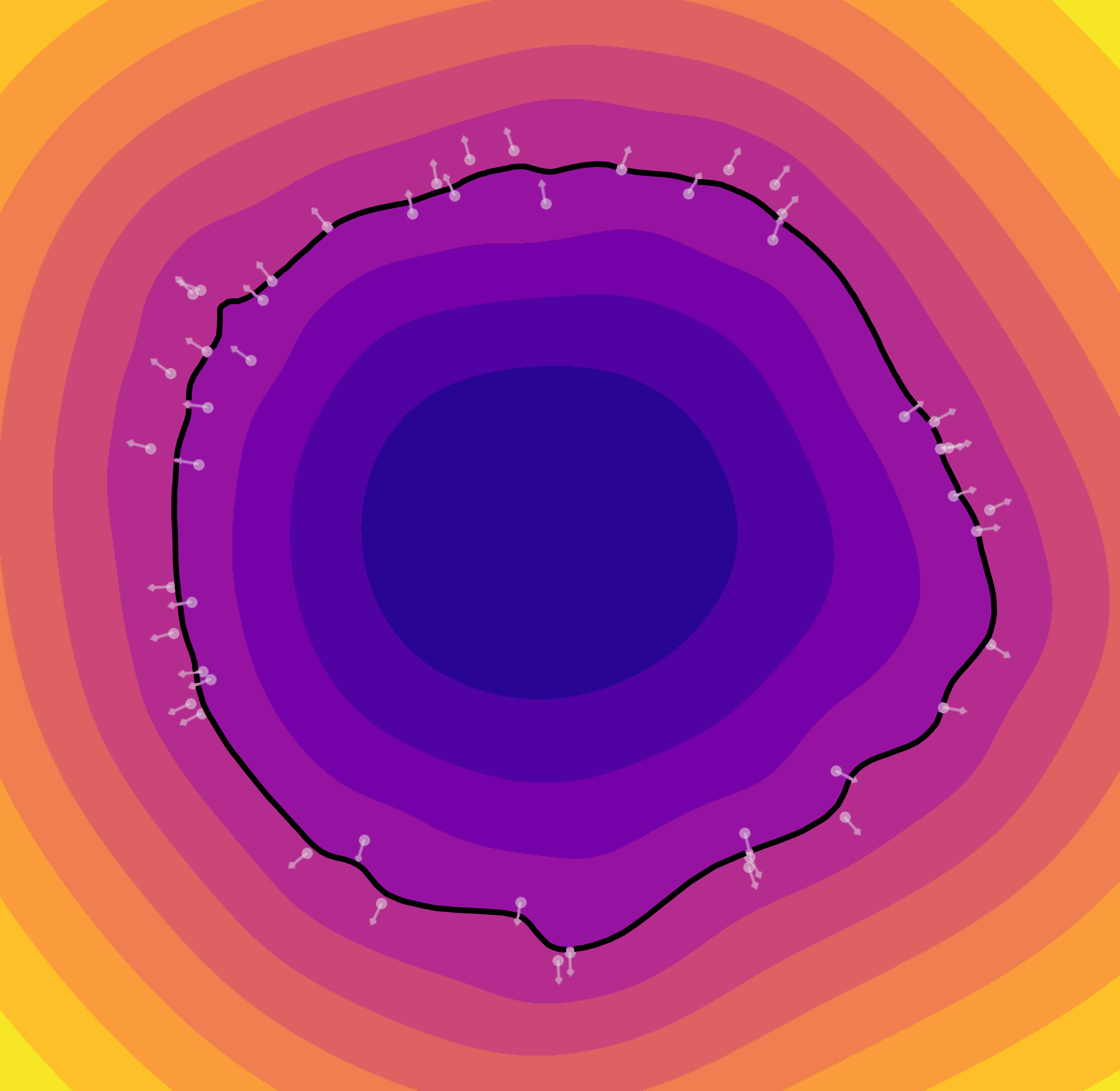}
    \endminipage
    \minipage{0.33\linewidth}
    \centering
    \includegraphics[width=0.99\linewidth]{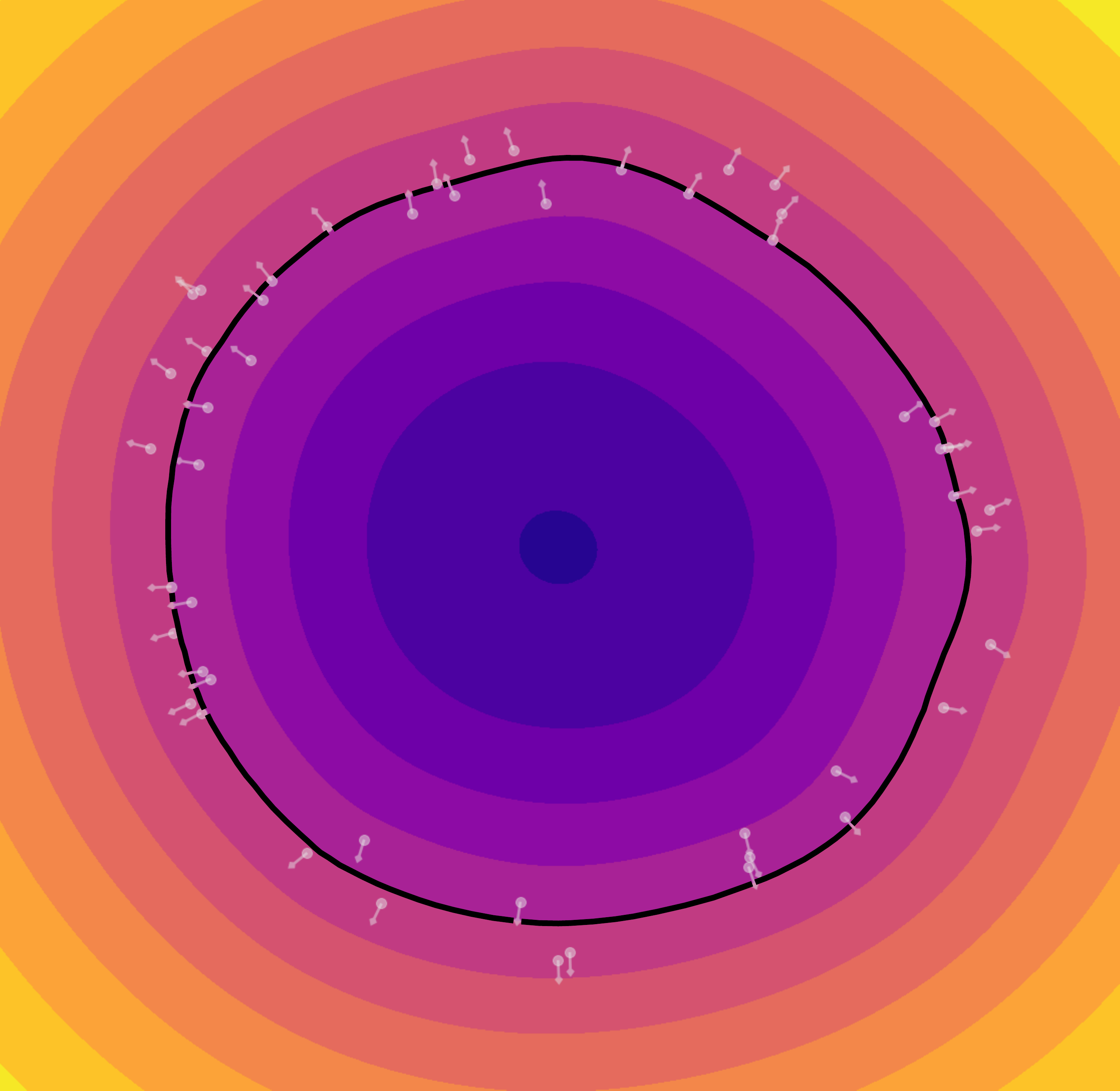}
    \endminipage\hfill
    \vspace{0.5em}
    \minipage{0.33\linewidth}
    \centering \footnotesize $\lambda = 0$
    \endminipage
    \minipage{0.33\linewidth}
    \centering \footnotesize $\lambda = 10^{-4}$
    \endminipage
    \minipage{0.33\linewidth}
    \centering \footnotesize $\lambda = 10^{-2}$
    \endminipage\hfill
    \vspace{1.0em}
    \\
    \minipage{0.25\linewidth}
    \centering
    \includegraphics[width=0.99\linewidth]{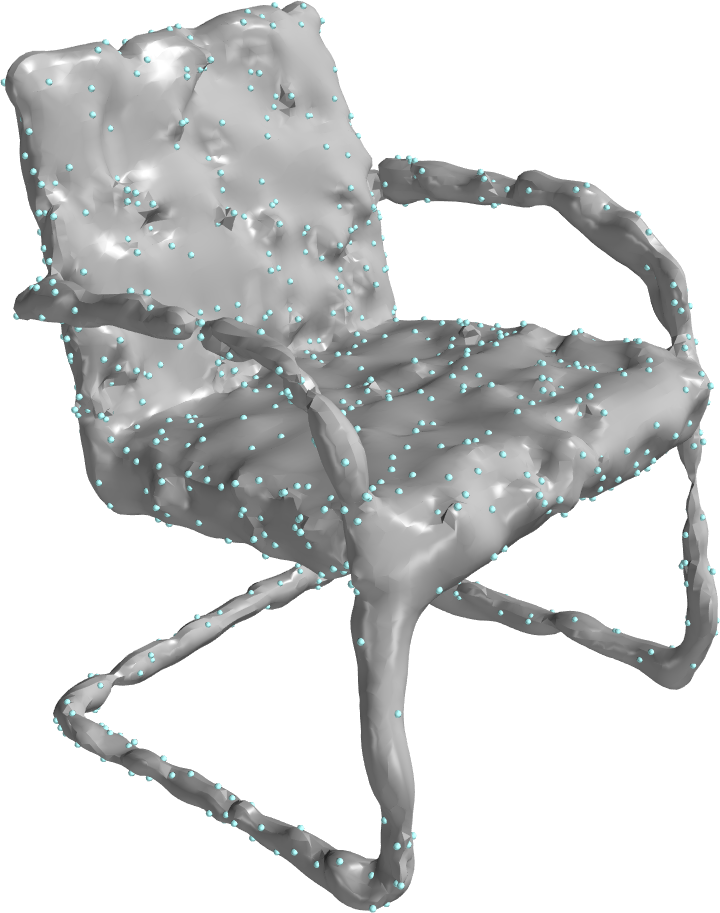}
    \endminipage
    \minipage{0.25\linewidth}
    \centering
    \includegraphics[width=0.99\linewidth]{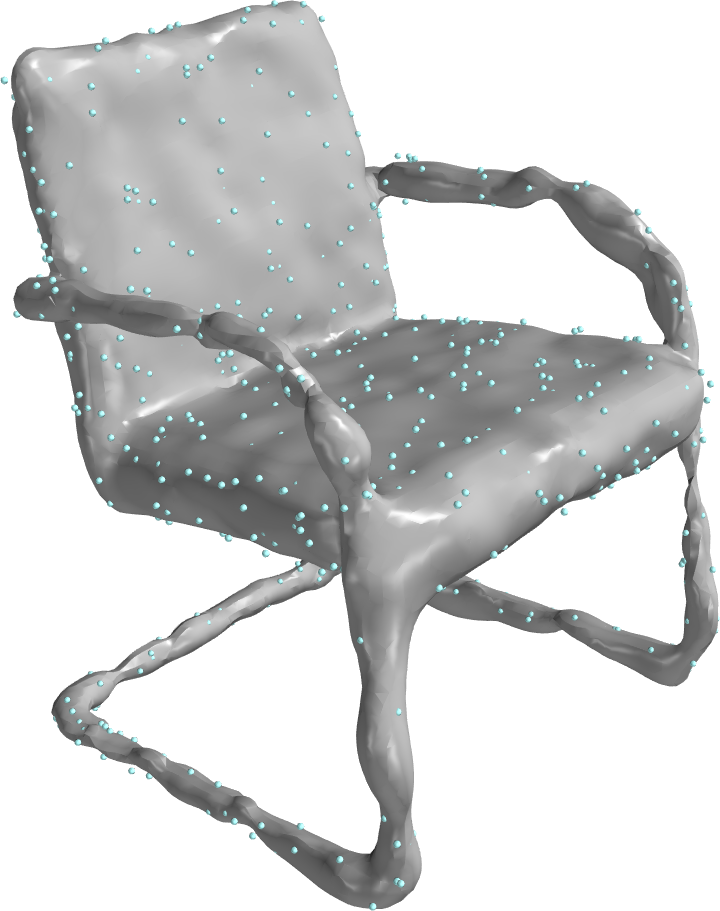}
    \endminipage
    \minipage{0.25\linewidth}
    \centering
    \includegraphics[width=0.99\linewidth]{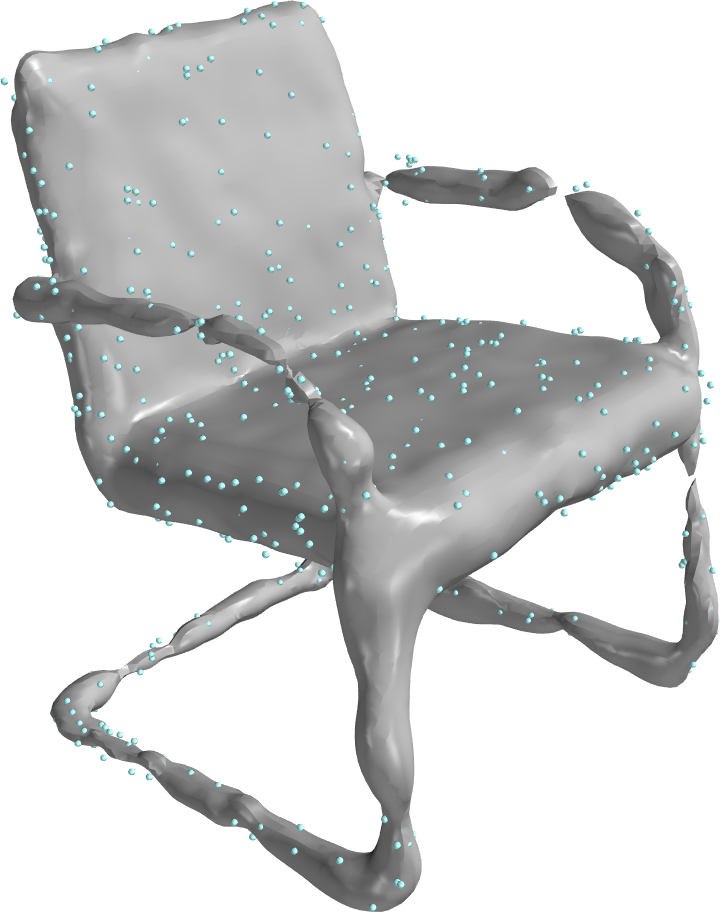}
    \endminipage
    \minipage{0.25\linewidth}
    \centering
    \includegraphics[width=0.99\linewidth]{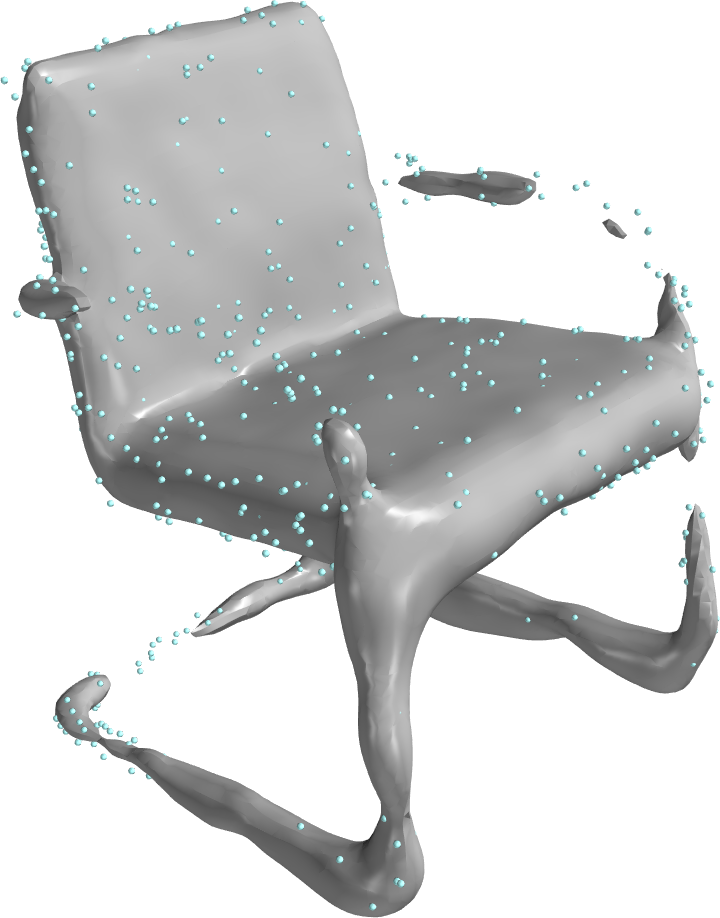}
    \endminipage\hfill
    \vspace{0.5em}
    \minipage{0.25\linewidth}
    \centering \footnotesize $\lambda = 0$
    \endminipage
    \minipage{0.25\linewidth}
    \centering \footnotesize $\lambda = 10^{-6}$
    \endminipage
    \minipage{0.25\linewidth}
    \centering \footnotesize $\lambda = 10^{-5}$
    \endminipage
    \minipage{0.25\linewidth}
    \centering \footnotesize $\lambda = 10^{-3}$
    \endminipage\hfill
    \vspace{0.5em}
    \caption{\emph{The effect of adding a regularization in 2D (Top) and 3D (Bottom).} Fitting the points and normals with no regularization yields a surface which perfectly interpolates the noisy input data (left images). Increasing $\lambda$ leads to smoother solutions which do not interpolate noise in the input. If $\lambda$ is too large, the output loses geometric details (\eg, the arms and base of the chair in the bottom right image).}\label{fig:regularization}
\end{figure}

\vspace{\titlepost}
To deal with noisy data, we can optionally add a simple regularizer term to our formulation that corresponds to penalizing the RKHS norm of the interpolant (``kernel ridge regression''). Concretely, we replace~\eqref{eq:linear_eqs_kernel} with
\begin{equation}\label{eq:linear_eqs_kernel_regularized}
    \sum_{i=1}^s K_\infty(x_j,x_i) \alpha^*_i + \delta_{ij} \lambda \, Id_{d+1} = \begin{bmatrix}
    y_j\\
    n_j
    \end{bmatrix} \in \RR^{d+1},
\end{equation}
for $j=1,\ldots,s$. The regularizer term affects the spectrum of the Gram matrix $K_{\infty}(x_j,x_i)$ by smoothing its smallest eigenvalues, with a similar effect to early stopping in gradient descent (see \eg~\cite{ali2019}). Figure \ref{fig:regularization} shows examples of applying this regularizer on 2D and 3D problems.

\vspace{\titlepre}
\subsection{Inductive Bias of Interpolants}\label{sec:implicit_bias}
Our interpolant $f^*$ belongs to the Hilbert space given by
\begin{footnotesize}
\begin{equation}\label{eq:RKHS}
\begin{split}
\begin{gathered}
\mathcal H = \big\{f(x) = \smallint c(w)\varphi_w(x) d\tau(w) \mid \|f\|_{\mathcal H} < + \infty \big\}, \mbox{ with }\\ 
         \|f\|^2_{\mathcal H} = \inf \big\{ \smallint c(w)^2 d \tau(w) \mid f(x) = \smallint c(w)\varphi_w(x) d\tau(w) \big\}
\end{gathered}
\end{split}
\end{equation}
\end{footnotesize}where $\tau(w)$ is a measure over the weights $w = (a,b)$ (\eg,  \eqref{eq:uniform_initialization} or \eqref{eq:gaussian_initialization}), and $c(a, b)$ can be viewed as a continuous analog of the outer-layer weights $c_i$ of the finite network given by \eqref{eq:finite_model}. The inductive bias of Neural Splines is thus determined by the RKHS norm in~\eqref{eq:RKHS} since our method outputs the interpolating function which minimizes that norm. 
As noted in~\cite{ongie2019function}, if $f(x) = \int c(w) \varphi_w(x) d\tau(w)$ and $d\tau(w) = d\tau_a(a) d\tau_b(b)$ then, by differentiating twice, we note that the Laplacian of $f$ is given by
\begin{equation}\label{eq:radon}
\Delta f(x) = \int_{\{ax + b = 0\}} c(a,b) d\tau_a(a).
\end{equation}
By comparing~\eqref{eq:RKHS} and~\eqref{eq:radon}, we see that the RKHS norm and the Laplacian of $f$ are closely related. More precisely, the Laplacian $\Delta f(x)$ is the \emph{Dual Radon Transform} of $c(a,b)$. 
Under certain assumptions, \eqref{eq:radon} can be inverted, yielding an explicit expression for $c$ in terms of $\Delta f$. Intuitively, this shows that bounding the RKHS norm imposes a constraint on the Laplacian of $f$, and thus encourages $f$ to be smooth. 
We report the following statement from~\cite{ongie2019function} and we refer to Appendix~\ref{sec:rkhs_norm} for more details.

\begin{proposition}\cite{ongie2019function}\label{prop:ongie} Let $f(x) = \int c(a,b) [ax + b]_+ d\tau(a,b)$, and the constant $\gamma_d = \frac{1}{2(2\pi)^{d-1}}$. If we assume that $c(a,b) = c(-a,-b)$ holds, then
\begin{equation*}
    c(a, b) = \gamma_d \frac{\mathcal{R} \{ (-\Delta)^\frac{d+1}{2} f(x) \}(a,b)}{\tau(a, b)}~,
\end{equation*} 
where $\mathcal{R}\{f\}(a, b)$ is the Radon Transform of $f$.
\end{proposition}

\vspace{\titlepre}
\subsection{Poisson Surface Reconstruction as a Kernel}
\begin{figure}
    \minipage{0.45\linewidth}
    \centering
    \includegraphics[width=0.75\linewidth, height=0.7\linewidth]{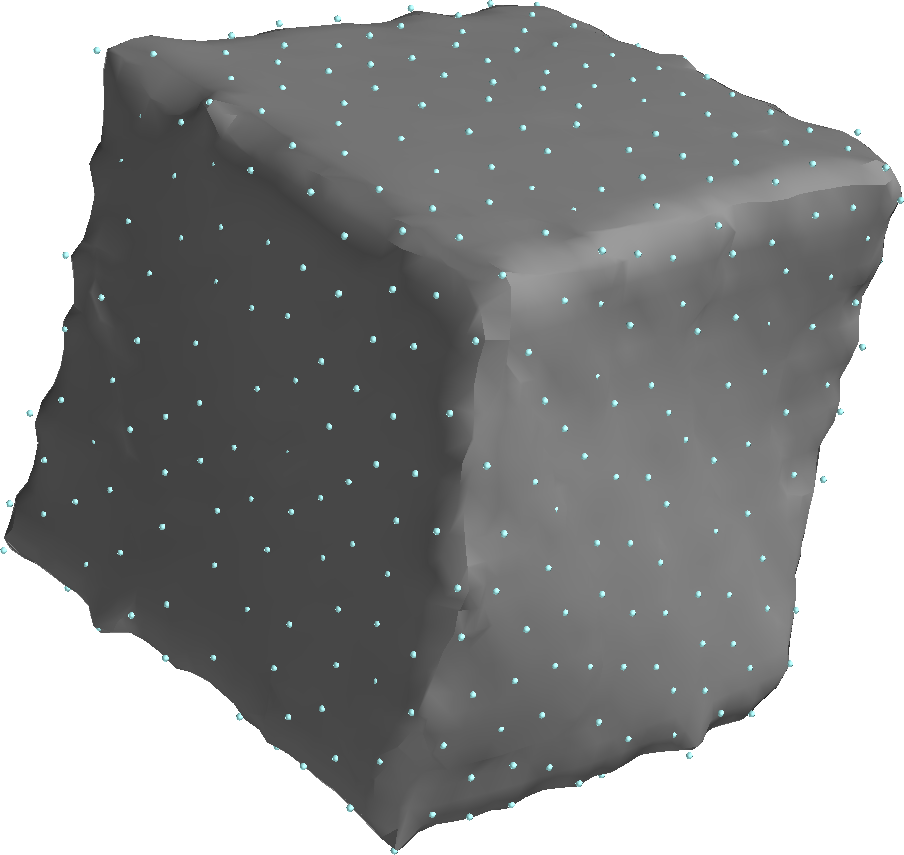}
    \endminipage\hfill
    \minipage{0.45\linewidth}
    \centering
    \includegraphics[width=0.75\linewidth, height=0.7\linewidth]{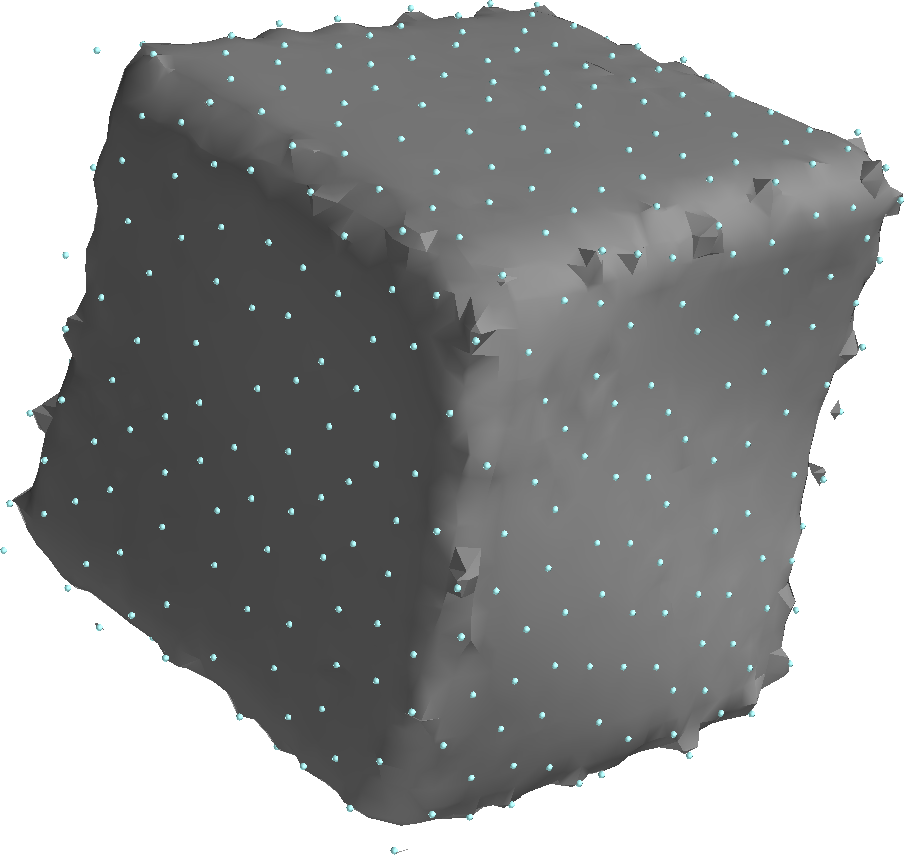}
    \endminipage\hfill
    \vspace{0.25em}
    \minipage{0.495\linewidth}
    \vspace{0.175em}
    \centering \footnotesize Approximate Poisson Kernel
    \endminipage\hfill
    \vspace{-0.175em}
    \minipage{0.495\linewidth}
    \centering \footnotesize Poisson Surface Reconstruction
    \endminipage\hfill
    \vspace{0.5em}
    \caption{\emph{Left:} Poisson reconstruction using the approximate Kernel \eqref{eq:k_approx}. \emph{Right:} Screened Poisson Reconstruction of the same input.}
    \label{fig:approx_poisson}
\end{figure}

\vspace{\titlepost}
\label{sec:PRK}
\newcommand{\F}{\mathcal{F}}
\newcommand{\Hk}{\mathcal{H}}
\newcommand{\R}{\mathbb{R}}
\newcommand{\sinc}{\mathrm{sinc}}
We cast Screened Poisson Surface Reconstruction \cite{kazhdan2013screened} in kernel form to facilitate comparisons. In its simplest form, Poisson reconstruction, extracts the level set of a smoothed indicator function determined as the solution of 
\begin{equation*}
-\Delta f  = \nabla \cdot V,
\end{equation*}
where $V$ is a vector field obtained from normals $n_i$ at samples $x_i$, and we use $f$ to denote the (smoothed) indicator function as it plays the same role as  $f$ in \eqref{eq:objective}. 
The equation above is closely related to \eqref{eq:objective}:  specifically, it is the equation for the minimizer of $\int_{\mathbb{R}^3}\|\nabla_x f(x)-V\|^2dx$ and the second term in \eqref{eq:objective} can be viewed as a approximation of this term by sampling at $x_i$. 
The screened form of Poisson reconstruction effectively adds the first term with $y_i =0$, as the indicator function at points of interest is supposed to be zero. 
For the Poisson equation, the solution can be explicitly written as an integral 
\begin{equation*}
f(x) = \int_{\mathbb{R}^3} \frac{\nabla_z \cdot V(z) dz}{|x-z|}.
\end{equation*}
The vector field $V$ is obtained by interpolating the normals using a fixed-grid spline basis and barycentric coordinates of the sample points with respect to the grid cell containing it. 
This is equivalent to using a non-translation invariant non-symmetric locally-supported kernel $K_B(z,x)$: 
\begin{equation*}
V(z) = \sum_i K_B(z,x_i) n_i~.
\end{equation*}
\begin{lemma}\label{lemma:poisson_kernel} 
Let $\{c_j \in \RR^3\}_{j=1}^g$ be a set of points arranged on a regular grid, $B_1(x-z)$ be the trilinear basis function, and $B_n(x-z)$ be a degree-$n$ spline basis function (See Appendix~\ref{sec:poisson_kernel_details} for equations for $B_1$ and $B_n$). The kernel corresponding to Poisson Surface Reconstruction is
\begin{equation}
\begin{aligned}
&K_{\text{\rm PR}}(x,x') =  \int_{\mathbb{R}^3} \frac{ K_B(z,x')  dz}{|x-z|},\\
\end{aligned}
\end{equation}
where $K_B(z,x) = \sum_j B_1(x-c_j) B_n(z-c_j)$.
\end{lemma}
To study the qualitative properties of this kernel, we replace $K_B(z,x)$ with a radial kernel $B^1_n(|z-x|)$ (see Appendix~\ref{sec:poisson_kernel_details}) which has qualitatively similar behavior (see Figure~\ref{fig:approx_poisson}). Since both $B^1_n$ and the Laplace kernel $\frac{1}{|x - z|}$ are radial functions, their convolution is also radial, yielding a translation-invariant radial approximation $K_{\text{PR}}^{\text{approx}}$ of $K_{\text{PR}}$:
\begin{equation}\label{eq:k_approx}
K^{\text{approx}}_{\text{PR}}(x,x') = \int_{\R^3} \frac{ B^1_n(|z-x'|)  dz}{|x-z|}.
\end{equation}
\begin{lemma}
The RKHS norm of the corresponding to the approximate Poisson kernel $K_{{\rm PR}}^{{\rm approx}}$ is 
\begin{equation}
\|f\|_{\mathcal H} = \int \frac{|\F[f]|^2}{\F[K^{\text{\rm approx}}_{\text{PR}}]}d\omega
\end{equation} where $\F[\cdot]$ is the Fourier transform.
\end{lemma} We discuss the kernel formulation of Poisson Reconstruction in more detail in Appendix~\ref{sec:poisson_kernel_details}.

\vspace{\titlepre}
\subsection{Fast and Scalable Implementation}\label{sec:fast_implementation}
\vspace{\titlepost}
We provide a fast and scalable implementation of Neural Spline kernels based on FALKON \cite{rudi2018falkon}, a recently-proposed solver for kernel-ridge-regression which runs in parallel on the GPU. While na\"{i}ve kernel ridge-regression with $N$ points requires solving and storing an $N \times N$ dense linear system, FALKON uses conjugate gradient descent requiring only $\mathcal{O}(N)$ storage, and $\sqrt{N}$ convergence (though in practice we find that even for very large inputs, we converge in fewer than 10 iterations). 
To speed up convergence, FALKON can optionally store an $M \times M$ preconditioner matrix in CPU memory (where $M \ll N$, see paragrph below). To maximize performance and reduce memory overhead, we rely on KeOps \cite{charlier2020kernel} to evaluate kernel matrix-vector products symbolically on the GPU, which means our implementation uses only a \emph{small constant} amount GPU memory and can be readily used on commodity hardware. Section~\ref{sec:performance} compares the performance of our implementation against other state of the art surface reconstruction techniques.  We note that in principle, low-dimensional kernel methods can be accelerated using fast multipole-based approaches  \cite{greengard1987fast} (in particular, in the context of 3D surface reconstruction this was used in \cite{carr2001reconstruction}); this yields optimal $O(N)$ time  complexity, for dense matrix-vector multiplication. 

\begin{figure}
    \minipage{0.245\linewidth}
    \centering
    \includegraphics[width=0.99\linewidth]{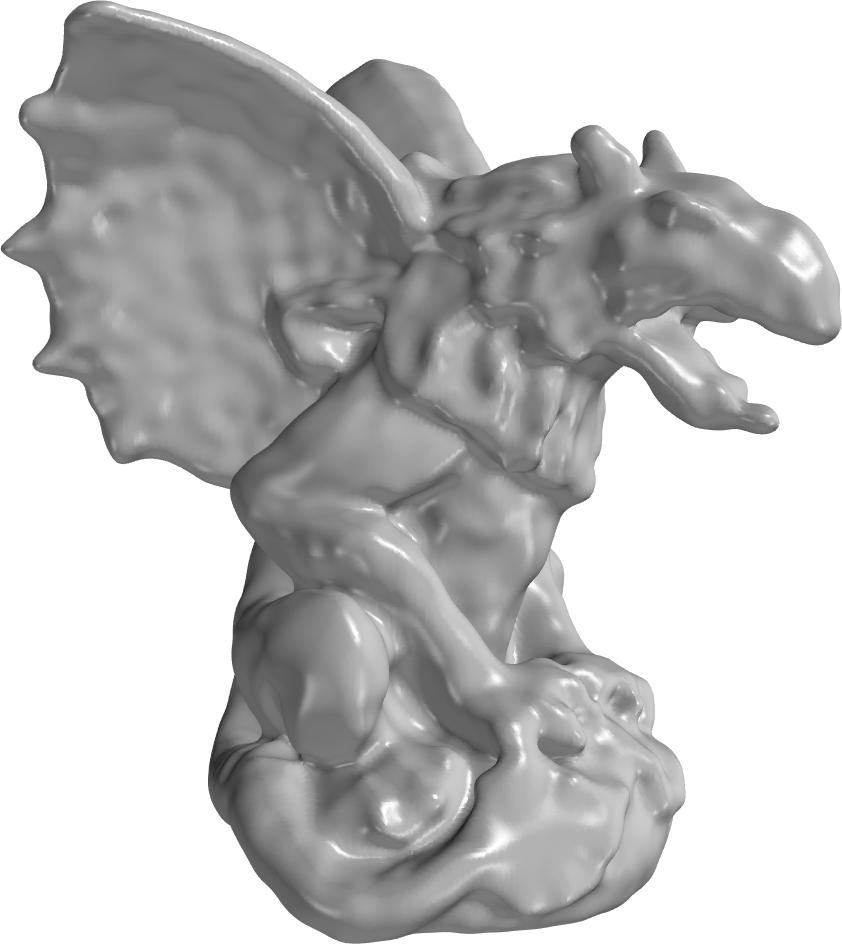}
    \endminipage\hfill
    \minipage{0.245\linewidth}
    \centering
    \includegraphics[width=0.99\linewidth]{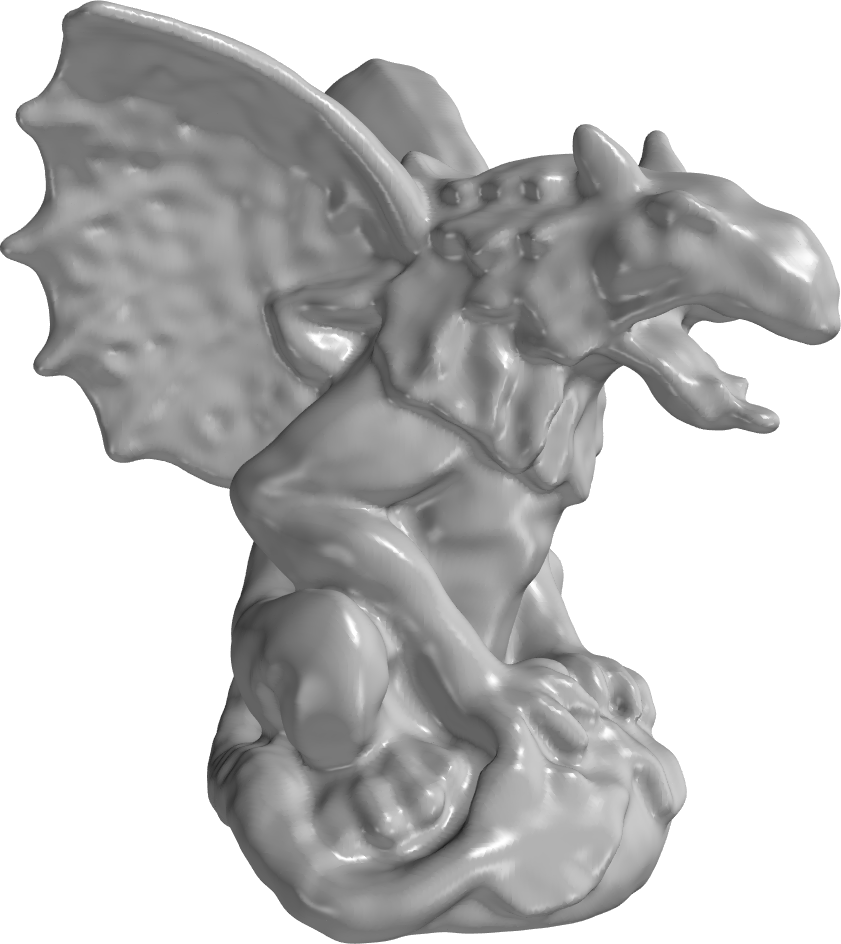}
    \endminipage\hfill
    \minipage{0.245\linewidth}
    \centering
    \includegraphics[width=0.99\linewidth]{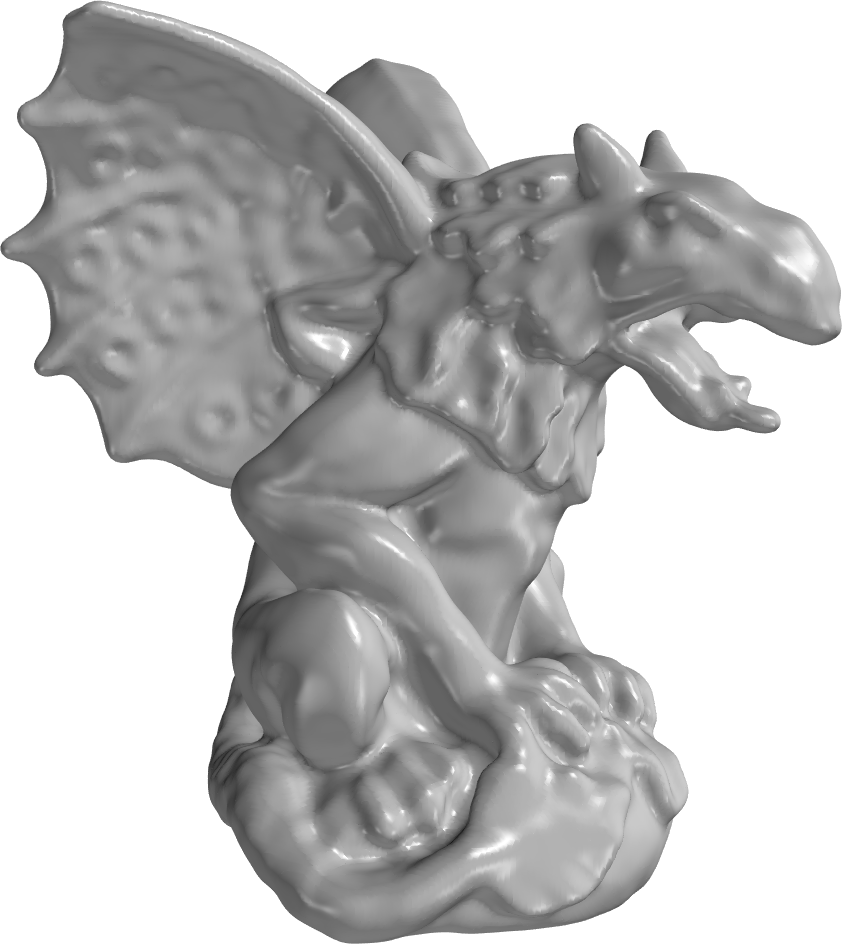}
    \endminipage\hfill
    \minipage{0.245\linewidth}
    \centering
    \includegraphics[width=0.99\linewidth]{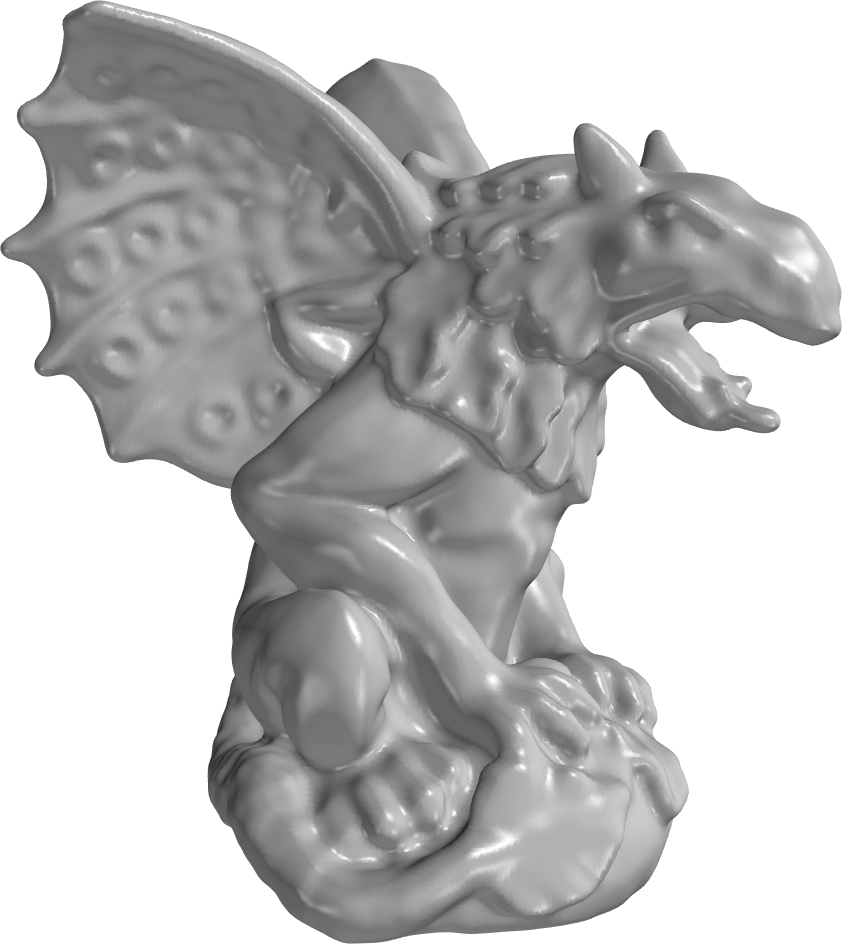}
    \endminipage\hfill
    \vspace{0.5em}
    \hfill
    \minipage{0.245\linewidth}
    \centering \footnotesize 5k Nystr\"{o}m
    \endminipage\hfill
    \minipage{0.245\linewidth}
    \centering \footnotesize 10k Nystr\"{o}m
    \endminipage\hfill
    \minipage{0.245\linewidth}
    \centering \footnotesize 15k Nystr\"{o}m
    \endminipage\hfill
    \minipage{0.245\linewidth}
    \centering \footnotesize 75k Nystr\"{o}m
    \endminipage\hfill
    \vspace{0.5em}
    \caption{Fitting a range scan with 100k points using varying numbers of Nystr\"{o}m samples. Larger numbers of Nystr\"{o}m samples will lead to reconstructions which preserve finer details (\eg the bumps on the wing of the gargoyle). In this case, 15\% of the input samples recovers approximately the same level of detail as 75\%. }
    \label{fig:nystrom}
\end{figure}

\paragraph{Nystr\"{o}m Subsampling}
Full kernel ridge regression predicts a function which is supported on every input point $x_i$ as in \eqref{eq:linear_eqs_kernel}, requiring $N$ coefficients to store the resulting function. We rely on Nystr\"{o}m sampling \cite{drineas2005nystrom} to instead produce a kernel function which is supported on a small $M$-sized subset of the input points (while still minimizing a loss on all the points). This is equivalent to approximating the kernel matrix with a low rank linear system. To choose Nystr\"{o}m samples, we leverage the geometric nature of our problem and select these by downsampling the input point cloud to have a blue-noise distribution using Bridson's algorithm \cite{bridson2007fast}. We demonstrate the effect of varying the number of Nystr\"{o}m samples qualitatively in Figure~\ref{fig:nystrom}. 

\vspace{\titlepre}
\section{Experiments and Results}
\vspace{\titlepost}
We now demonstrate the effectiveness of Neural Splines on the task of surface reconstruction. For all the experiments in this section, we used the analytical form of the kernel \eqref{eq:kernel_infinite} with Gaussian initialization \eqref{eq:uniform_initialization}. Appendix~\ref{sec:us_vs_us} compares the uniform \eqref{eq:uniform_initialization} and Gaussian \eqref{eq:gaussian_initialization} kernels, showing almost no measurable difference in the reconstructions produced by either. We compare the empirical and analytical kernels in detail in Appendix~\ref{sec:empirical_vs_analytical}. An implementation of Neural Splines is available at \url{https://github.com/fwilliams/neural-splines}.

\vspace{\titlepre}
\subsection{Sparse Reconstruction on Shapenet}\label{sec:benchmark} 
\vspace{\titlepost}
We performed a quantitative evaluation on a subset of the Shapenet dataset \cite{chang2015shapenet} to demonstrate that the inductive bias of Neural Splines is particularly effective for reconstructing surfaces from sparse points. We chose 1024 random points and normals sampled from the surface of 20 shapes per category across 13 categories (totalling 260 shapes).
Using this dataset, we compared our method against Implicit Geometric Regularization (IGR) \cite{gropp2020implicit}, SIREN \cite{sitzmann2020implicit}, Fourier Feature Networks \cite{tancik2020fourier}, Biharmonic RBF (Biharmonic) \cite{carr2001reconstruction}, SVM surface modelling (SVR) \cite{NIPS2004_2724}, and Screened Poisson Surface Reconstruction (Poisson) \cite{kazhdan2013screened}. The first three techniques are modern neural network based methods, while the latter three techniques are classical methods based on kernels or solving a PDE.
As criteria for the benchmark, we consider the Intersection over Union (IoU) and Chamfer Distance between the reconstructed shapes and the ground truth shapes. The former metric captures the accuracy of the predicted occupancy function, while the latter metric captures the accuracy of the predicted surface. Under both metrics, our method outperforms all other methods by a large margin. Table~\ref{tbl:quantitative_stats} reports quantitative results for the experiment and Figure~\ref{fig:comparison_3d} shows visual results on a few models. We report per-category results in Appendix~\ref{sec:per_category_stats} and show many more figures in Appendix~\ref{sec:allfigs}.

\newcommand{\snfigspacing}{0.1427}
\newcommand{\snfigscale}{0.7}
\begin{figure*}
    \minipage{\snfigspacing\linewidth}
    \centering
    \includegraphics[width=\snfigscale\linewidth]{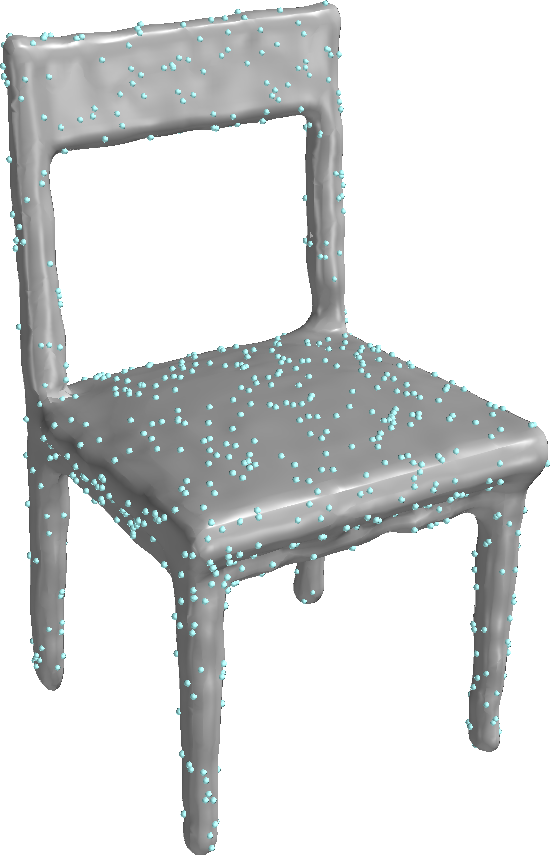}
    \endminipage\hfill
    \minipage{\snfigspacing\linewidth}
    \centering
    \includegraphics[width=\snfigscale\linewidth]{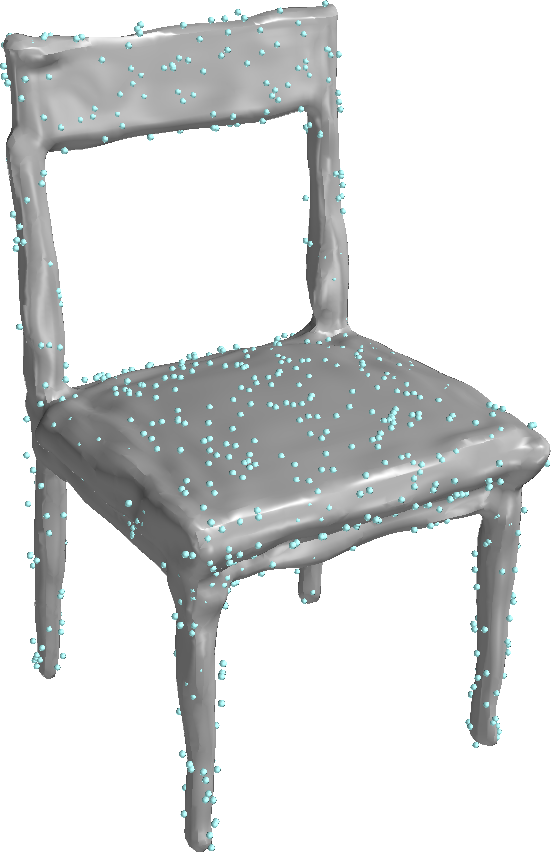}
    \endminipage\hfill
    \minipage{\snfigspacing\linewidth}
    \centering
    \includegraphics[width=\snfigscale\linewidth]{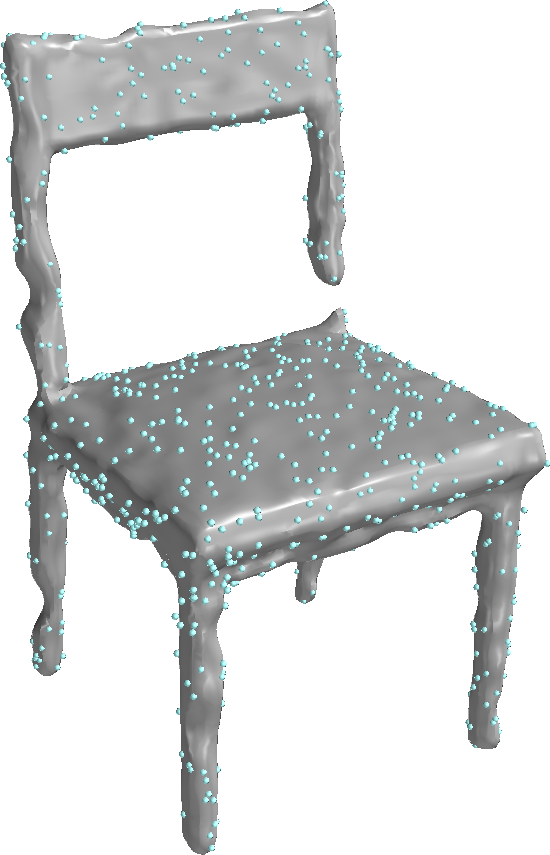}
    \endminipage\hfill
    \minipage{\snfigspacing\linewidth}
    \centering
    \includegraphics[width=\snfigscale\linewidth]{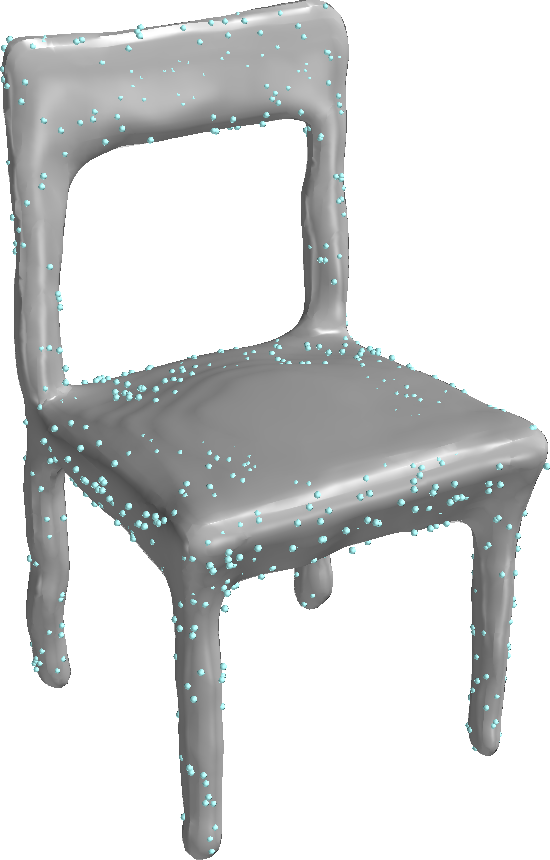}
    \endminipage\hfill
    \minipage{\snfigspacing\linewidth}
    \centering
    \includegraphics[width=\snfigscale\linewidth]{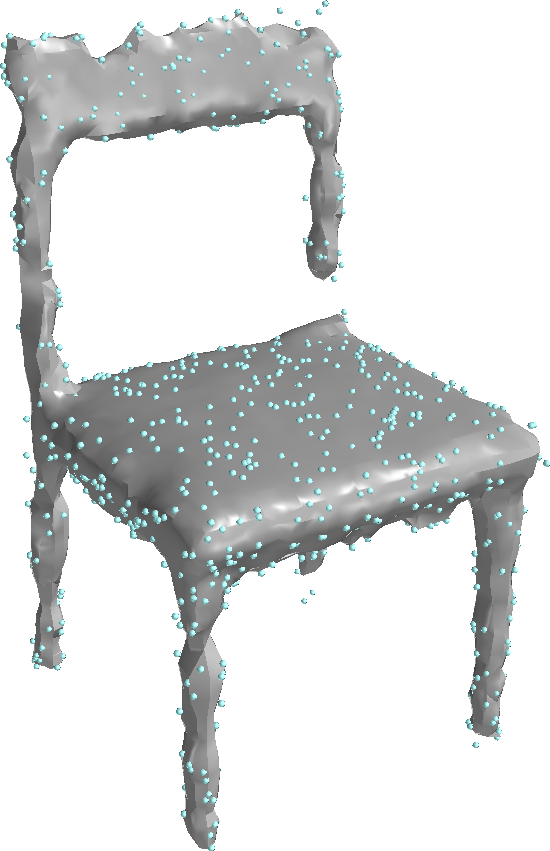}
    \endminipage\hfill
    \minipage{\snfigspacing\linewidth}
    \centering
    \includegraphics[width=\snfigscale\linewidth]{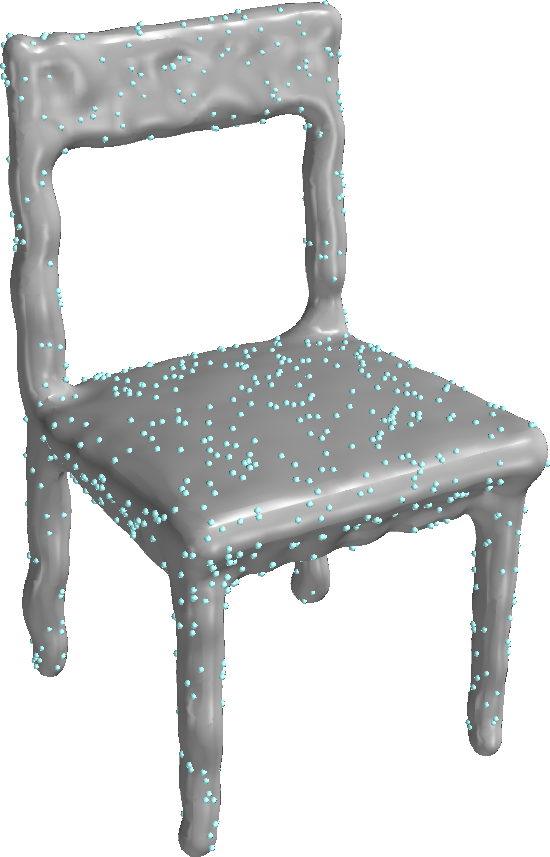}
    \endminipage\hfill
    \minipage{\snfigspacing\linewidth}
    \centering
    \includegraphics[width=\snfigscale\linewidth]{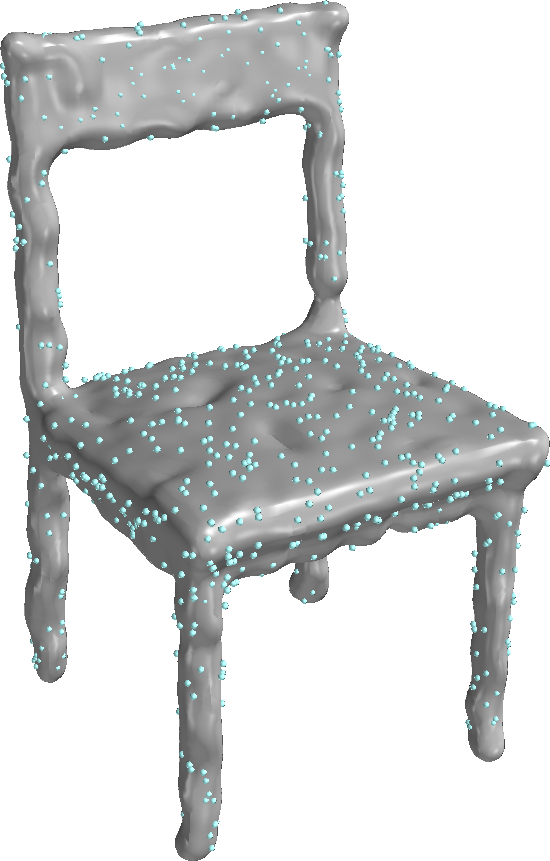}
    \endminipage\hfill
    \\
    \minipage{\snfigspacing\linewidth}
    \centering
    \includegraphics[width=\snfigscale\linewidth]{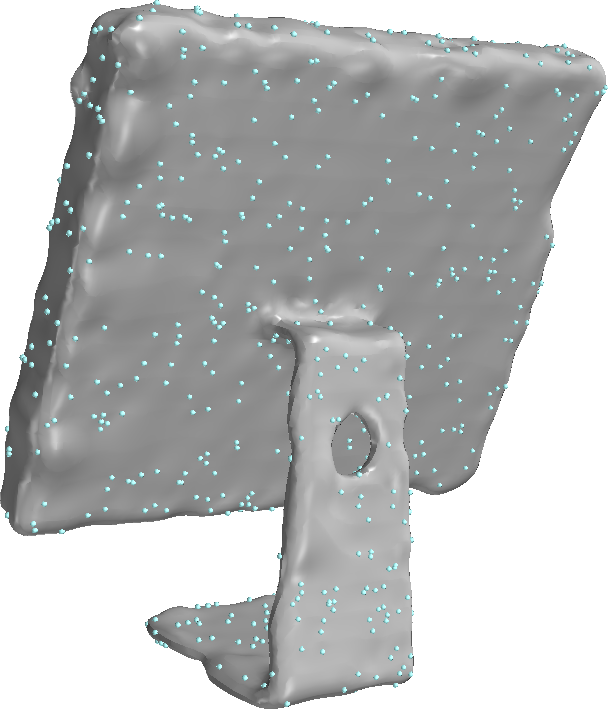}
    \endminipage\hfill
    \minipage{\snfigspacing\linewidth}
    \centering
    \includegraphics[width=\snfigscale\linewidth]{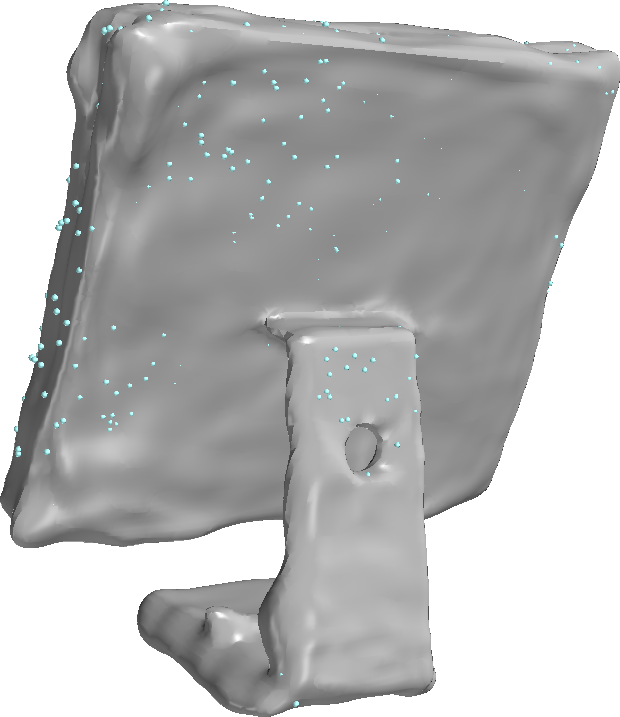}
    \endminipage\hfill
    \minipage{\snfigspacing\linewidth}
    \centering
    \includegraphics[width=\snfigscale\linewidth]{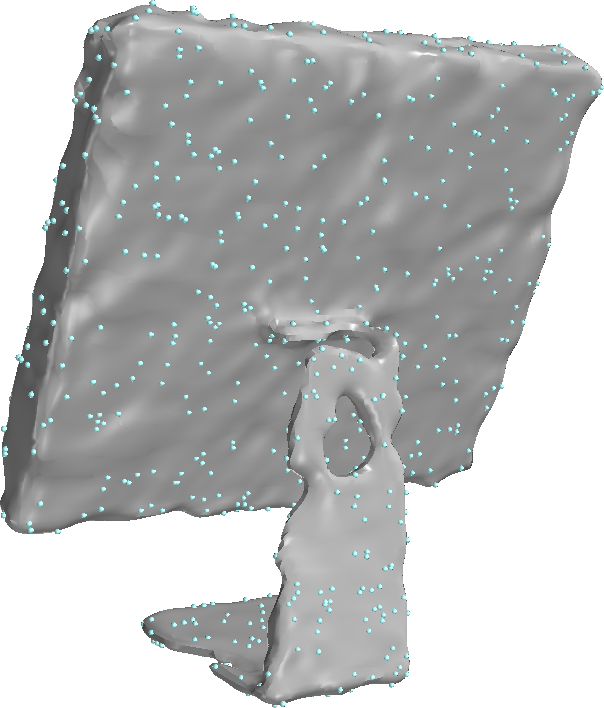}
    \endminipage\hfill
    \minipage{\snfigspacing\linewidth}
    \centering
    \includegraphics[width=\snfigscale\linewidth]{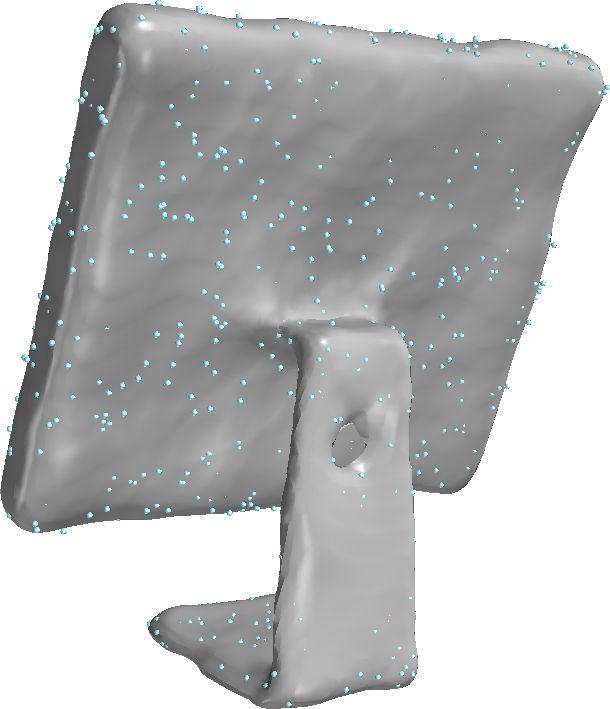}
    \endminipage\hfill
    \minipage{\snfigspacing\linewidth}
    \centering
    \includegraphics[width=\snfigscale\linewidth]{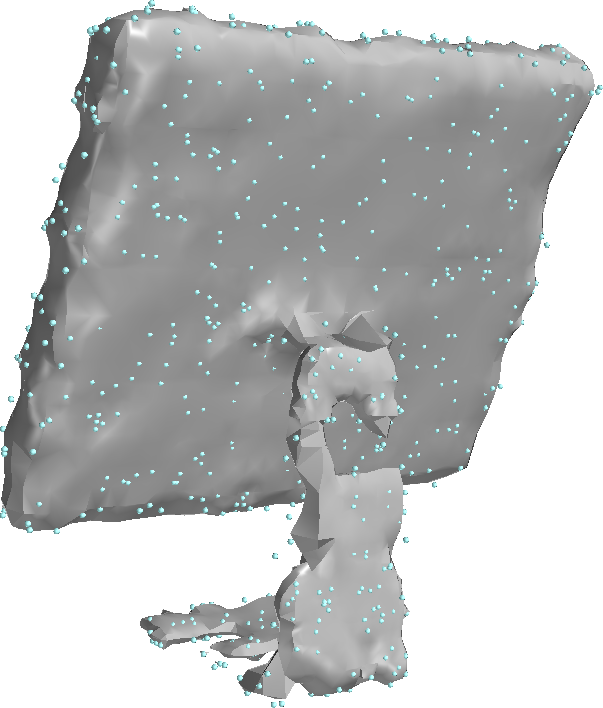}
    \endminipage\hfill
    \minipage{\snfigspacing\linewidth}
    \centering
    \includegraphics[width=\snfigscale\linewidth]{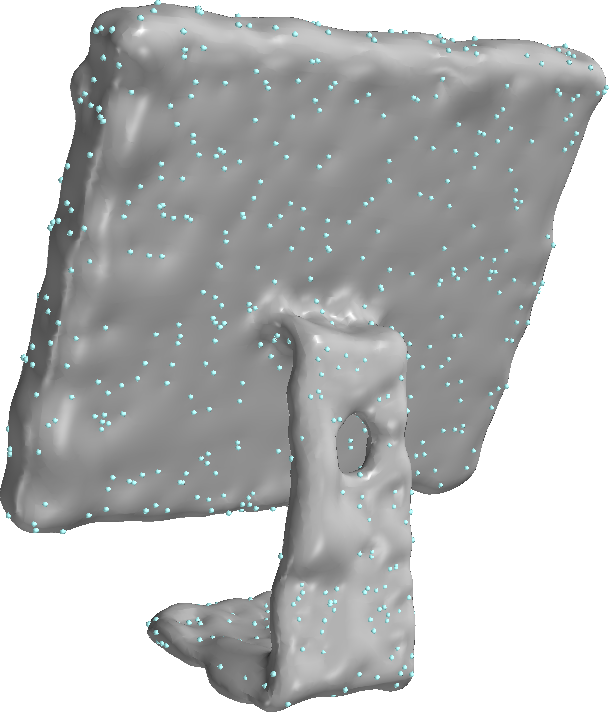}
    \endminipage\hfill
    \minipage{\snfigspacing\linewidth}
    \centering
    \includegraphics[width=\snfigscale\linewidth]{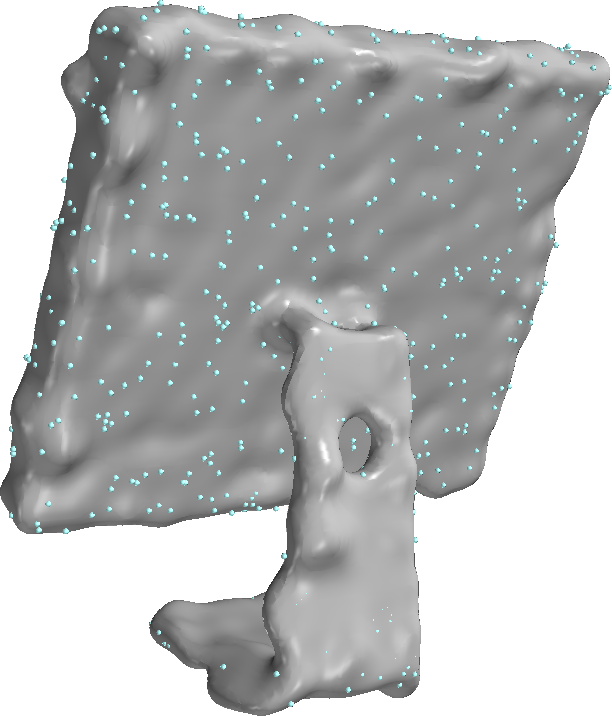}
    \endminipage\hfill
    \vspace{0.5em}
    \\
    \minipage{\snfigspacing\linewidth}
    \centering
    \includegraphics[width=\snfigscale\linewidth]{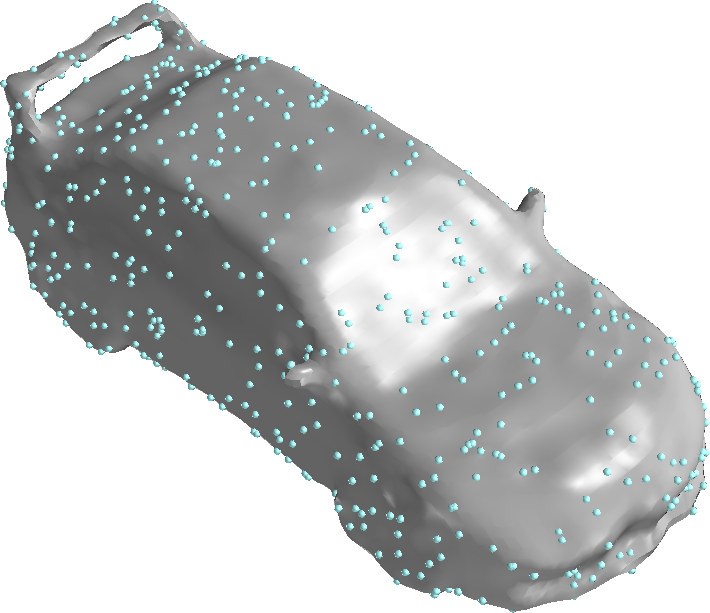}
    \endminipage\hfill
    \minipage{\snfigspacing\linewidth}
    \centering
    \includegraphics[width=\snfigscale\linewidth]{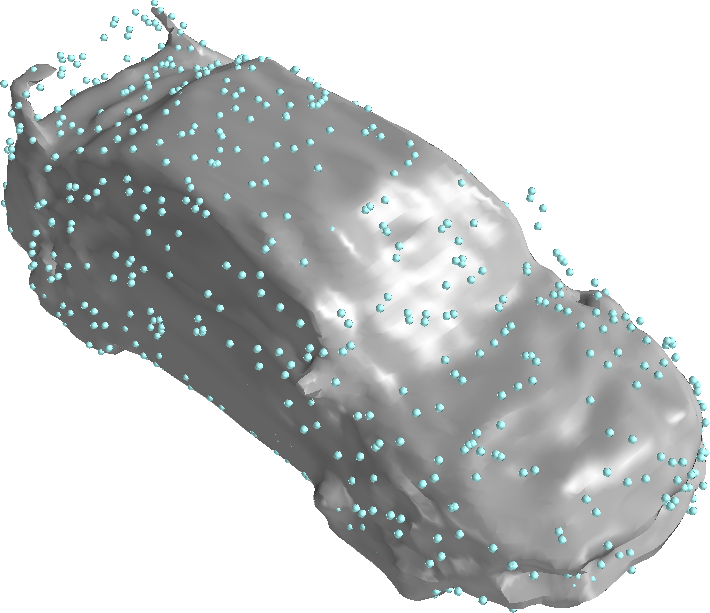}
    \endminipage\hfill
    \minipage{\snfigspacing\linewidth}
    \centering
    \includegraphics[width=\snfigscale\linewidth]{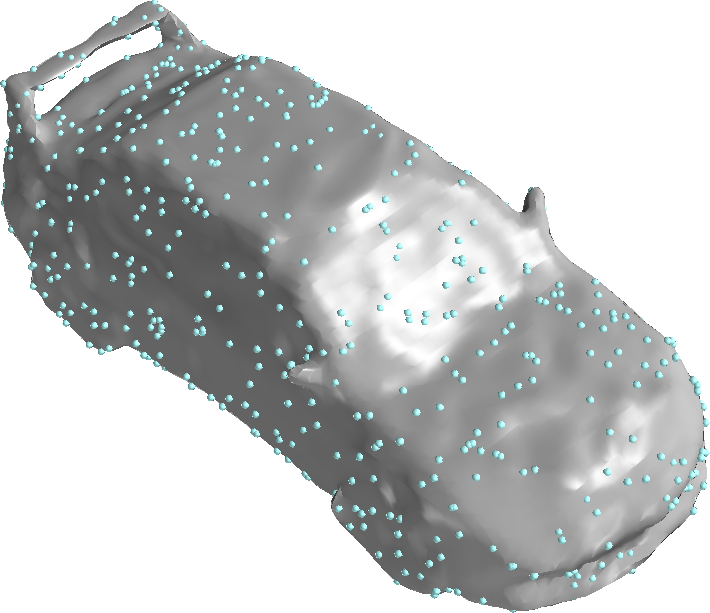}
    \endminipage\hfill
    \minipage{\snfigspacing\linewidth}
    \centering
    \includegraphics[width=\snfigscale\linewidth]{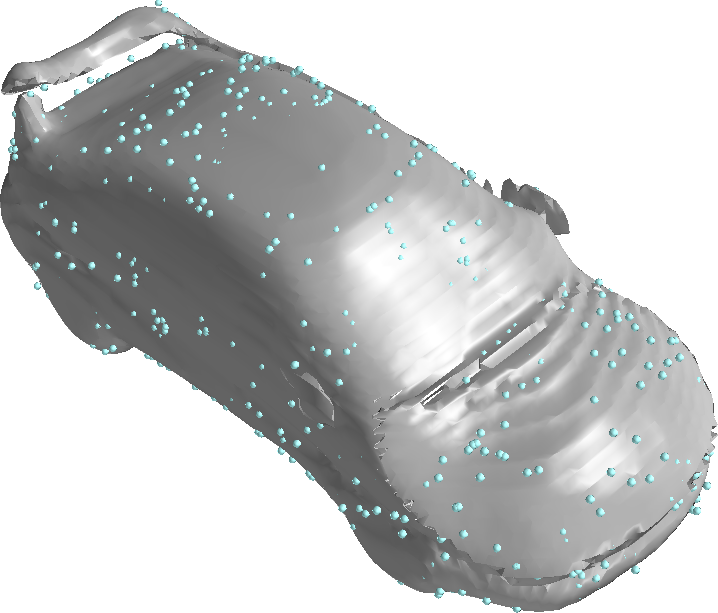}
    \endminipage\hfill
    \minipage{\snfigspacing\linewidth}
    \centering
    \includegraphics[width=\snfigscale\linewidth]{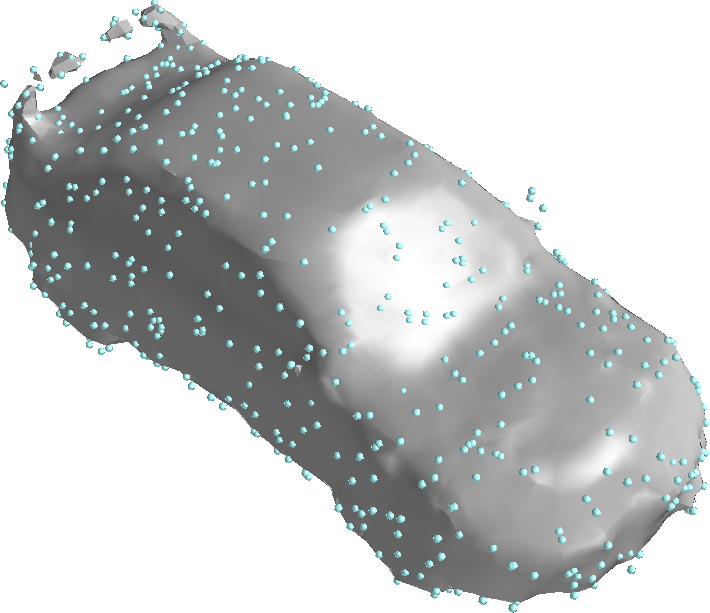}
    \endminipage\hfill
    \minipage{\snfigspacing\linewidth}
    \centering
    \includegraphics[width=\snfigscale\linewidth]{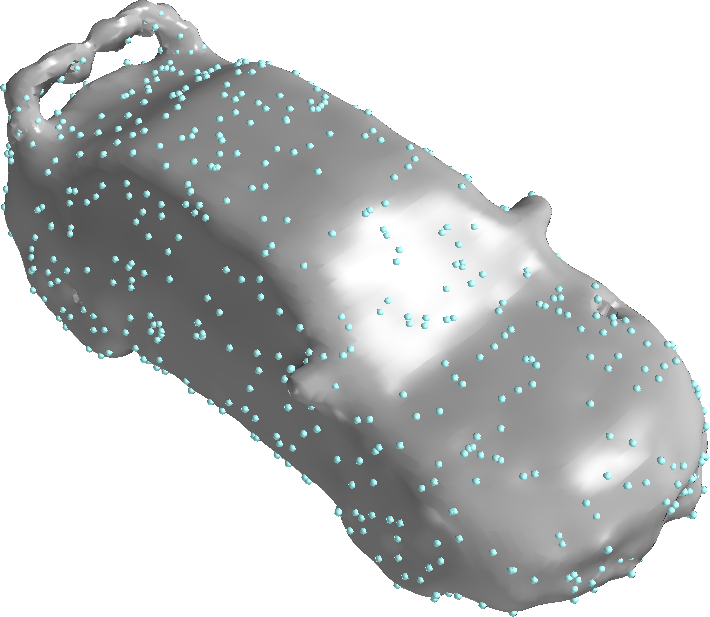}
    \endminipage\hfill
    \minipage{\snfigspacing\linewidth}
    \centering
    \includegraphics[width=\snfigscale\linewidth]{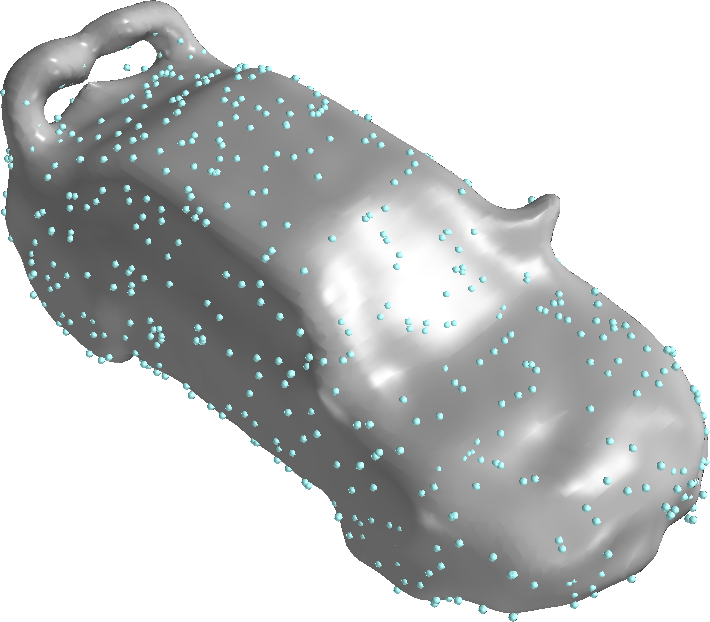}
    \endminipage\hfill
    \vspace{0.75em}
    \hfill
    \minipage{\snfigspacing\linewidth}
    \centering \footnotesize \textbf{Ours}
    \endminipage\hfill
    \minipage{\snfigspacing\linewidth}
    \centering \footnotesize IGR
    \endminipage\hfill
    \minipage{\snfigspacing\linewidth}
    \centering \footnotesize SIREN
    \endminipage\hfill
    \minipage{\snfigspacing\linewidth}
    \centering \footnotesize Fourier Feats
    \endminipage\hfill
    \minipage{\snfigspacing\linewidth}
    \centering \footnotesize Poisson
    \endminipage\hfill
    \minipage{\snfigspacing\linewidth}
    \centering \footnotesize Biharmonic
    \endminipage\hfill
    \minipage{\snfigspacing\linewidth}
    \centering \footnotesize SVR
    \endminipage\hfill
    \vspace{0.5em}
    \caption{\emph{Comparisons between reconstruction techniques on ShapeNet models.} For techniques requiring parameter sweeps, we show the result with the highest IoU. See Appendix~\ref{sec:allfigs} for more figures from each Shapenet class.}
    \label{fig:comparison_3d}
\end{figure*}

\begin{table}
    \begin{scriptsize}
    \begin{center}
    \begin{tabular}{c | c | c | c | c }
                                 & Method     & mean                & median           & std \\
        \hline
        \multirow{6}{*}{IoU}     & Screened Poisson \cite{kazhdan2013screened}  & 0.6340           & 0.6728           & 0.1577 \\
                                 & IGR \cite{gropp2020implicit}                 & 0.8102           & 0.8480           & 0.1519 \\
                                 & SIREN \cite{sitzmann2020implicit}            & 0.8268           & 0.9097           & 0.2329 \\
                                 & Fourier Feat. Nets \cite{tancik2020fourier}  & 0.8218           & 0.8396           & 0.0989 \\
                                 & Biharmonic RBF \cite{carr2001reconstruction} & 0.8247           & 0.8642           & 0.1350 \\
                                 & SVR \cite{NIPS2004_2724}                     & 0.7625           & 0.7819           & 0.1300 \\
                                 & \textbf{Ours}                                & \textbf{0.8973}   & \textbf{0.9230}  & 0.0871 \\
        \hline
        \multirow{6}{*}{Chamfer} & Screened Poisson \cite{kazhdan2013screened}  & 2.22e-4          & 1.70e-4          & 1.76e-4 \\
                                 & IGR \cite{gropp2020implicit}                 & 5.12e-4          & 1.13e-4          & 2.15e-3 \\
                                 & SIREN \cite{sitzmann2020implicit}            & 1.03e-4          & 5.28e-5          & 1.93e-4 \\
                                 & Fourier Feat. Nets \cite{tancik2020fourier}  & 9.12e-5          & 8.65e-5          & 3.36e-5 \\
                                 & Biharmonic RBF \cite{carr2001reconstruction} & 1.11e-4          & 8.97e-5          & 7.06e-5 \\
                                 & SVR \cite{NIPS2004_2724}                     & 1.14e-4          & 1.04e-4          & 5.99e-5 \\
                                 & \textbf{Ours}                                &\textbf{5.36e-5}  & \textbf{4.06e-5} & 3.64e-5 \\
    \end{tabular}
    \end{center}
    \end{scriptsize}
    \caption{Quantitative results on ShapeNet: Our method quantitatively outperforms state of the art neural network based methods and classical methods by a large marchin in both IoU and Chamfer distances.}
    \label{tbl:quantitative_stats}
\end{table}

\paragraph{Parameter Selection} Several of the above methods have parameters which can be tuned to increase performance. To ensure a fair comparison, we ran parameters sweeps for each Shapenet model for these methods, reporting the maximimum of each metric under consideration (see Appendix~\ref{sec:experiment_details_shapenet} for a detailed description). For our method we \emph{did not} tune parameters. We used no regularization and 1024 as Nystr\"{o}m samples for all models.

\vspace{\titlepre}
\subsection{Surface Reconstruction Benchmark}\label{sec:srb_comp}
\vspace{\titlepost}
We evaluated our method on the Surface Reconstruction Benchmark \cite{berger2013benchmark} which consists of simulated noisy range scans (points and normals) taken from 5 shapes with challenging properties such as complex topologies, sharp features, and small surface details. We evaluate our method against Deep Geometric Prior (DGP) \cite{Williams_2019}, Implicit Geometric Regularization (IGR) \cite{gropp2020implicit}, SIREN \cite{sitzmann2020implicit}, and Fourier Feature Networks (FFN) \cite{tancik2020fourier}. We remark that DGP establishes itself as superior to a dozen other classical methods on this benchmark and IGR furhter outperforms DGP. As in \cite{gropp2020implicit} and \cite{Williams_2019}, Table~\ref{tab:srb_results} reports the Hausdorff ($d_H$) and Chamfer ($d_C$) distances between the reconstruction and ground-truth. We also report the one sided Hausdorff ($d_{\vec{H}}$) and Chamfer ($d_{\vec{C}}$) distances between the scan and the reconstruction, which measures how much the reconstructions overfits noise in the input. Our reconstructions are quantitatively closer to ground truth on all but one model. Figure~\ref{fig:srb_main} shows visual examples of a few models from the benchmark. All models are shown in Appendix~\ref{sec:allfigs}. As with the Shapenet benchmark, we did parameter sweeps for those methods which have tunable parameters, choosing the best model for each metric under consideration. See Appendix~\ref{sec:experiment_details_srb} for details.
\begin{figure}
    \minipage{0.25\linewidth}
    \centering
    \includegraphics[width=0.97\linewidth]{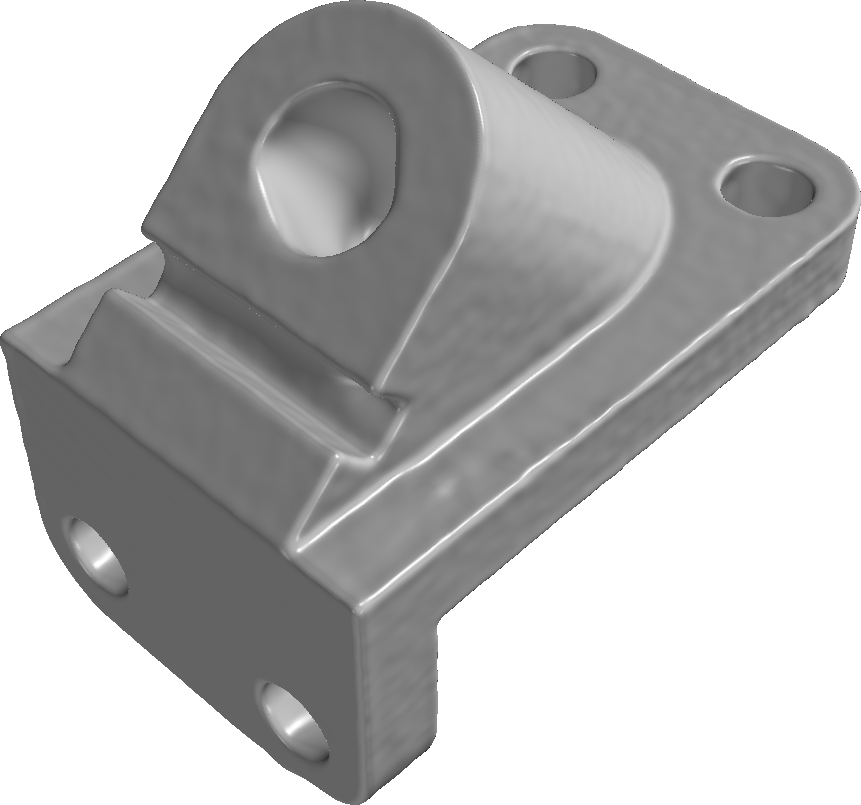}
    \endminipage\hfill
    \minipage{0.25\linewidth}
    \centering
    \includegraphics[width=0.97\linewidth]{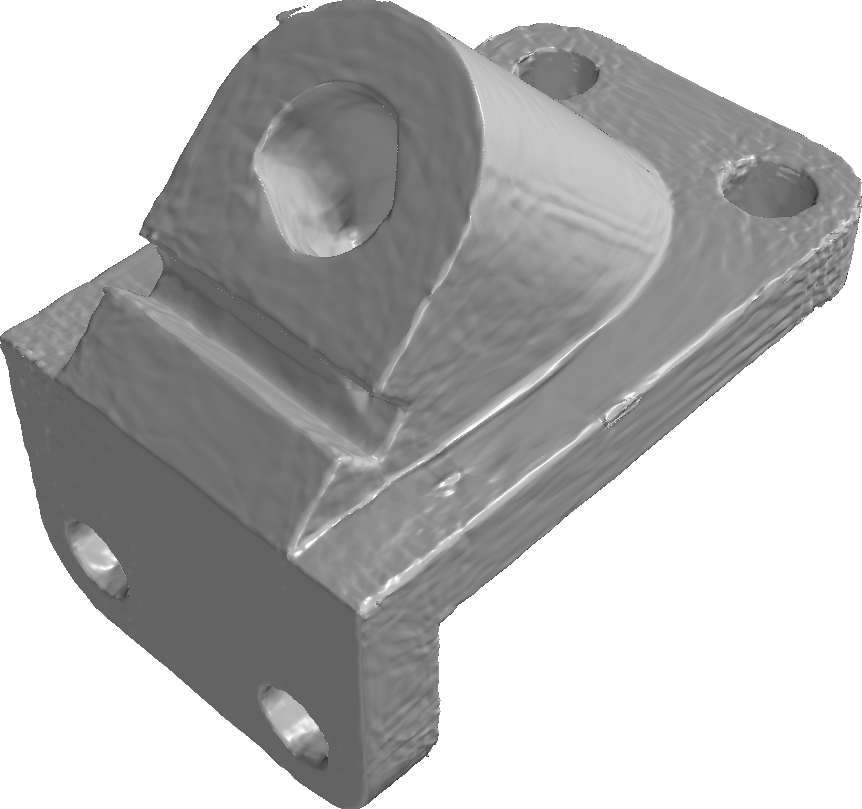}
    \endminipage\hfill
    \minipage{0.25\linewidth}
    \centering
    \includegraphics[width=0.97\linewidth]{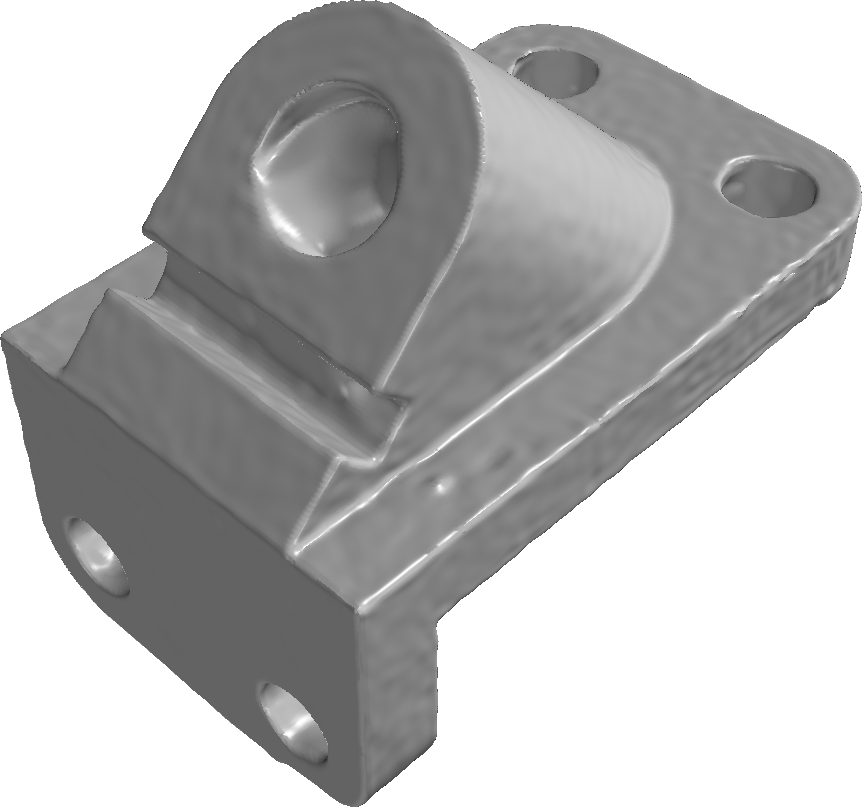}
    \endminipage\hfill
    \minipage{0.25\linewidth}
    \centering
    \includegraphics[width=0.97\linewidth]{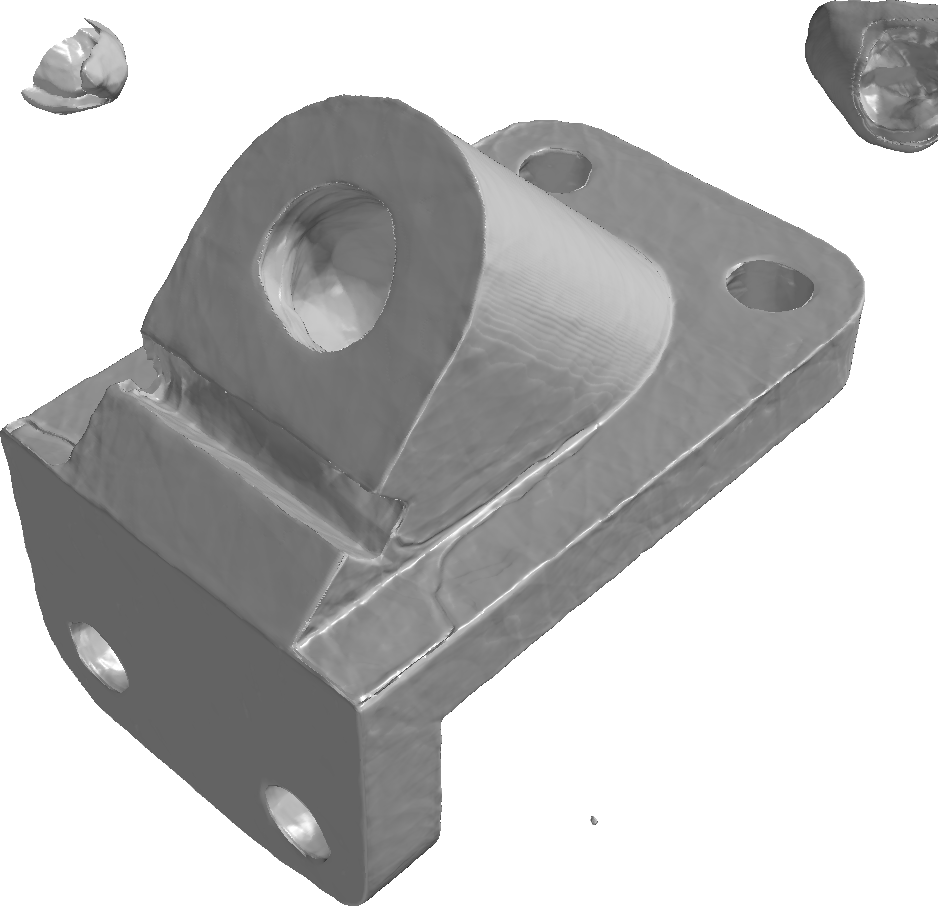}
    \endminipage\hfill
    \\
    \minipage{0.25\linewidth}
    \centering
    \includegraphics[width=0.97\linewidth]{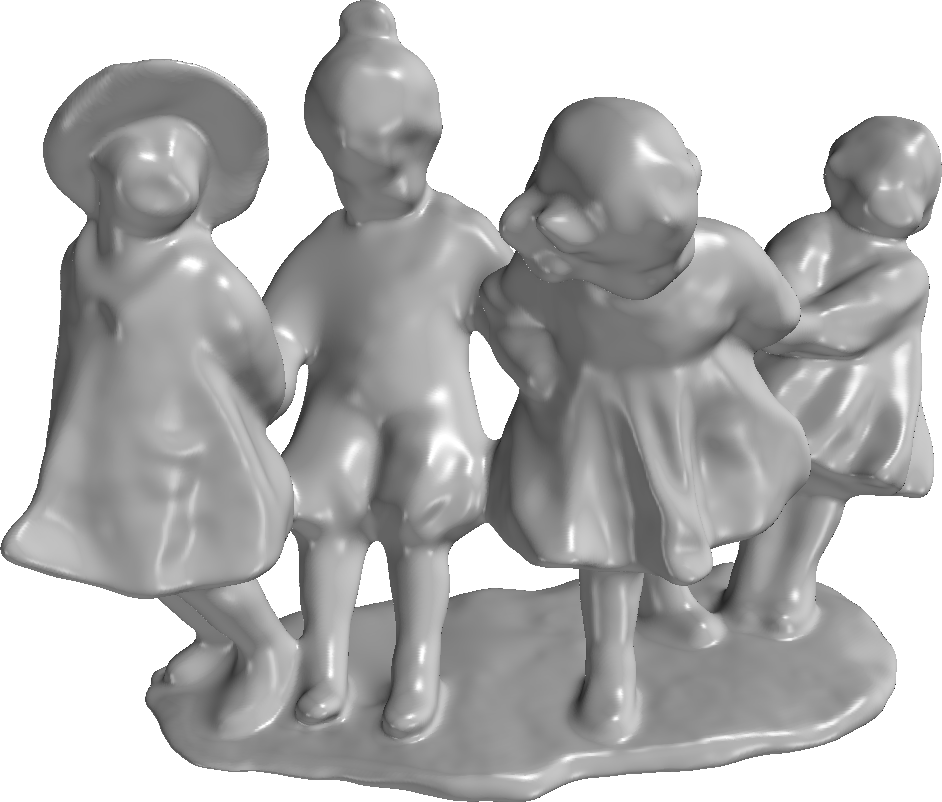}
    \endminipage\hfill
    \minipage{0.25\linewidth}
    \centering
    \includegraphics[width=0.97\linewidth]{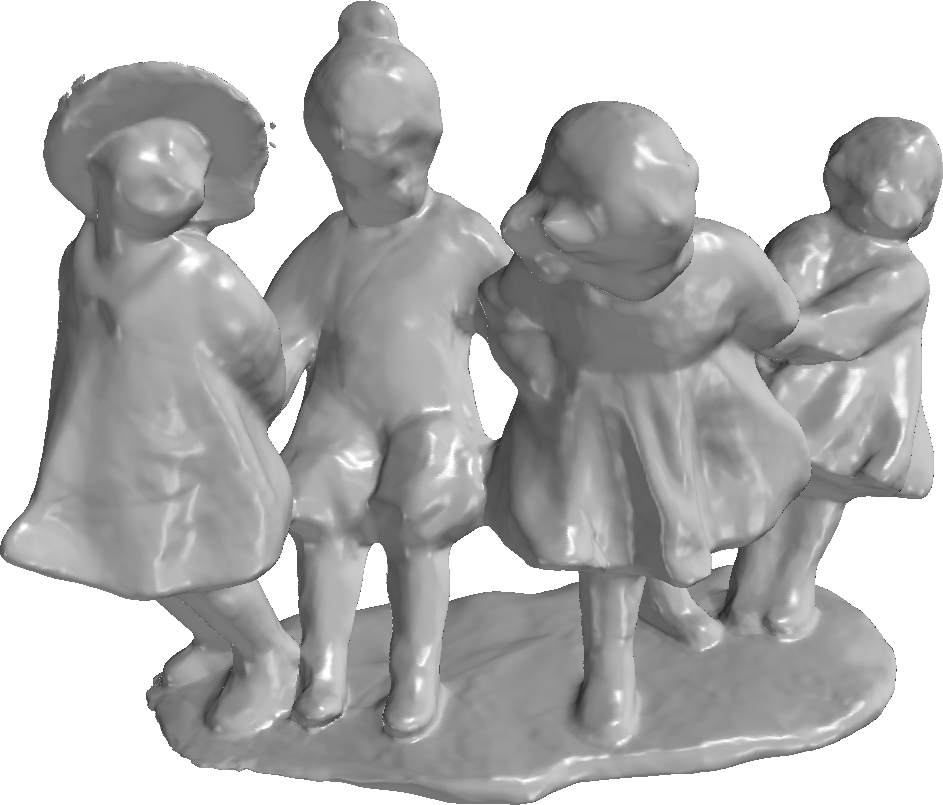}
    \endminipage\hfill
    \minipage{0.25\linewidth}
    \centering
    \includegraphics[width=0.97\linewidth]{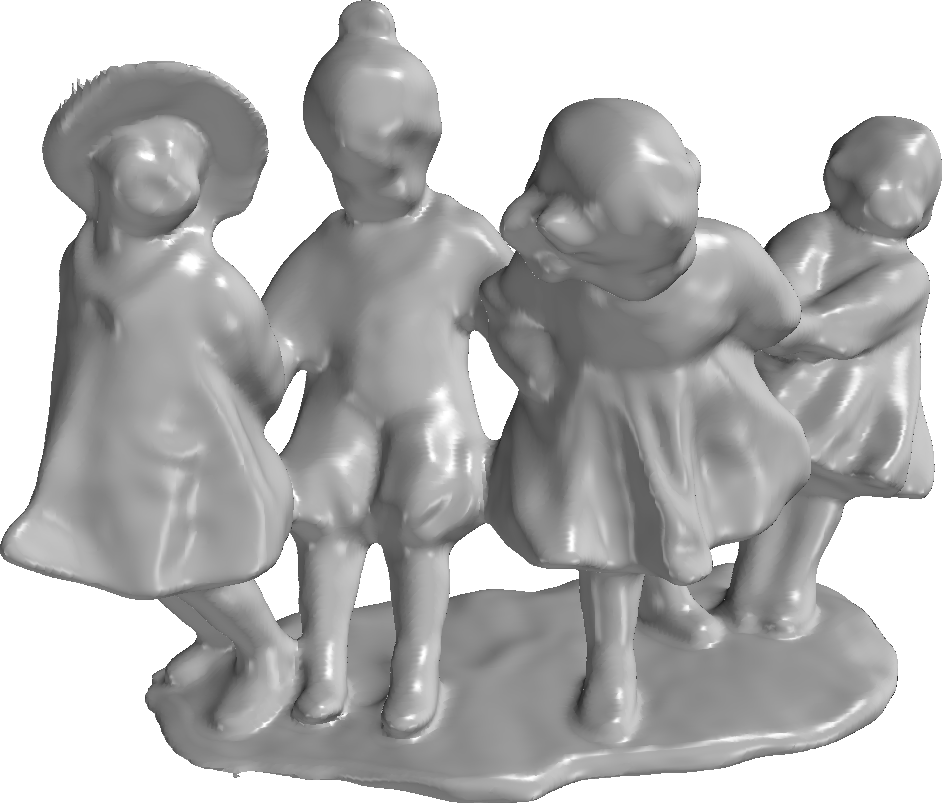}
    \endminipage\hfill
    \minipage{0.25\linewidth}
    \centering
    \includegraphics[width=0.97\linewidth]{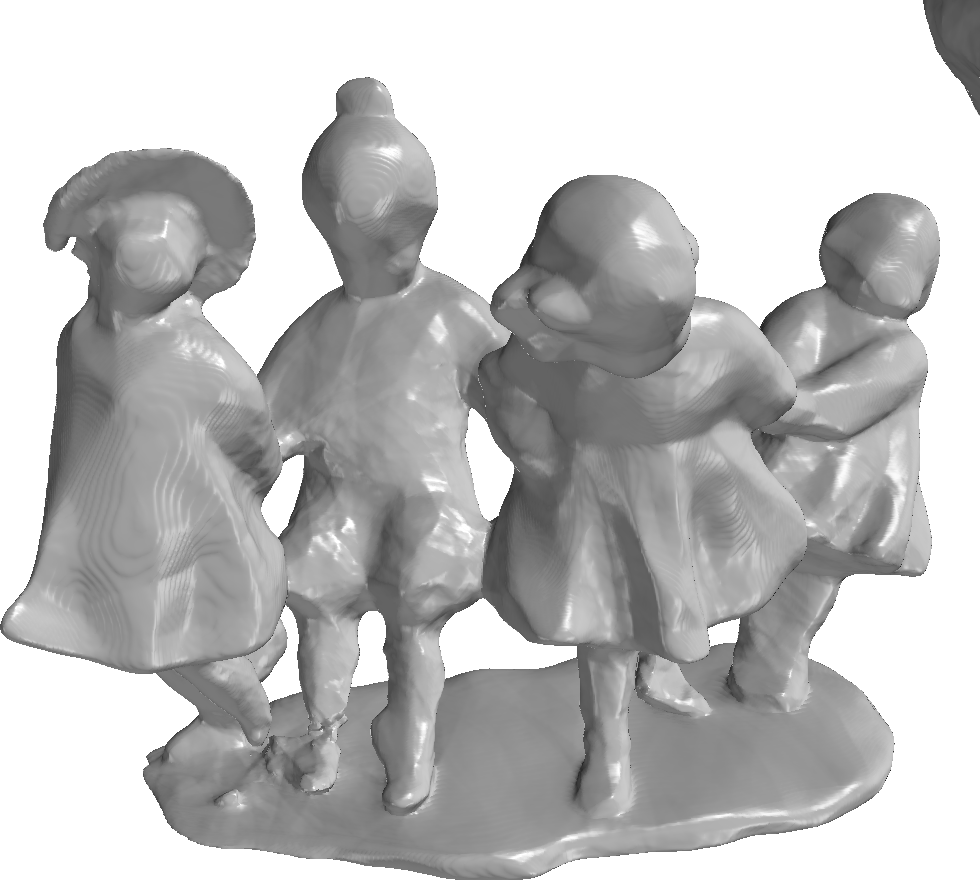}
    \endminipage\hfill
    \\
    \minipage{0.25\linewidth}
    \centering
    \includegraphics[width=0.97\linewidth]{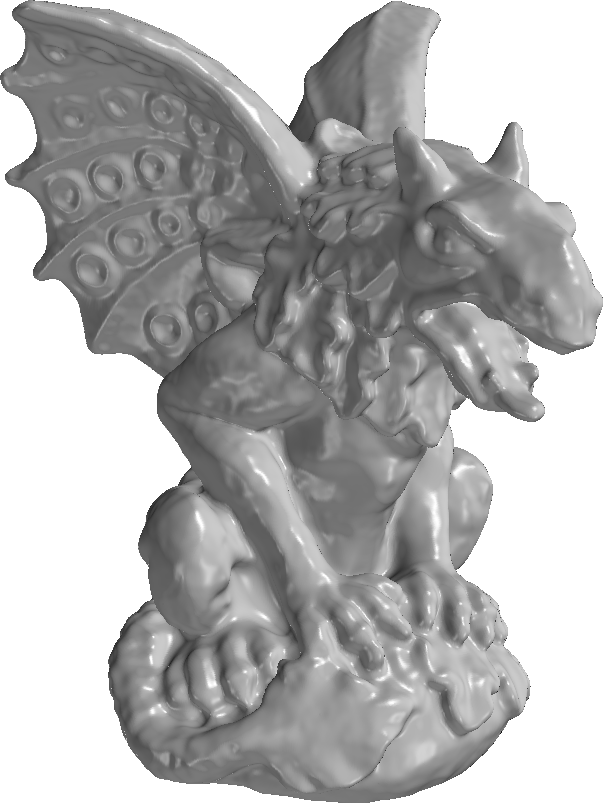}
    \endminipage\hfill
    \minipage{0.25\linewidth}
    \centering
    \includegraphics[width=0.97\linewidth]{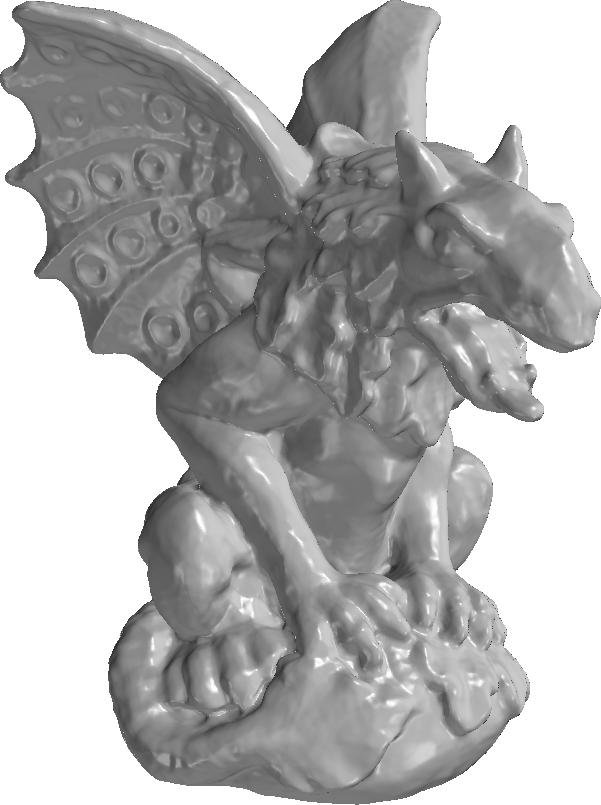}
    \endminipage\hfill
    \minipage{0.25\linewidth}
    \centering
    \includegraphics[width=0.97\linewidth]{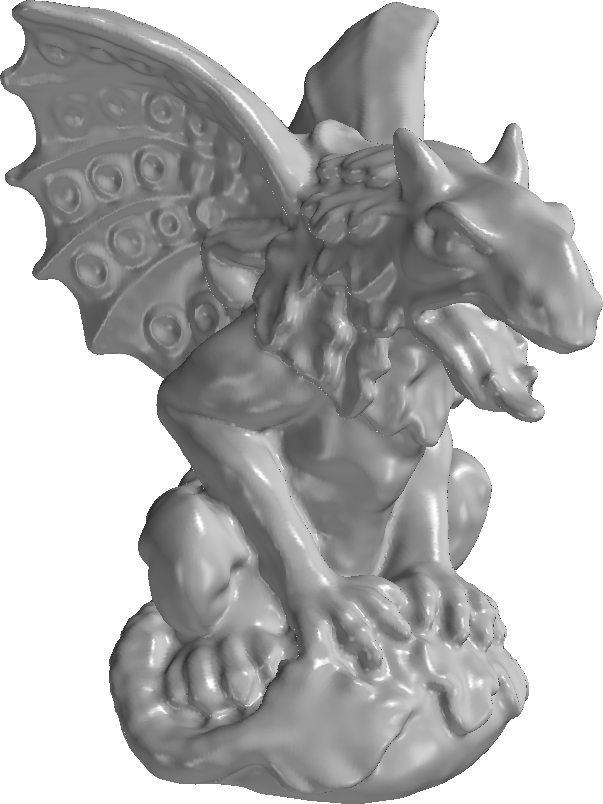}
    \endminipage\hfill
    \minipage{0.25\linewidth}
    \centering
    \includegraphics[width=0.97\linewidth]{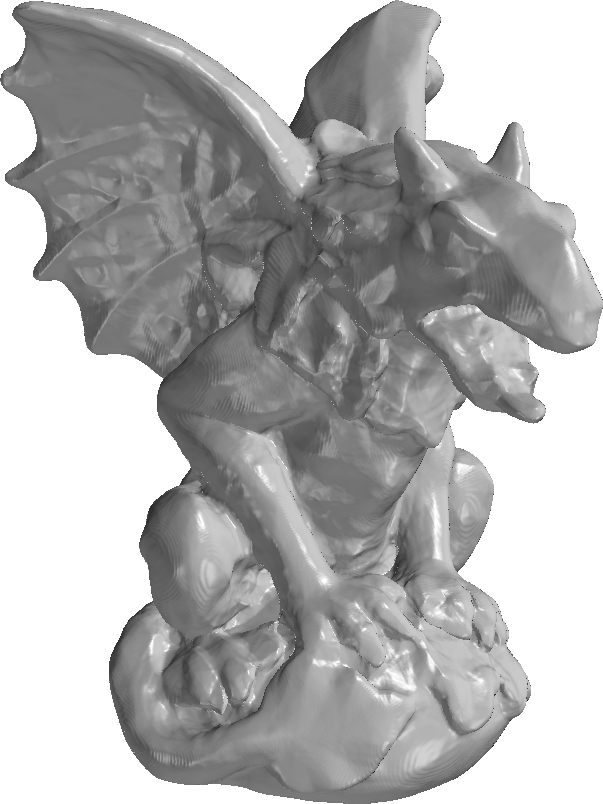}
    \endminipage\hfill
    \vspace{0.75em}
    \hfill
    \minipage{0.25\linewidth}
    \centering \footnotesize \textbf{Ours}
    \endminipage\hfill
    \minipage{0.25\linewidth}
    \centering \footnotesize IGR
    \endminipage\hfill
    \minipage{0.25\linewidth}
    \centering \footnotesize SIREN
    \endminipage\hfill
    \minipage{0.25\linewidth}
    \centering \footnotesize Fourier Feats
    \endminipage\hfill
    \vspace{0.5em}
    \caption{\emph{Comparisons between reconstruction techniques on the Surface Reconstruction Benchmark.} For techniques requiring parameter sweeps, we show the result with the lowest Chamfer Distance. See Appendix~\ref{sec:allfigs} for more figures from each model.}
    \label{fig:srb_main}
\end{figure}

\begin{table}
    \begin{scriptsize}
    \begin{center}
    \begin{tabular}{c | c | c | c | c | c}
        \multicolumn{2}{c}{} & \multicolumn{2}{|c|}{Ground Truth} &  \multicolumn{2}{c}{Scans} \\
        
                                   & Method                              & $d_C$          & $d_H$           & $d_{\vec{C}}$ & $d_{\vec{H}}$\\
        \hline
        
        \multirow{5}{*}{Anchor}    & DGP \cite{Williams_2019}            & 0.33           & 8.82           & 0.08  & 2.79 \\
                                   & IGR \cite{gropp2020implicit}        & \textbf{0.22}  & 4.71           & 0.12  & 1.32 \\
                                   & SIREN \cite{sitzmann2020implicit}   & 0.32           & 8.19           & 0.11  & 2.43 \\
                                   & FFN \cite{tancik2020fourier}        & 0.31           & 4.49           & 0.10  & 0.10 \\
                                   & \textbf{Ours}                       & \textbf{0.22}  & \textbf{4.65}  & 0.11  & 1.11 \\
        \hline
        
        \multirow{5}{*}{Daratech}  & DGP \cite{Williams_2019}            & \textbf{0.2}   & 3.14           & 0.04  & 1.89\\
                                   & IGR \cite{gropp2020implicit}        & 0.25           & 4.01           & 0.08  & 1.59\\
                                   & SIREN \cite{sitzmann2020implicit}   & 0.21           & 4.30           & 0.09  & 1.77\\
                                   & FFN \cite{tancik2020fourier}        & 0.34           & 5.97           & 0.10  & 0.10\\
                                   & \textbf{Ours}                       & 0.21           & 4.35           & 0.08  & 1.14\\
        \hline

        \multirow{5}{*}{DC}        & DGP \cite{Williams_2019}            & 0.18           & 4.31           & 0.04  & 2.53\\
                                   & IGR \cite{gropp2020implicit}        & 0.17           & 2.22           & 0.09  & 2.61\\
                                   & SIREN \cite{sitzmann2020implicit}   & 0.15           & 2.18           & 0.06  & 2.76\\
                                   & FFN \cite{tancik2020fourier}        & 0.20           & 2.87           & 0.10  & 0.12\\
                                   & \textbf{Ours}                       & \textbf{0.14}  & \textbf{1.35}  & 0.06  & 2.75\\
        \hline
        
        \multirow{5}{*}{Gargoyle}  & DGP \cite{Williams_2019}            & 0.21           & 5.98           & 0.06  & 3.41\\
                                   & IGR \cite{gropp2020implicit}        & \textbf{0.16}  & 3.52           & 0.06  & 0.81\\
                                   & SIREN \cite{sitzmann2020implicit}   & 0.17           & 4.64           & 0.08  & 0.91\\
                                   & FFN \cite{tancik2020fourier}        & 0.22           & 5.04           & 0.09  & 0.09\\
                                   & \textbf{Ours}                       & \textbf{0.16}  & \textbf{3.20}  & 0.08  & 2.75\\
        \hline
        
        \multirow{5}{*}{Lord Quas} & DGP \cite{Williams_2019}            & 0.14          & 3.67           & 0.04  & 2.03\\
                                   & IGR \cite{gropp2020implicit}        & \textbf{0.12} & 1.17           & 0.07  & 0.98\\
                                   & SIREN \cite{sitzmann2020implicit}   & 0.17          & 0.82           & 0.12  & 0.76\\
                                   & FFN \cite{tancik2020fourier}        & 0.35          & 3.90           & 0.06  & 0.06\\
                                   & \textbf{Ours}                       & \textbf{0.12} & \textbf{0.69}  & 0.05  & 0.62\\
    \end{tabular}
    \end{center}
    \end{scriptsize}
    \caption{Quantitative results on the Surface Reconstruction Benchmark: The \emph{Ground Truth} contains the Chamfer ($d_C$) and Hausdorff ($d_H$) distances between the reconstruction and ground truth. The \emph{Scans} column contains the one sided Chamfer ($d_{\vec{C}}$) and Hausdorff  ($d_{\vec{H}}$) distance between the reconstruction and noisy inputs which measures how much the reconstruction overfits noise in the input.}
    \label{tab:srb_results}
\end{table}

\vspace{\titlepre}
\subsection{Large Scale Reconstruction of Full Scenes} 
\begin{figure*}
    \centering
    \includegraphics[width=0.97\linewidth]{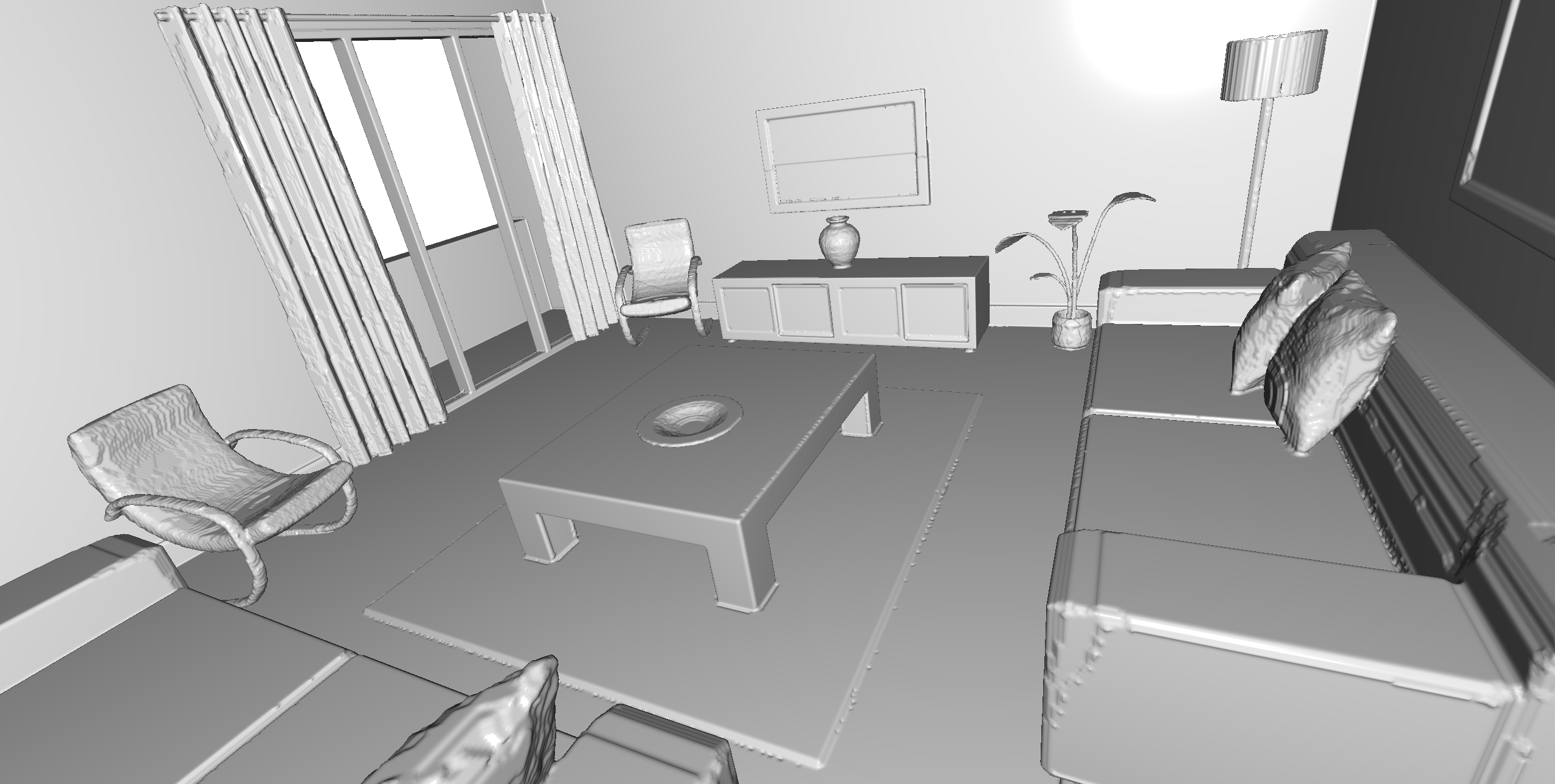}
    \caption{Using our method to reconstruct a full scene consisting of 9 million points.}
    \label{fig:living_room}
\end{figure*}

\vspace{\titlepost}
Figure~\ref{fig:living_room} shows a full scene consisting of 9 million points reconstructed using our method. This is the same scene used in~\cite{sitzmann2020implicit}, and contains many thin features that are difficult to reconstruct (\eg, curtains, plant, and lampshade). To generate the input point cloud for this experiment, we densely sampled a mesh extracted from a $2048^3$ occupancy grid of the scene. To perform the reconstruction, we subdivided the space into 8x8x8 cells containing between 10k and 500k samples and reconstructed each cell interdependently using up to 15k Nystr\"{o}m samples. The whole process takes 1.5 hours on a machine with an NVIDIA 1080Ti GPU.

\vspace{\titlepre}
\subsection{Timing and Performance}\label{sec:performance}
\vspace{\titlepost}
Table~\ref{tbl:timings} compares the average running time and GPU memory usage of our method and others when reconstructing point clouds from the Surface Reconstruction Benchmark described in Section~\ref{sec:srb_comp}. To ensure a fair comparison we ran all neural network based methods for 5000 iterations. We do not report exact CPU memory usage for the methods since it is hard to measure but we remark that, by observation, system memory usage never exceeded 3GiB for any of the methods. Appendix~\ref{sec:full_perf} reports the running times and memory usages for individual models in the benchmark.

\begin{table}
\begin{center}
\begin{scriptsize}
    \begin{tabular}{c|c|c|c|c|c}
                      & FFN     & IGR      & SIREN    & Poisson  & Ours \\
        \hline
        Runtime (sec)  & 337.71  & 1413.94  & 599.00   & 1.64     & 11.91 \\
        Max VRAM (MiB) & 3711    & 5093     & 1607     & N.A.     & 5285
    \end{tabular}
\end{scriptsize}
\end{center}
    \caption{Average runtime and GPU memory utilization of various methods on the Surface Reconstion Benchmark \cite{berger2013benchmark} models. All Neural Network based methods were run for 5000 iterations. }
    \label{tbl:timings}
\end{table}

\vspace{\titlepre}
\subsection{Our Method versus Neural Networks} 
\vspace{\titlepost}
\begin{figure}
    \minipage{0.25\linewidth}
    \centering
    \includegraphics[width=0.97\linewidth]{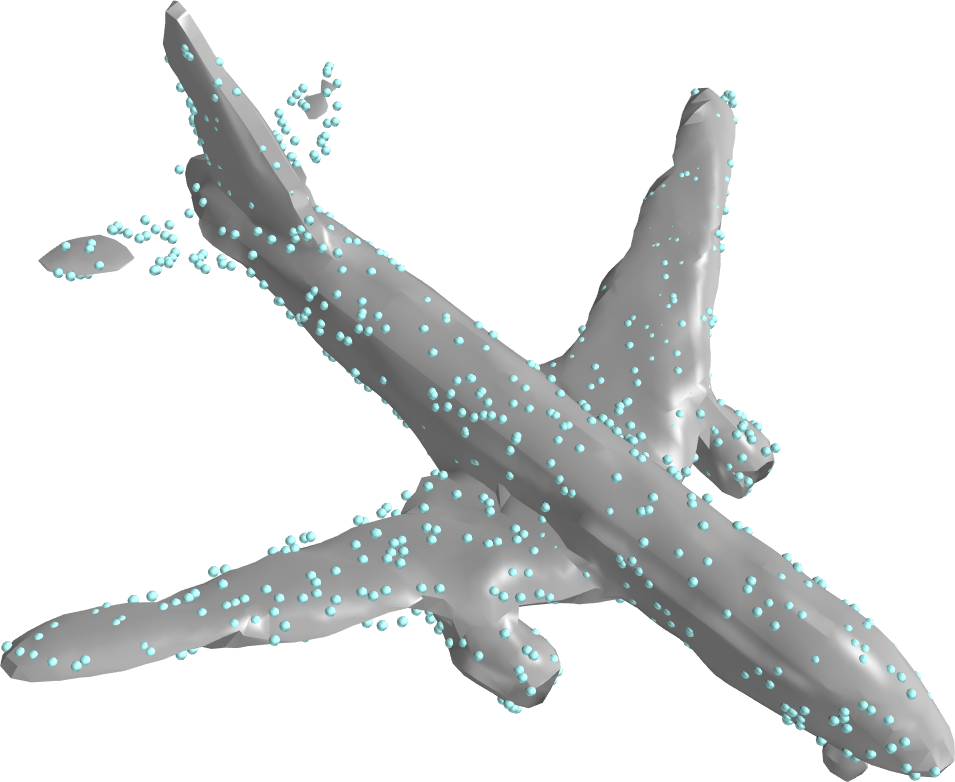}
    \endminipage
    \minipage{0.25\linewidth}
    \centering
    \includegraphics[width=0.97\linewidth]{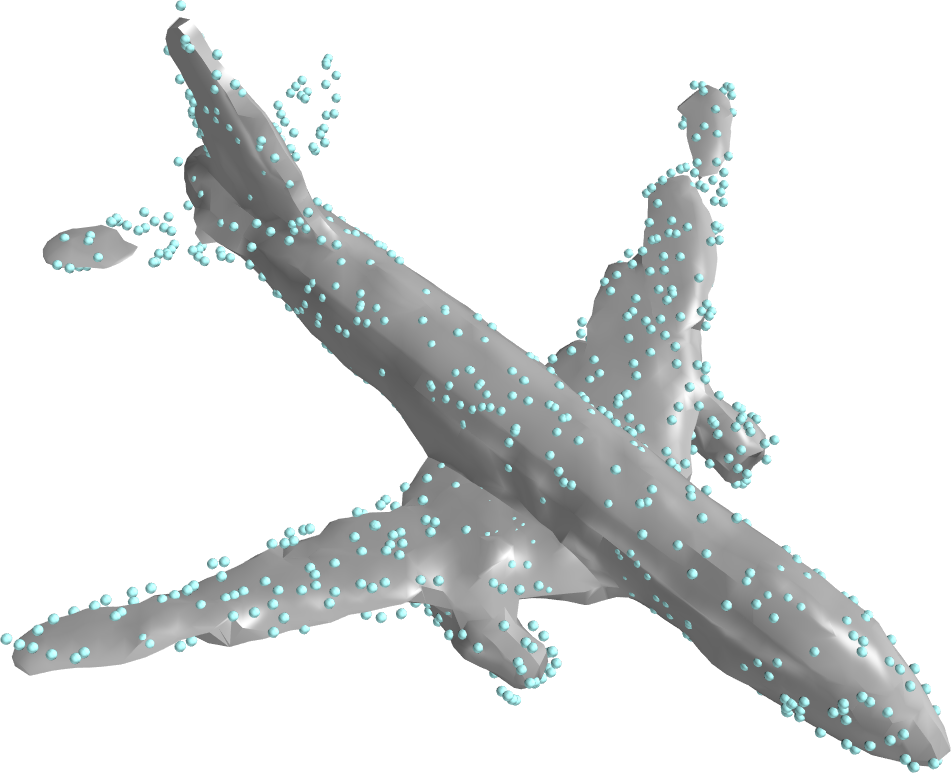}
    \endminipage
    \minipage{0.25\linewidth}
    \centering
    \includegraphics[width=0.97\linewidth]{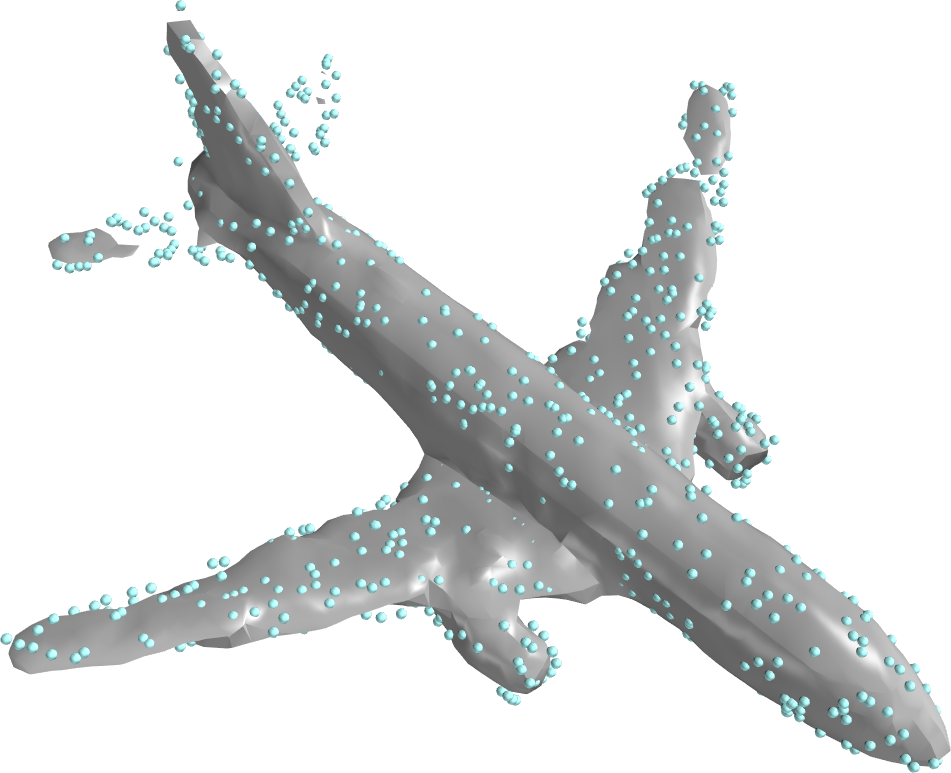}
    \endminipage
    \minipage{0.25\linewidth}
    \centering
    \includegraphics[width=0.97\linewidth]{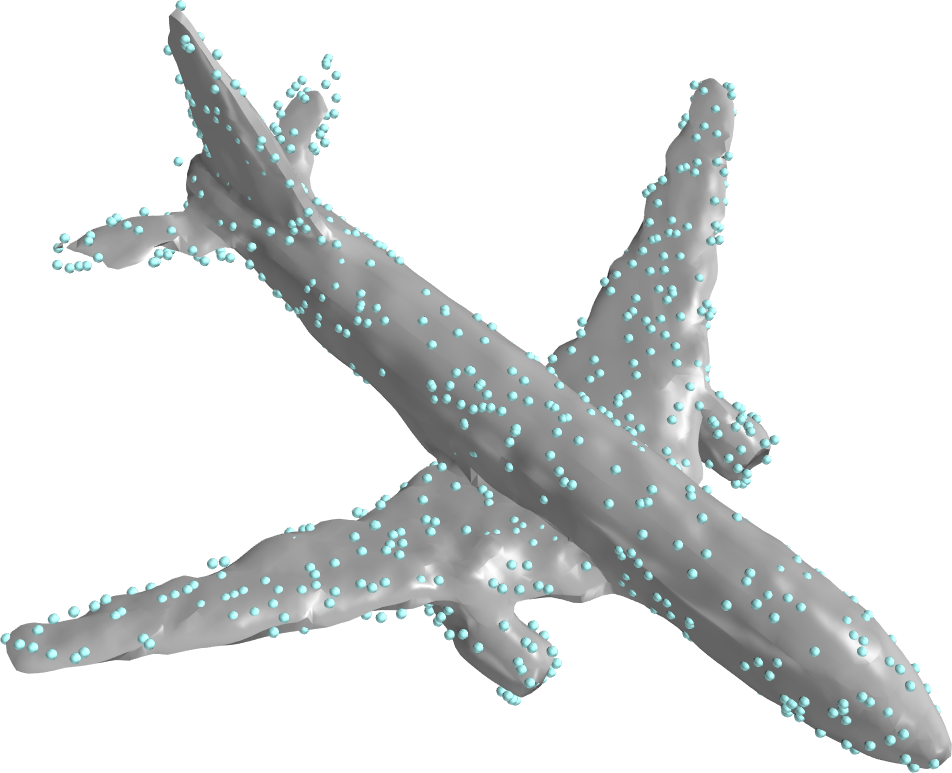}
    \endminipage\hfill
    \vspace{0.5em}
    \minipage{0.25\linewidth}
    \centering
    \centering \footnotesize 5k Iter
    \endminipage
    \minipage{0.25\linewidth}
    \centering
    \centering \footnotesize 50k Iter
    \endminipage
    \minipage{0.25\linewidth}
    \centering
    \centering \footnotesize 75k Iter
    \endminipage
    \minipage{0.25\linewidth}
    \centering
    \centering \footnotesize 100k Iter
    \endminipage\hfill
    \vspace{0.5em}
    \caption{Implicit Geometric Regularization \cite{gropp2020implicit} and other methods based on ReLU networks  suffer from slow convergence for sharp and thin features (\eg, the tail of the airplane only begins to appear after 100k iterations).}
    \label{fig:igr_slow_convergence}
\end{figure}
\begin{figure}
    \minipage{0.33\linewidth}
    \centering
    \includegraphics[width=0.97\linewidth]{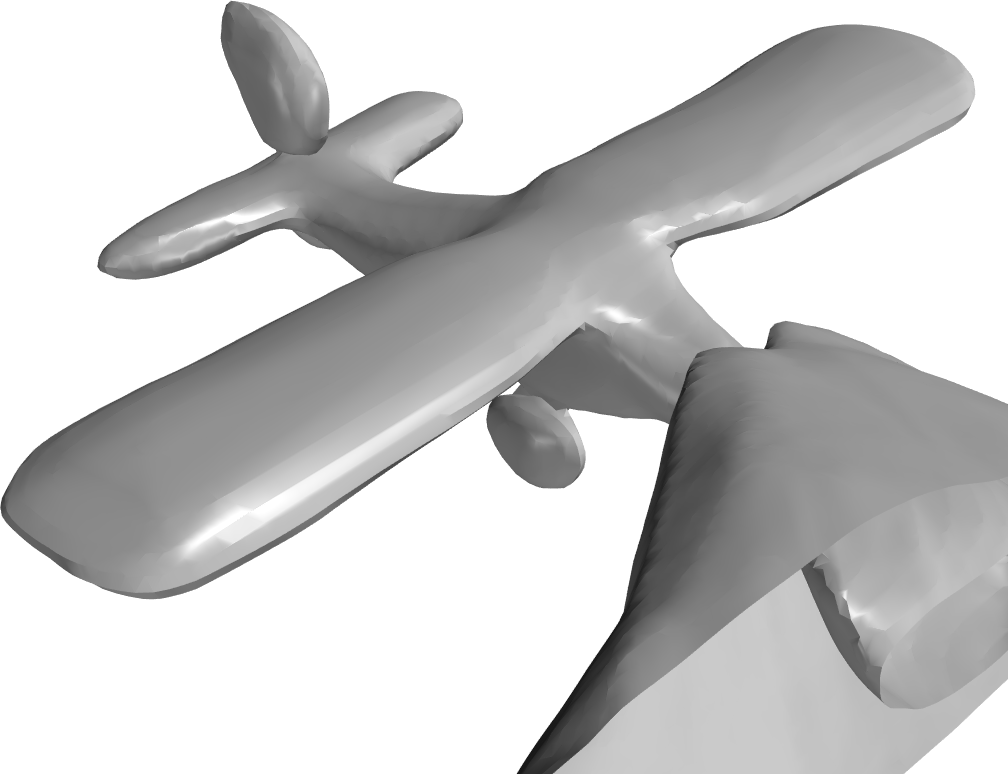}
    \endminipage
    \minipage{0.33\linewidth}
    \centering
    \includegraphics[width=0.97\linewidth]{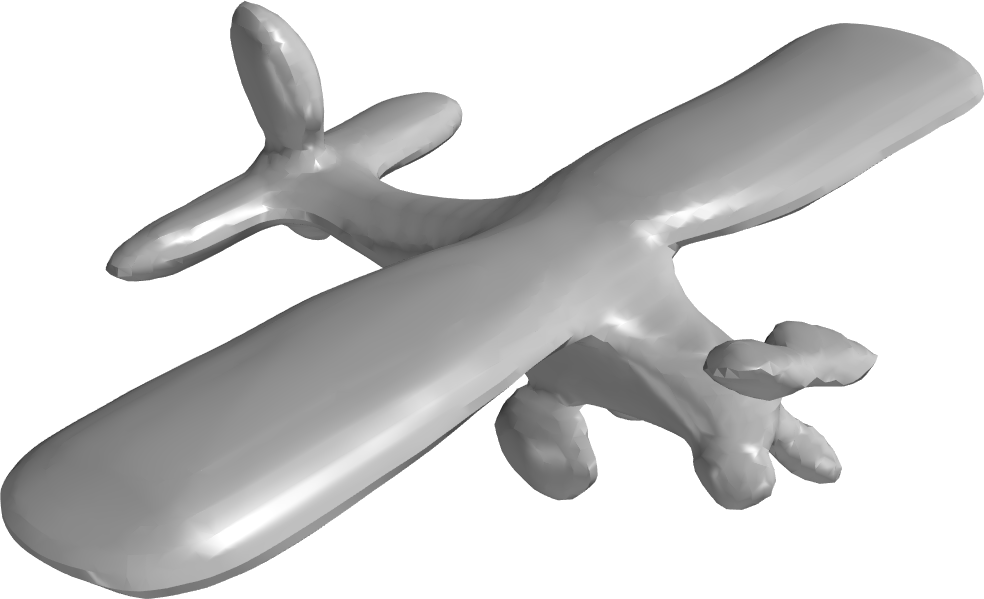}
    \endminipage
    \minipage{0.33\linewidth}
    \centering
    \includegraphics[width=0.97\linewidth]{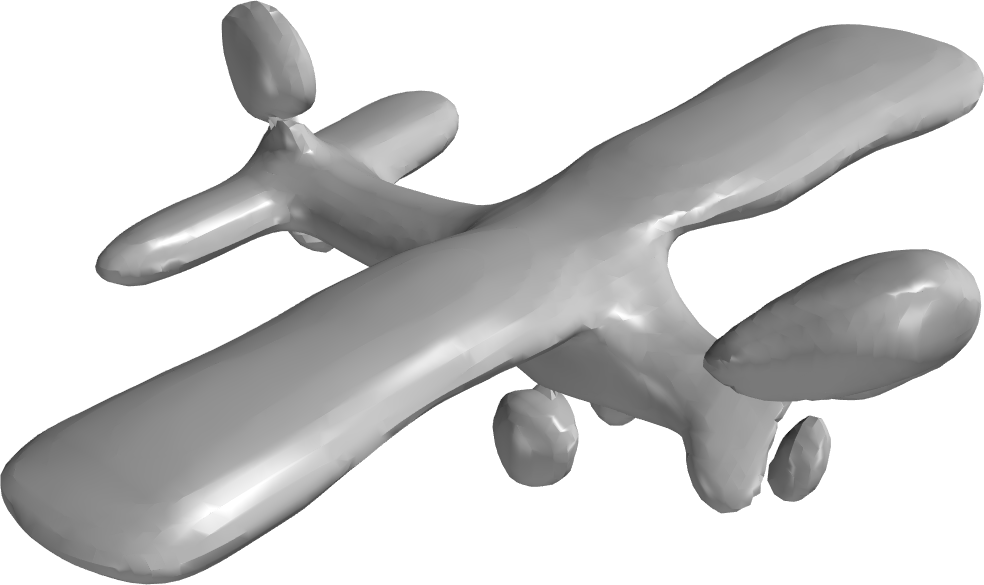}
    \endminipage\hfill
    \vspace{0.5em}
    \minipage{0.33\linewidth}
    \centering
    \centering \footnotesize $\sigma = 0.5$
    \endminipage
    \minipage{0.33\linewidth}
    \centering
    \centering \footnotesize $\sigma = 0.6$
    \endminipage
    \minipage{0.33\linewidth}
    \centering
    \centering \footnotesize  $\sigma = 0.7$
    \endminipage\hfill
    \vspace{0.5em}
    \caption{Fourier Feature Networks with Gaussian features \cite{tancik2020fourier} are sensitive to the choice of variance ($\sigma$) of the Gaussian when data is sparse. The images above show the result of training a network with the same initialization and Fourier features with different initial scales for the Gaussian distribution. Small changes in $\sigma$ lead to large changes in the final reconstruction.}
    \label{fig:ffn_sensitive}
\end{figure}
Implicit Geometric Regularization \cite{gropp2020implicit} demonstrates that ReLU networks have a natural inductive bias making them good at reconstructing shapes. However, methods based on ReLU networks suffer from slow convergence (see Figure~\ref{fig:igr_slow_convergence}). SIREN \cite{sitzmann2020implicit} drastically improves convergence speed by replacing ReLU with sinusoidal activations and a clever initialization, however, SIREN occasionally underfits on sparse inputs and requires an additional loss computed on points in the volume around the shape to prevent reconstruction artefacts. This loss increase runtimes, and suggest that SIREN's inductive bias may not be ideal for sparse reconstruction tasks. By projecting input points onto random Fourier features, Fourier Feature Networks (FFNs) \cite{tancik2020fourier} present a principled approach rooted in Kernel methods to control the inductive bias of the solution as well as convergence speed. While FFNs are capable of producing high quality solutions, they are sensitive to the choice of feature distribution when data is sparse (see Figure~\ref{fig:ffn_sensitive}) and thus require tuning to work well. In contrast, our technique converges in seconds (Section~\ref{sec:performance}), has a well suited inductive bias for shape representation (Section~\ref{sec:implicit_bias}), and requires minimal parameter tuning.

\vspace{\titlepre}
\section{Conclusion and Future Work}
\vspace{\titlepost}
We have shown that Neural Spline kernels arising from infinitely wide shallow ReLU networks are very effective tools for 3D surface reconstruction, outperforming state-of-the-art methods while being computationally efficient and conceptually simple. In a sense, our work bridges the gap between traditional reconstruction methods and modern neural network based methods by leveraging the deep connection between neural networks and kernels. 

We remark that our kernel formulation is fully differentiable. In the future, we hope to integrate Neural Splines into deep learning pipelines and apply them to other 3D tasks such as shape completion and sparse reconstruction. On the theory side, we would like to %
investigate and compare the approximation properties of different kernels (in particular those arising from infinite width sinusoidal networks) in the context of 3D reconstruction. %

\vspace{-1.5em}
\paragraph{Acknowledgements}This work is partially supported by the Alfred P. Sloan Foundation, NSF RI-1816753, NSF CAREER CIF 1845360, NSF CHS-1901091, and Samsung Electronics.
\newpage
{\small
\bibliographystyle{ieee_fullname}
\bibliography{main.bib}
}

\newpage
\appendix
\onecolumn

\section{Additional Experiments}

\subsection{Detailed description of Shapenet Experiment}\label{sec:experiment_details_shapenet}
\vspace{-0.35em}
Many of the methods we compared against on Shapenet have tunable parameters which can drastically alter the quality of reconstructed outputs. To ensure a proper comparison, we ran sweeps over these parameters where appropriate choosing the best reconstruction for each model under both metrics (Chamfer and IoU). For our method we did no parameter sweeps, using no regularization and 1024 Nystr\"{o}m samples for each model in the dataset. We describe the exerimental methodology for each method in the benchmark below.

\vspace{-0.95em}

\paragraph{Implicit Geometric Regularization \cite{gropp2020implicit}} We trained each model for 5k iterations with Adam and a learning rate of 0.001 using the same parameters and architecture as proposed in the original paper. We included the normals in the loss with the parameter $\tau$ set to 1. The Eikonal regularization term $\lambda$ was set to 0.1. While IGR can slightly improve by using a very large number of iterations. doing so is prohibitively slow over many models. Figure~\ref{fig:igr_slow_convergence} motivates our choice of iterations, demonstrating only a slight improvements between 5k and 100k iterations (the latter which required 2 hours of fitting on a NVIDIA-1080-Ti GPU).

\vspace{-0.95em}

\paragraph{Screened Poisson Surface Reconstruction \cite{kazhdan2013screened}} We considered every possible combination of the following parameters: the \emph{octree depth} in $[6, 7, 8, 9]$, the number of \emph{points per leaf} in $[1, 2, 3, 5, 10]$, the \emph{point weight} (which controls the degree to which the method interpolates the input) in $[4.0, 100.0, 1000.0]$. Since all the shapes in the benchmark are watertight meshes, we used Dirichlet boundary constraints for the reconstruction.

\vspace{-0.95em}

\paragraph{SIREN \cite{sitzmann2020implicit}} We trained used a SIREN network with 5 hidden layers each containing 256 neurons for 5000 iterations, using a learning rate of 1e-4 with Adam. We used the same loss for shapes as in the original SIREN paper, sampling an additional $N$ (where $N$ is the number of input points) points in an axis aligned bounding box whose diagonal is 10\% larger than the object's bounding box. We minimize the same loss for shapes as the SIREN paper (see section 4.2 in the original SIREN paper).

\vspace{-0.95em}

\paragraph{Fourier Feature Networks \cite{tancik2020fourier}} We used an 8-layer ReLU MLP with 256 Fourier features sampled from a Gaussian distribution. This is the same architecture as the shape representation experiment in the original paper. For each model in the benchmark, we did a parameter sweep on the variance $\sigma$ of the Gaussian distribution, considering $\sigma \in \{0.1, 0.25, 0.5. 0.6, 0.7, 0.8, 0.9, 1.0, 1.25, 1.5, 3.0\}$. The range of parameters was chosen by empirical verification on 3 models from the airplanes, benches, and cars categories. 

\vspace{-0.95em}

\paragraph{SVR \cite{NIPS2004_2724}} As in the original paper, we use a Gaussian kernel to perform support vector regression. To generate occupancy samples, we augmented the input points with an ''inside`` and ''outside`` point by perturbing them by $\pm \epsilon$ along the normal at that point. We used $\epsilon=0.01$ forall the models. For each model we did a joint parameter sweep over the regularization parameter $C \in \{1.0, 0.1, 0.01, 0.001, 0.0001\}$ and the variance parameter $\sigma \in \{0.002. 0.001, 0.0004, 0.0002, 0.0001\}$. The range of parameters was chosen by empirical verification on 3 models from the airplanes, benches, and cars categories. 

\vspace{-0.75em}

\paragraph{Biharmonic RBF \cite{carr2001reconstruction}} To generate occupancy samples, we augmented the input points with an ''inside`` and ''outside`` point by perturbing them by $\pm\ \epsilon$ along the normal at that point. We used $\epsilon=0.01$ for all the models. The biharmonic function is very simple $\phi(r) = r$ where $r = \|x_i - x_j\|$ and does require tuning parameters.

\subsection{Detailed Description of Surface Reconstrucion Benchmark Experiment}\label{sec:experiment_details_srb}
\vspace{-0.35em}
For the surface reconstruction benchmark we only included comparisons against neural network based methods since \cite{Williams_2019} performed an extensive comparison against traditional methods and clearly established itself as superior. We used the same experimental setup and parameter sweeps as in the Shapenet benchmark (Section~\ref{sec:experiment_details_shapenet}) for IGR, SIREN and Fourier feature networks. We verified that the range of parameters used for Fourier feature networks was valid by qualitative verification on the models (there are only 5 so this is straightforward). For our method we use 15000 Nystr\"{o}m samples. Additionally, since noise is present in the input data, we performed a modest parameter sweep over regularization parameters considering $\lambda \in \{0.0, 1\mathrm{e}{-13}, 1\mathrm{e}{-12}, 1\mathrm{e}{-11}, 1\mathrm{e}{-10}\}$. We remark that a full parameter sweep with our method requires less time than fitting a single model with competing methods (See timings in Section~\ref{sec:full_perf}).

\subsection{Additional Figures}\label{sec:allfigs}
\begin{figure}[H]
    \minipage{0.1427\linewidth}
    \centering
    \includegraphics[width=0.97\linewidth]{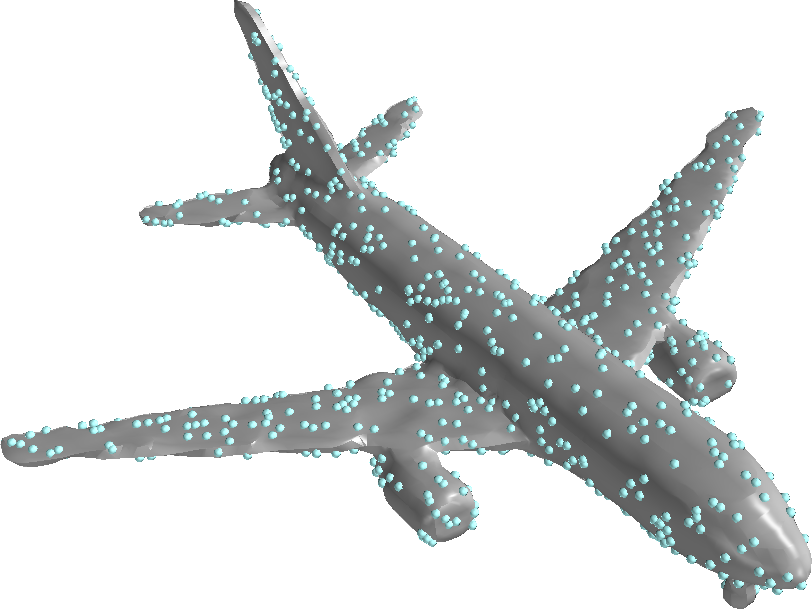}
    \endminipage\hfill
    \minipage{0.1427\linewidth}
    \centering
    \includegraphics[width=0.97\linewidth]{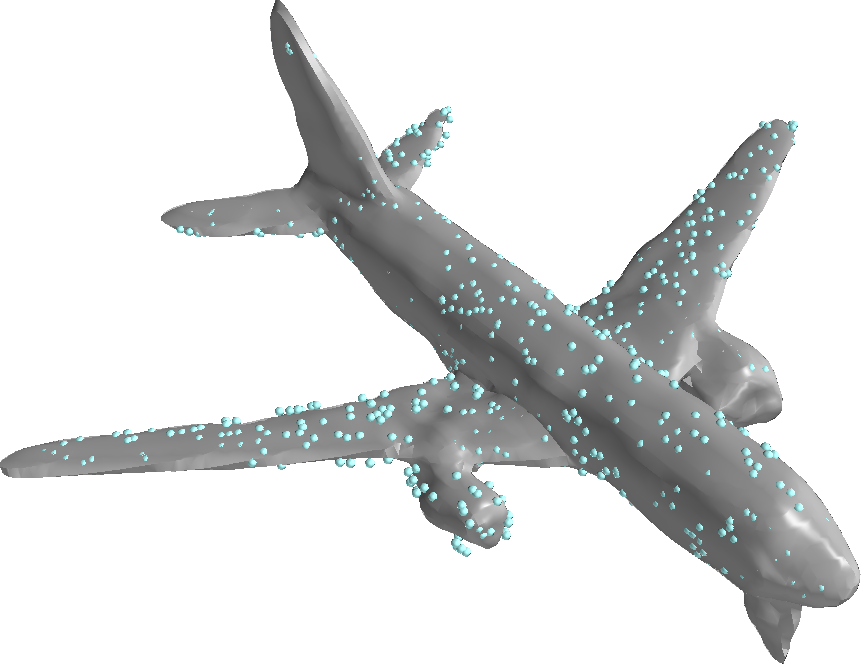}
    \endminipage\hfill
    \minipage{0.1427\linewidth}
    \centering
    \includegraphics[width=0.97\linewidth]{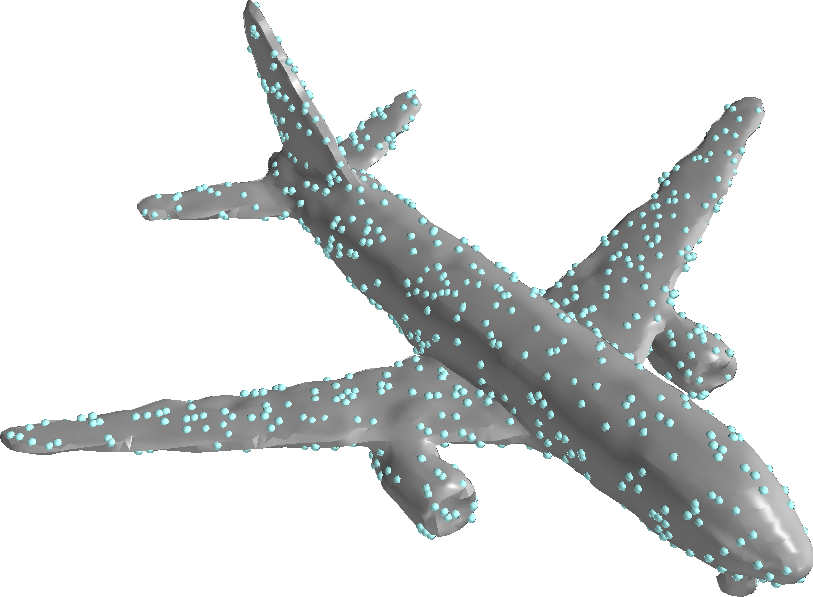}
    \endminipage\hfill
    \minipage{0.1427\linewidth}
    \centering
    \includegraphics[width=0.97\linewidth]{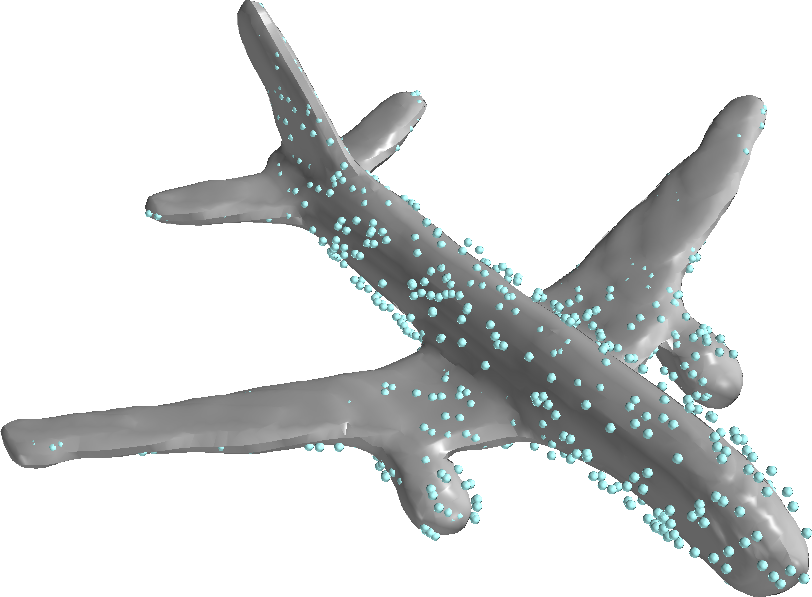}
    \endminipage\hfill
    \minipage{0.1427\linewidth}
    \centering
    \includegraphics[width=0.97\linewidth]{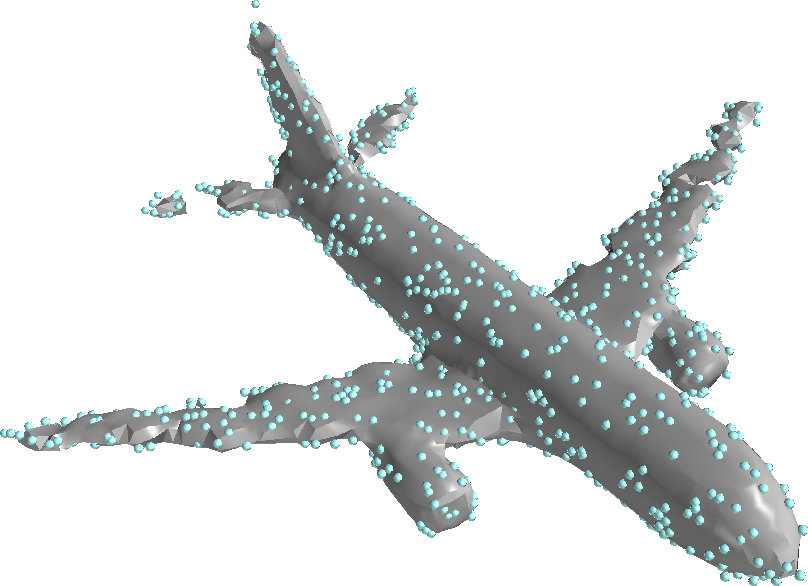}
    \endminipage\hfill
    \minipage{0.1427\linewidth}
    \centering
    \includegraphics[width=0.97\linewidth]{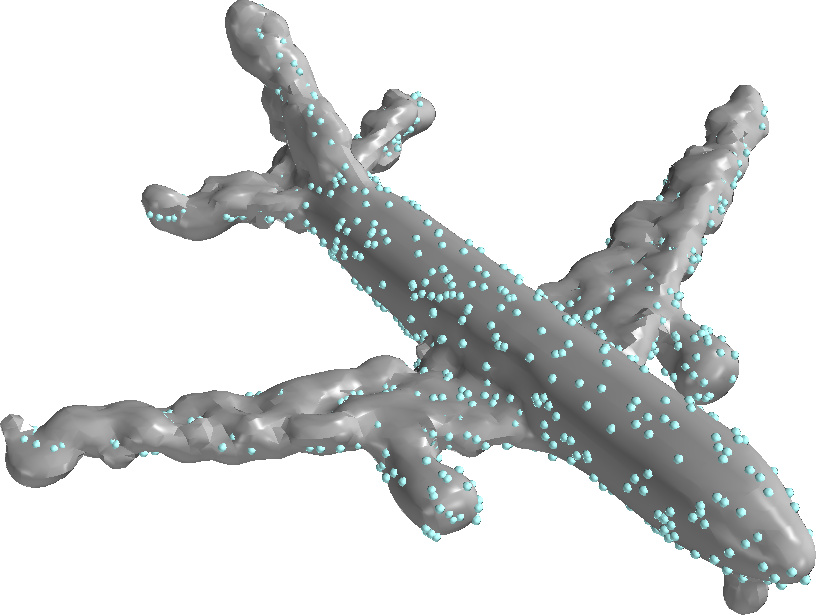}
    \endminipage\hfill
    \minipage{0.1427\linewidth}
    \centering
    \includegraphics[width=0.97\linewidth]{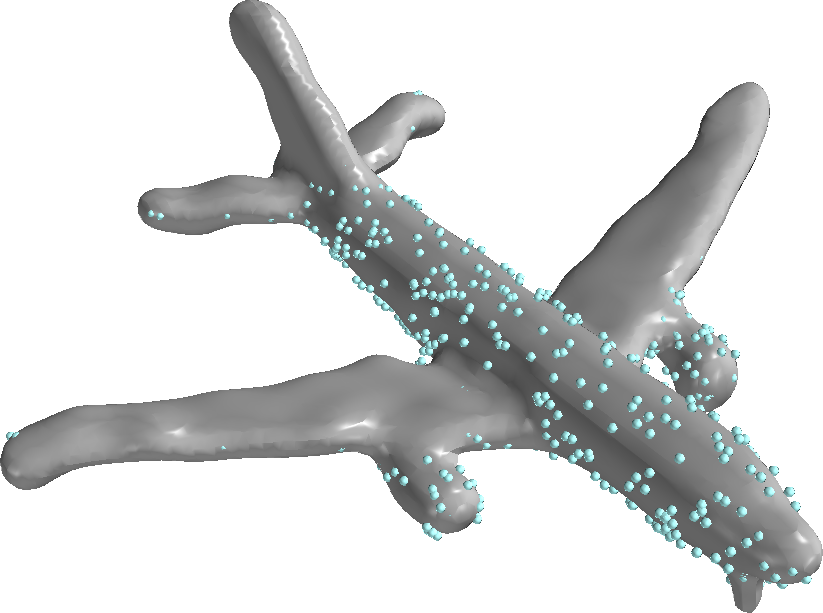}
    \endminipage\hfill
 \\
    \minipage{0.1427\linewidth}
    \centering
    \includegraphics[width=0.97\linewidth]{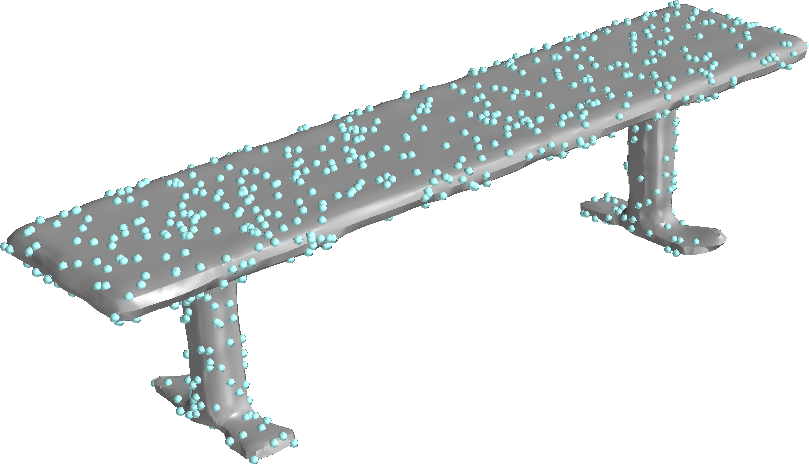}
    \endminipage\hfill
    \minipage{0.1427\linewidth}
    \centering
    \includegraphics[width=0.97\linewidth]{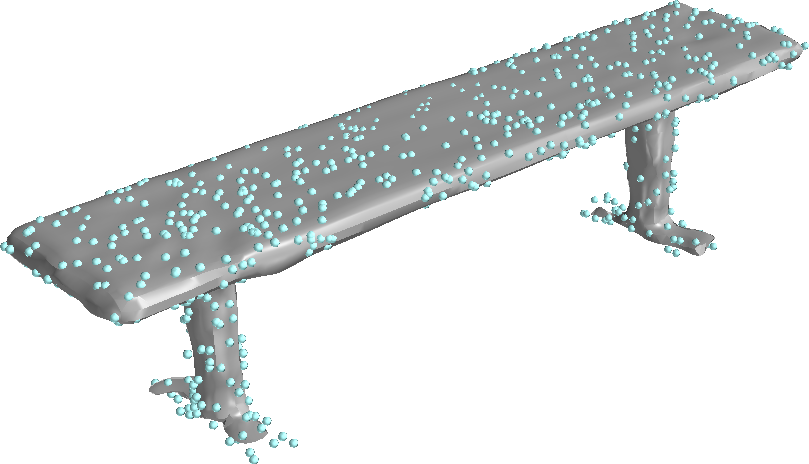}
    \endminipage\hfill
    \minipage{0.1427\linewidth}
    \centering
    \includegraphics[width=0.97\linewidth]{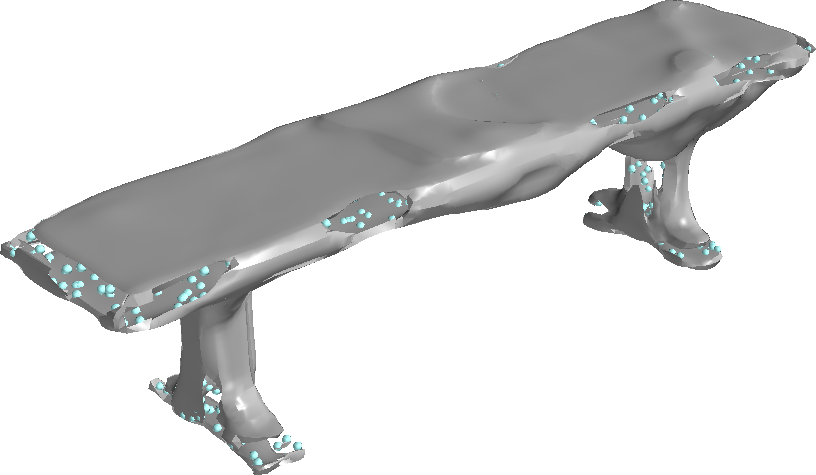}
    \endminipage\hfill
    \minipage{0.1427\linewidth}
    \centering
    \includegraphics[width=0.97\linewidth]{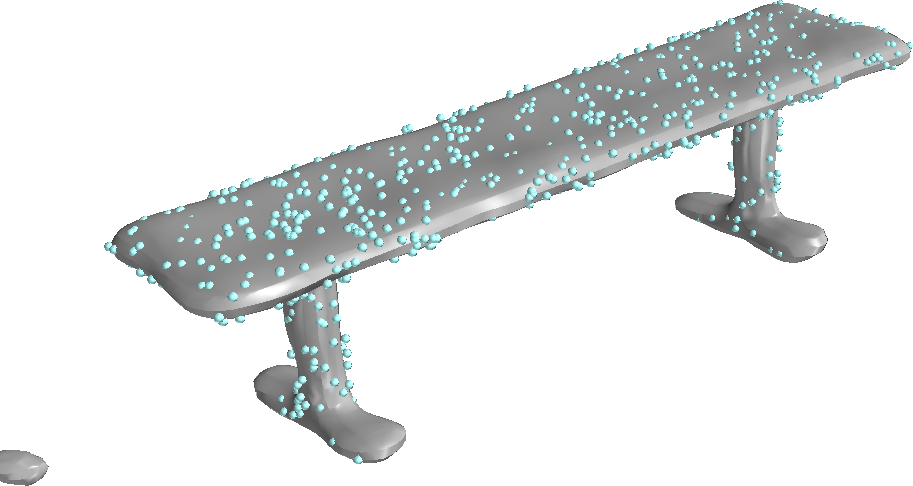}
    \endminipage\hfill
    \minipage{0.1427\linewidth}
    \centering
    \includegraphics[width=0.97\linewidth]{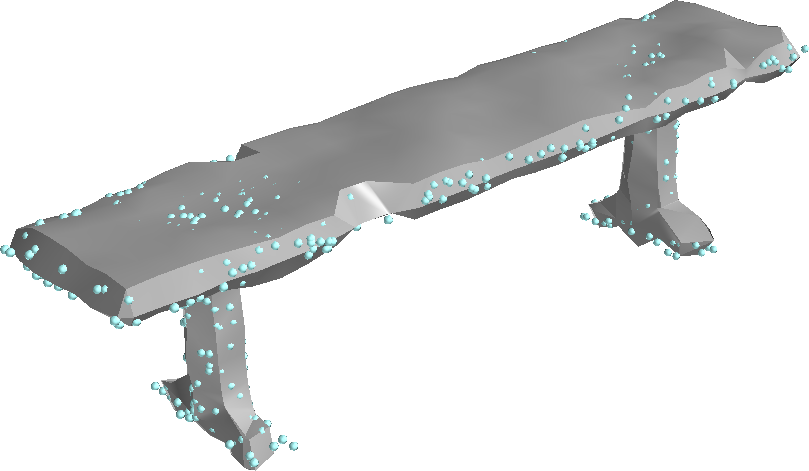}
    \endminipage\hfill
    \minipage{0.1427\linewidth}
    \centering
    \includegraphics[width=0.97\linewidth]{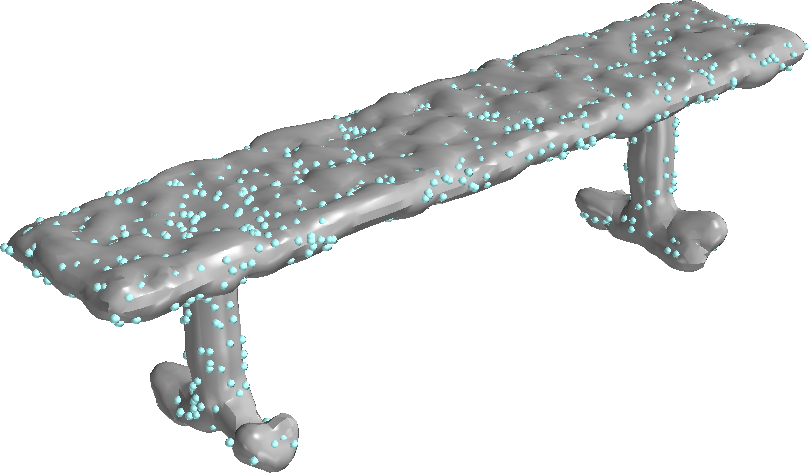}
    \endminipage\hfill
    \minipage{0.1427\linewidth}
    \centering
    \includegraphics[width=0.97\linewidth]{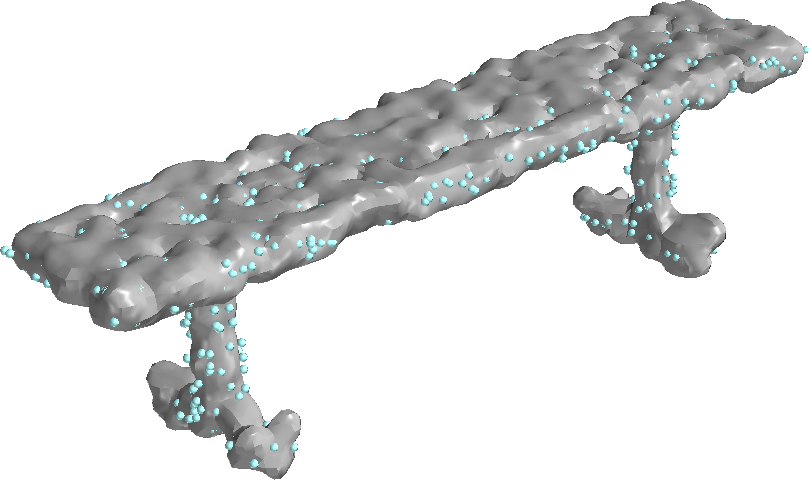}
    \endminipage\hfill
 \\
    \minipage{0.1427\linewidth}
    \centering
    \includegraphics[width=0.97\linewidth]{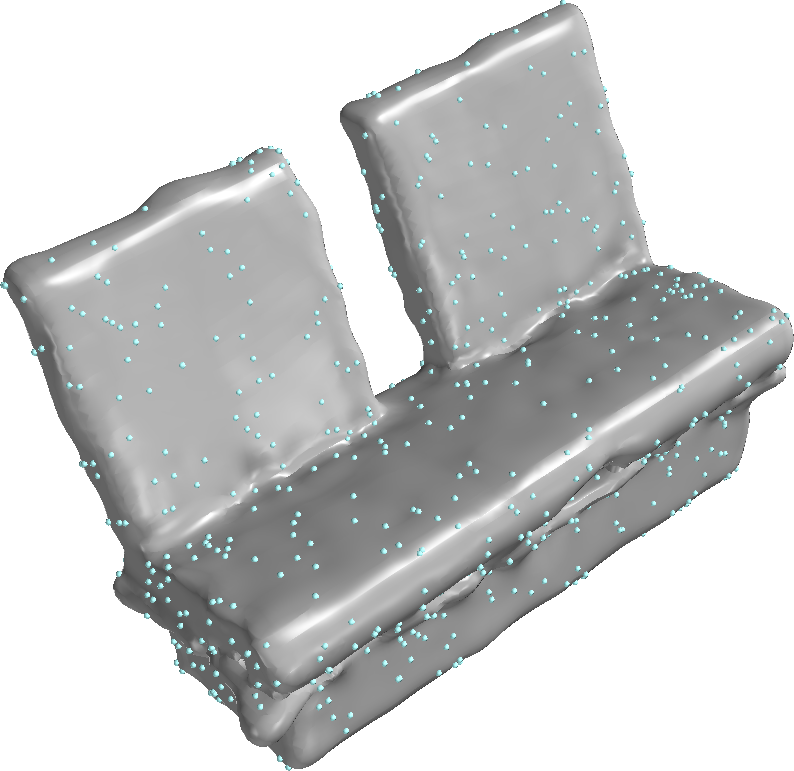}
    \endminipage\hfill
    \minipage{0.1427\linewidth}
    \centering
    \includegraphics[width=0.97\linewidth]{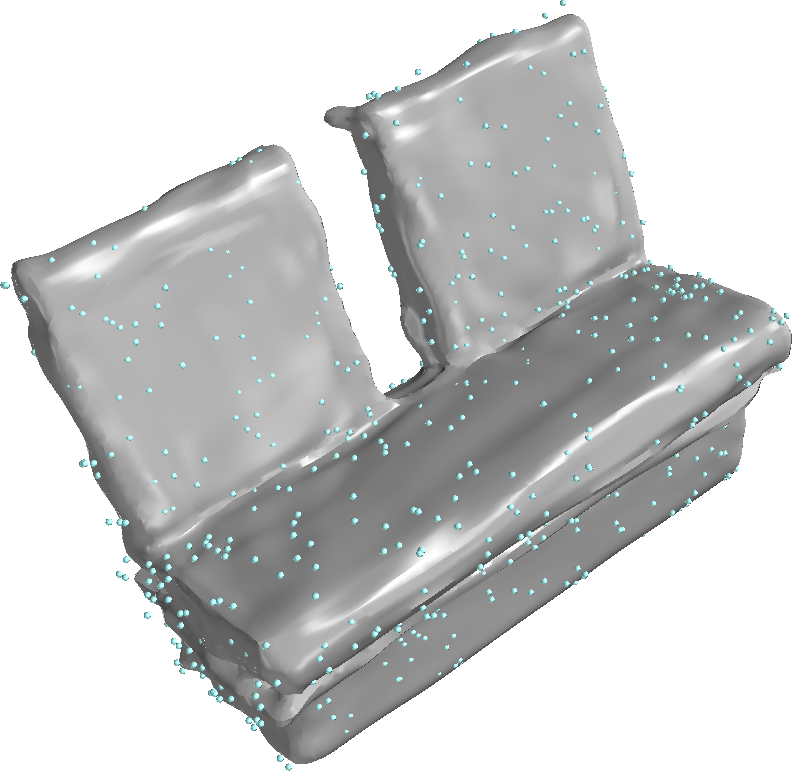}
    \endminipage\hfill
    \minipage{0.1427\linewidth}
    \centering
    \includegraphics[width=0.97\linewidth]{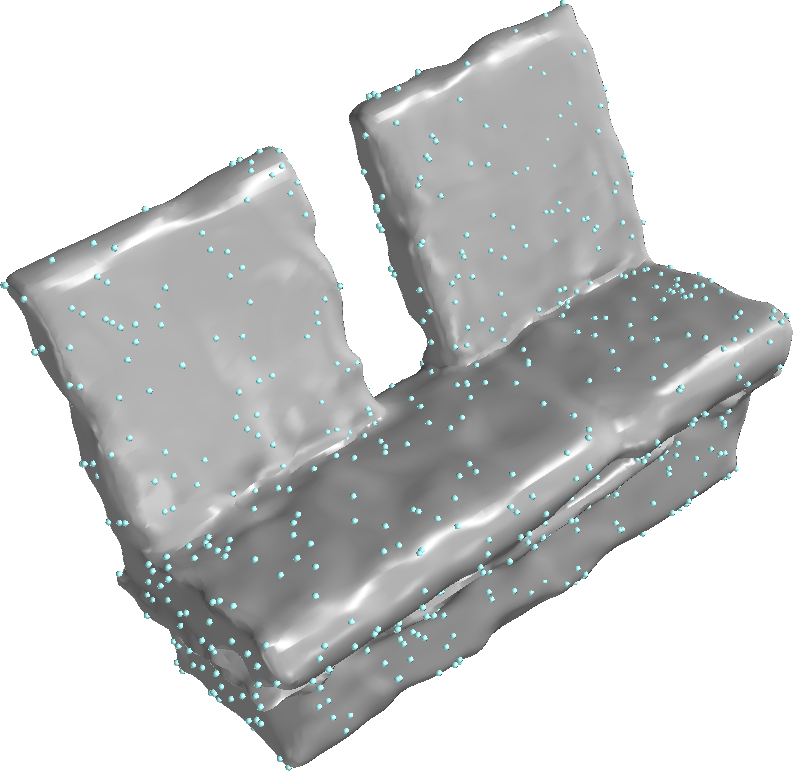}
    \endminipage\hfill
    \minipage{0.1427\linewidth}
    \centering
    \includegraphics[width=0.97\linewidth]{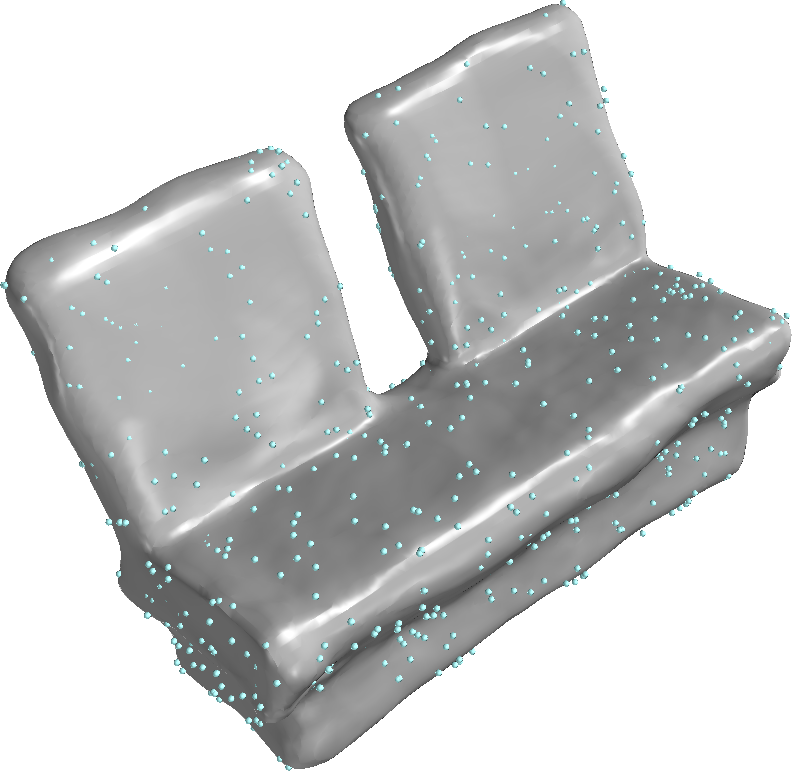}
    \endminipage\hfill
    \minipage{0.1427\linewidth}
    \centering
    \includegraphics[width=0.97\linewidth]{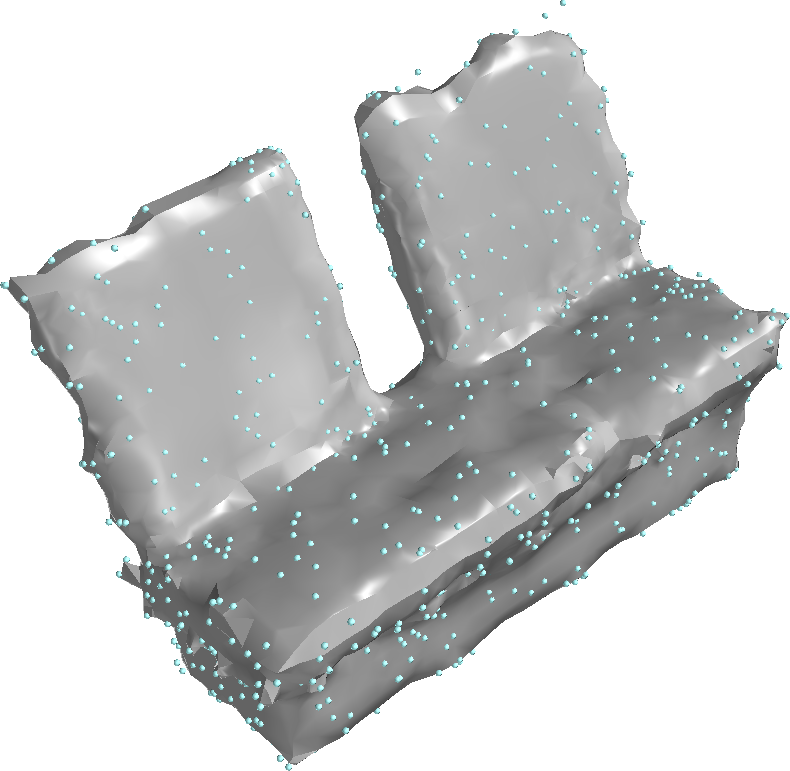}
    \endminipage\hfill
    \minipage{0.1427\linewidth}
    \centering
    \includegraphics[width=0.97\linewidth]{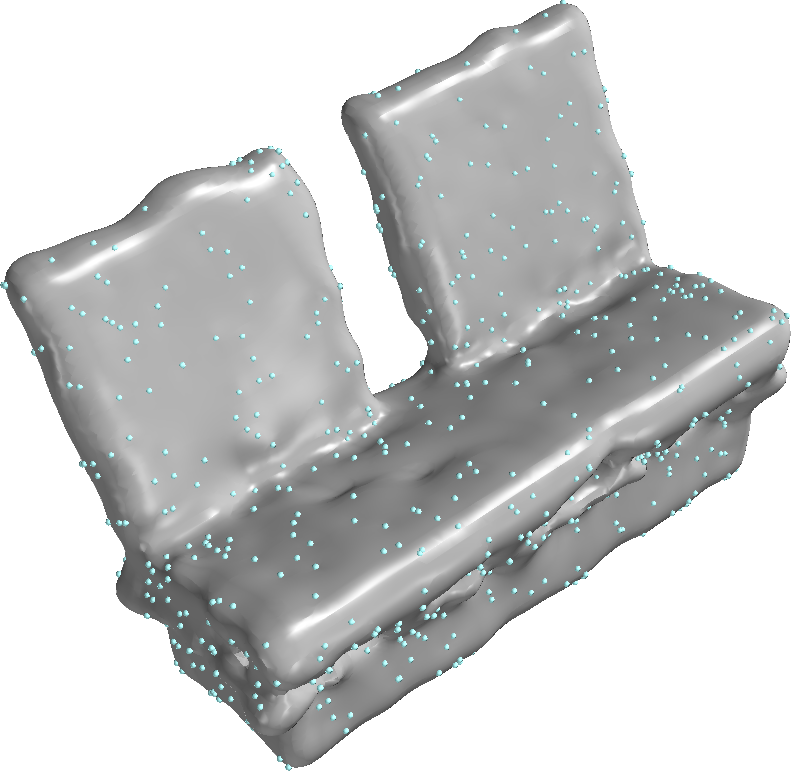}
    \endminipage\hfill
    \minipage{0.1427\linewidth}
    \centering
    \includegraphics[width=0.97\linewidth]{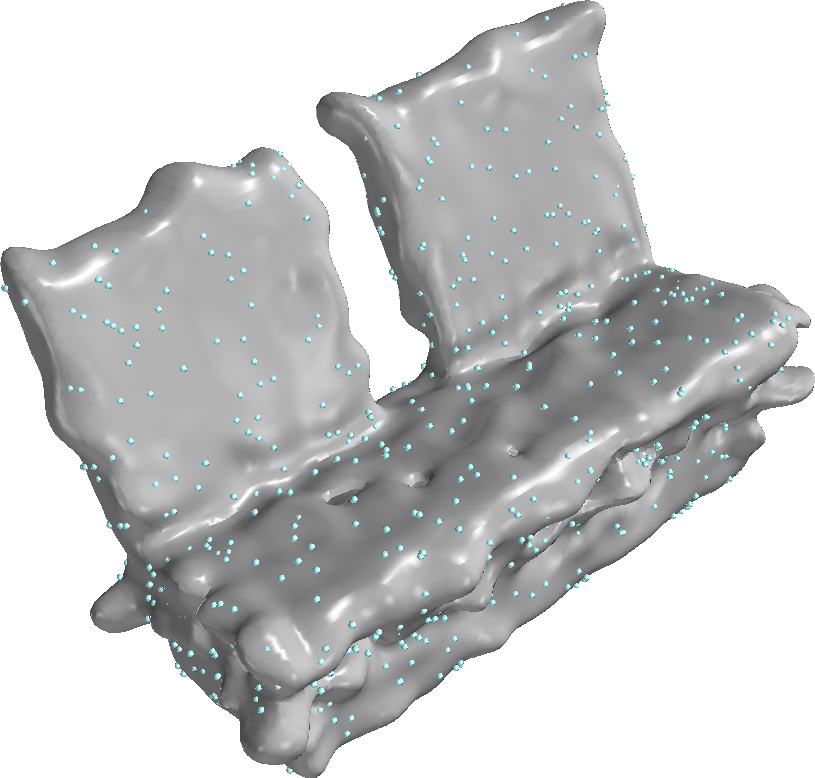}
    \endminipage\hfill
 \\
    \minipage{0.1427\linewidth}
    \centering
    \includegraphics[width=0.97\linewidth]{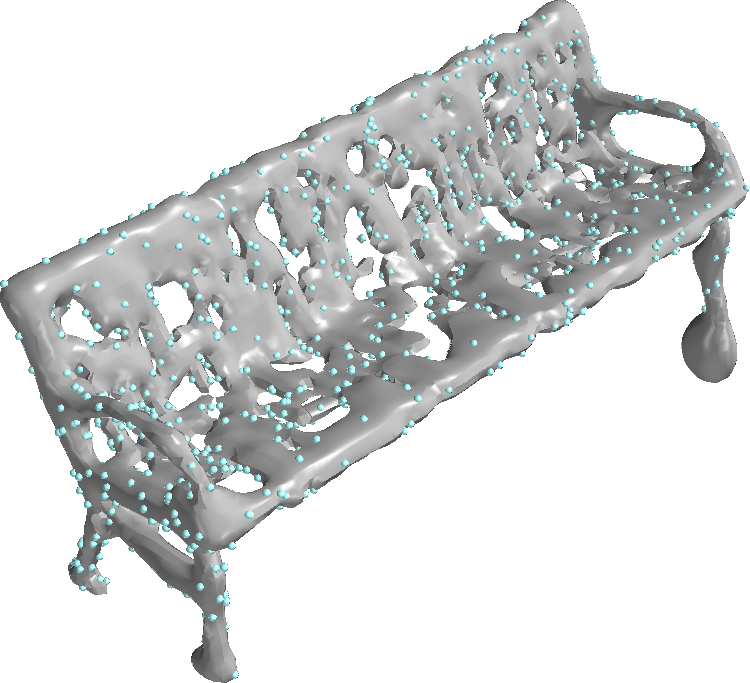}
    \endminipage\hfill
    \minipage{0.1427\linewidth}
    \centering
    \includegraphics[width=0.97\linewidth]{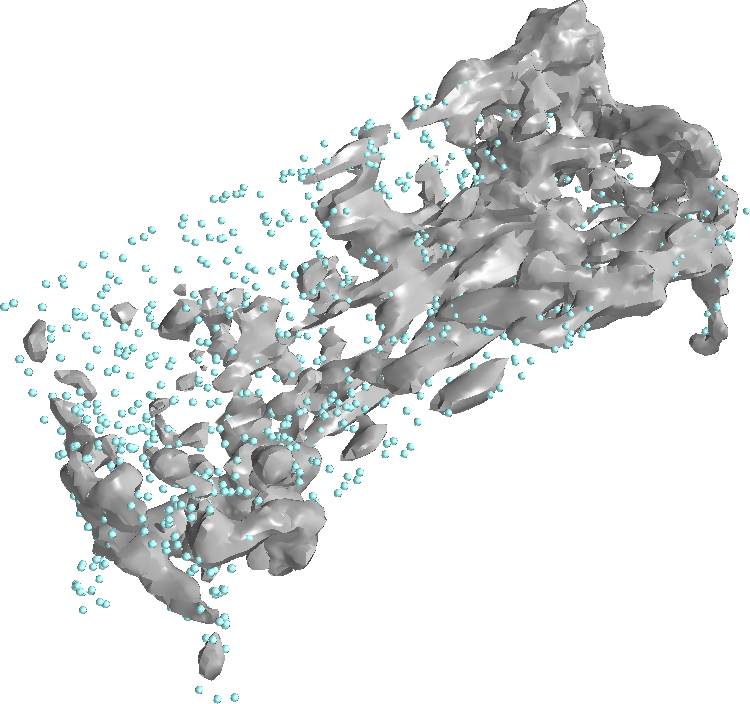}
    \endminipage\hfill
    \minipage{0.1427\linewidth}
    \centering
    \includegraphics[width=0.97\linewidth]{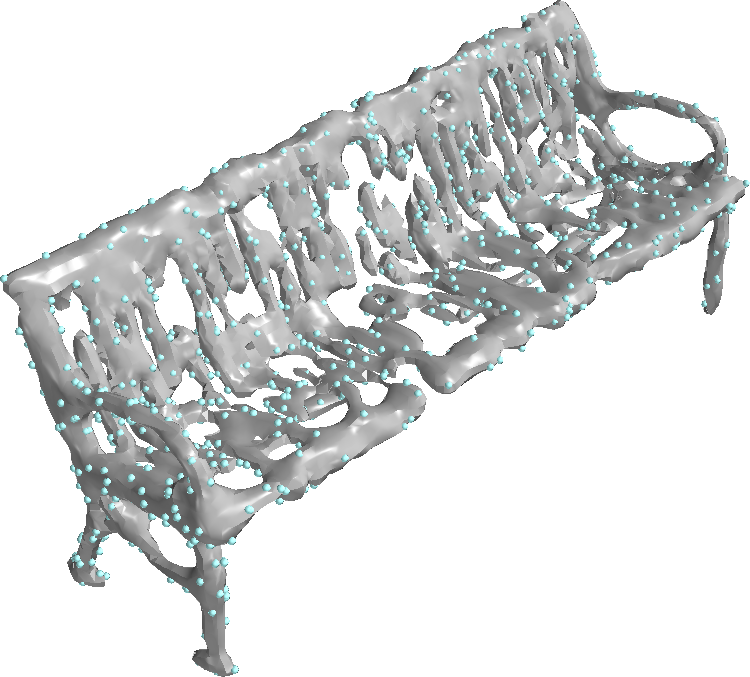}
    \endminipage\hfill
    \minipage{0.1427\linewidth}
    \centering
    \includegraphics[width=0.97\linewidth]{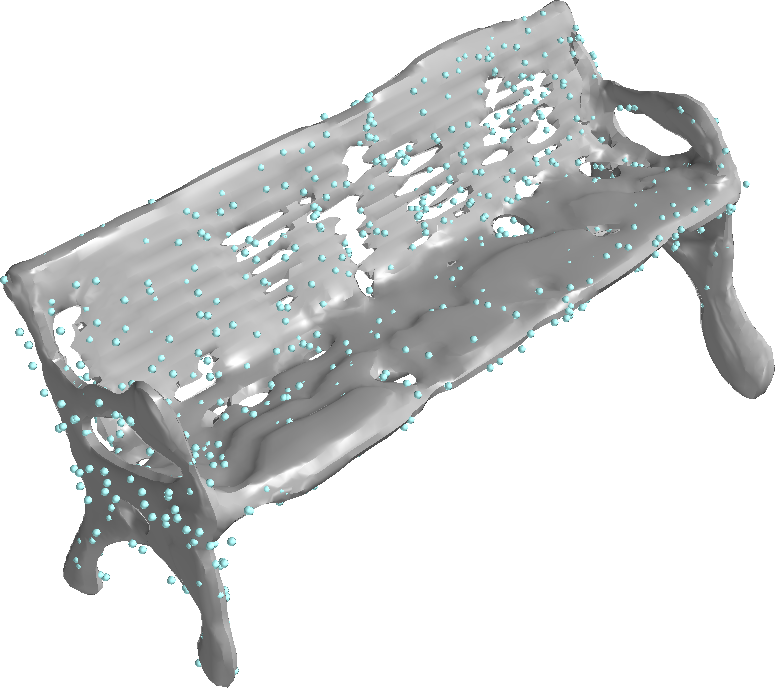}
    \endminipage\hfill
    \minipage{0.1427\linewidth}
    \centering
    \includegraphics[width=0.97\linewidth]{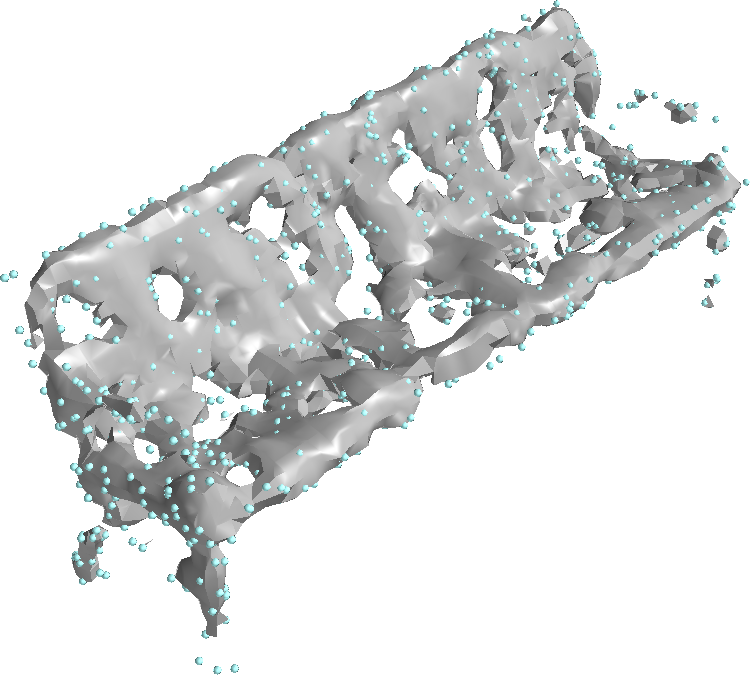}
    \endminipage\hfill
    \minipage{0.1427\linewidth}
    \centering
    \includegraphics[width=0.97\linewidth]{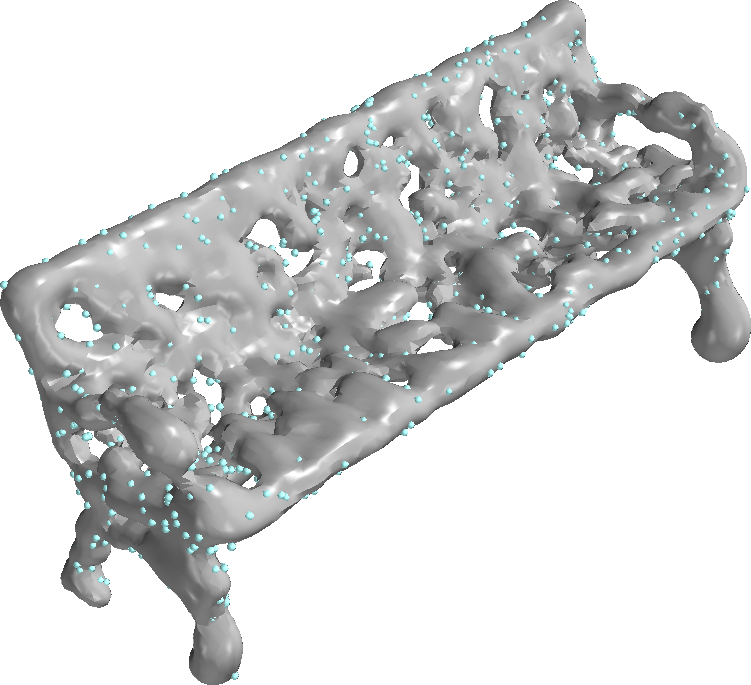}
    \endminipage\hfill
    \minipage{0.1427\linewidth}
    \centering
    \includegraphics[width=0.97\linewidth]{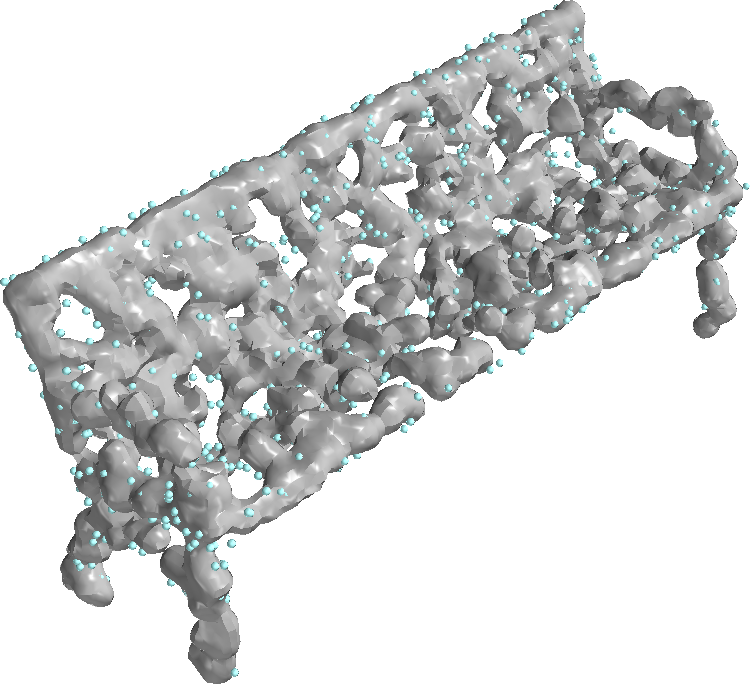}
    \endminipage\hfill
\\
    \minipage{0.1427\linewidth}
    \centering
    \includegraphics[width=0.97\linewidth]{figures/shapenet_benchmark/car_recon_6/figure_n-splines-g.png}
    \endminipage\hfill
    \minipage{0.1427\linewidth}
    \centering
    \includegraphics[width=0.97\linewidth]{figures/shapenet_benchmark/car_recon_6/figure_igr.png}
    \endminipage\hfill
    \minipage{0.1427\linewidth}
    \centering
    \includegraphics[width=0.97\linewidth]{figures/shapenet_benchmark/car_recon_6/figure_siren.png}
    \endminipage\hfill
    \minipage{0.1427\linewidth}
    \centering
    \includegraphics[width=0.97\linewidth]{figures/shapenet_benchmark/car_recon_6/figure_ffk.png}
    \endminipage\hfill
    \minipage{0.1427\linewidth}
    \centering
    \includegraphics[width=0.97\linewidth]{figures/shapenet_benchmark/car_recon_6/figure_poisson.png}
    \endminipage\hfill
    \minipage{0.1427\linewidth}
    \centering
    \includegraphics[width=0.97\linewidth]{figures/shapenet_benchmark/car_recon_6/figure_rbf.png}
    \endminipage\hfill
    \minipage{0.1427\linewidth}
    \centering
    \includegraphics[width=0.97\linewidth]{figures/shapenet_benchmark/car_recon_6/figure_svr.png}
    \endminipage\hfill
 \\
    \minipage{0.1427\linewidth}
    \centering
    \includegraphics[width=0.97\linewidth]{figures/shapenet_benchmark/chair_recon_11/figure_n-splines-g.png}
    \endminipage\hfill
    \minipage{0.1427\linewidth}
    \centering
    \includegraphics[width=0.97\linewidth]{figures/shapenet_benchmark/chair_recon_11/figure_igr.png}
    \endminipage\hfill
    \minipage{0.1427\linewidth}
    \centering
    \includegraphics[width=0.97\linewidth]{figures/shapenet_benchmark/chair_recon_11/figure_siren.png}
    \endminipage\hfill
    \minipage{0.1427\linewidth}
    \centering
    \includegraphics[width=0.97\linewidth]{figures/shapenet_benchmark/chair_recon_11/figure_ffk.png}
    \endminipage\hfill
    \minipage{0.1427\linewidth}
    \centering
    \includegraphics[width=0.97\linewidth]{figures/shapenet_benchmark/chair_recon_11/figure_poisson.png}
    \endminipage\hfill
    \minipage{0.1427\linewidth}
    \centering
    \includegraphics[width=0.97\linewidth]{figures/shapenet_benchmark/chair_recon_11/figure_rbf.png}
    \endminipage\hfill
    \minipage{0.1427\linewidth}
    \centering
    \includegraphics[width=0.97\linewidth]{figures/shapenet_benchmark/chair_recon_11/figure_svr.png}
    \endminipage\hfill
    \vspace{0.5em}
\\
    \minipage{0.1427\linewidth}
    \centering
    \includegraphics[width=0.97\linewidth]{figures/shapenet_benchmark/display_recon_16/figure_n-splines-g.png}
    \endminipage\hfill
    \minipage{0.1427\linewidth}
    \centering
    \includegraphics[width=0.97\linewidth]{figures/shapenet_benchmark/display_recon_16/figure_igr.png}
    \endminipage\hfill
    \minipage{0.1427\linewidth}
    \centering
    \includegraphics[width=0.97\linewidth]{figures/shapenet_benchmark/display_recon_16/figure_siren.png}
    \endminipage\hfill
    \minipage{0.1427\linewidth}
    \centering
    \includegraphics[width=0.97\linewidth]{figures/shapenet_benchmark/display_recon_16/figure_ffk.png}
    \endminipage\hfill
    \minipage{0.1427\linewidth}
    \centering
    \includegraphics[width=0.97\linewidth]{figures/shapenet_benchmark/display_recon_16/figure_poisson.png}
    \endminipage\hfill
    \minipage{0.1427\linewidth}
    \centering
    \includegraphics[width=0.97\linewidth]{figures/shapenet_benchmark/display_recon_16/figure_rbf.png}
    \endminipage\hfill
    \minipage{0.1427\linewidth}
    \centering
    \includegraphics[width=0.97\linewidth]{figures/shapenet_benchmark/display_recon_16/figure_svr.png}
    \endminipage\hfill
\\
    \minipage{0.1427\linewidth}
    \centering
    \includegraphics[width=0.97\linewidth]{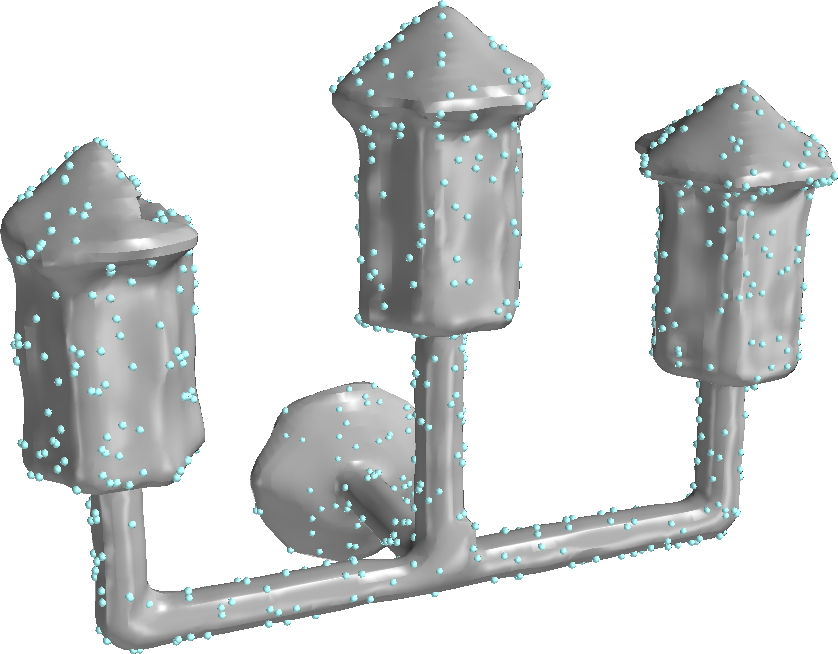}
    \endminipage\hfill
    \minipage{0.1427\linewidth}
    \centering
    \includegraphics[width=0.97\linewidth]{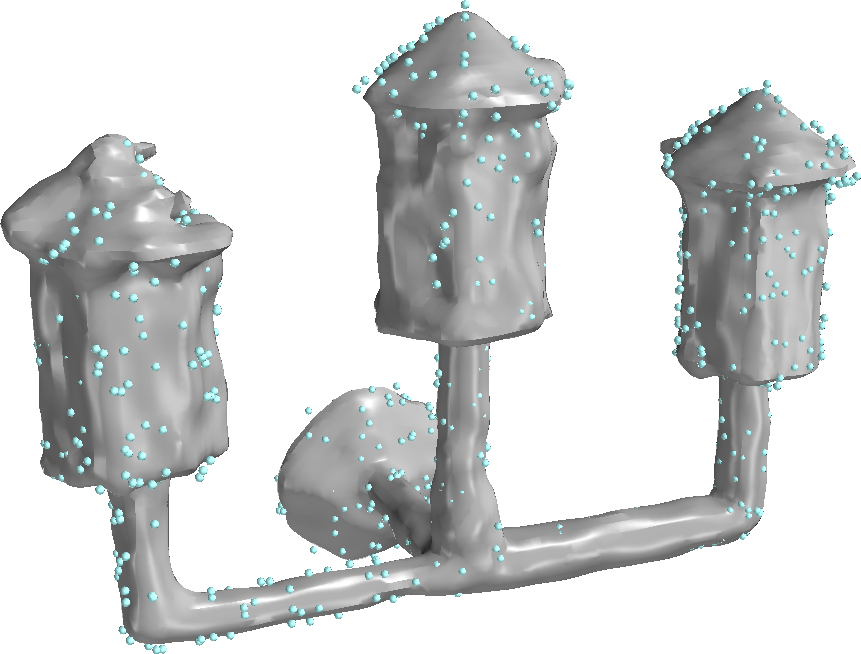}
    \endminipage\hfill
    \minipage{0.1427\linewidth}
    \centering
    \includegraphics[width=0.97\linewidth]{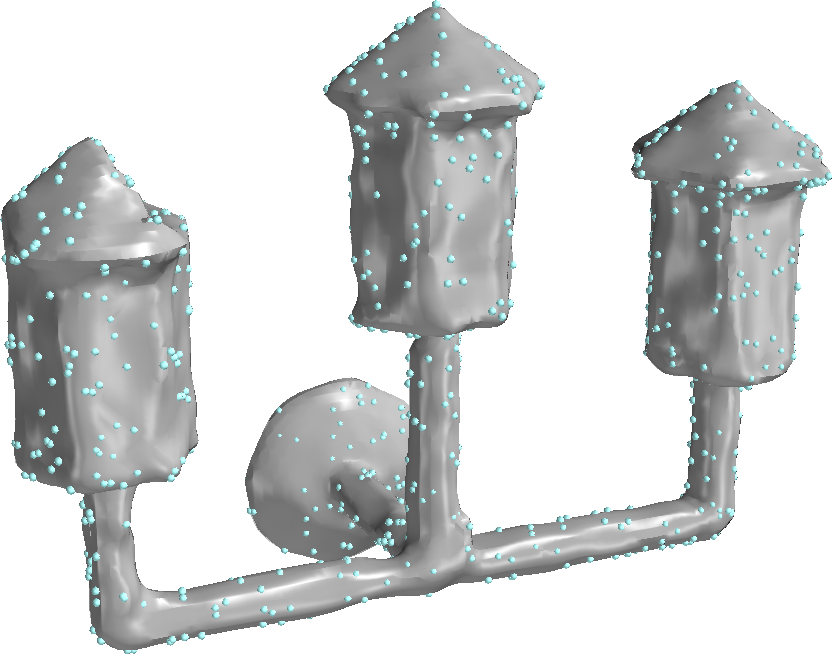}
    \endminipage\hfill
    \minipage{0.1427\linewidth}
    \centering
    \includegraphics[width=0.97\linewidth]{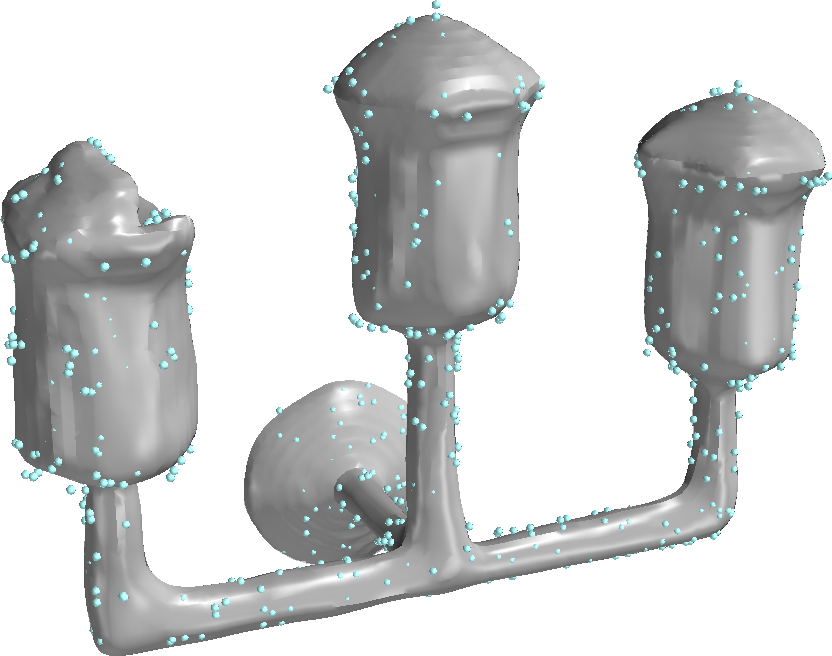}
    \endminipage\hfill
    \minipage{0.1427\linewidth}
    \centering
    \includegraphics[width=0.97\linewidth]{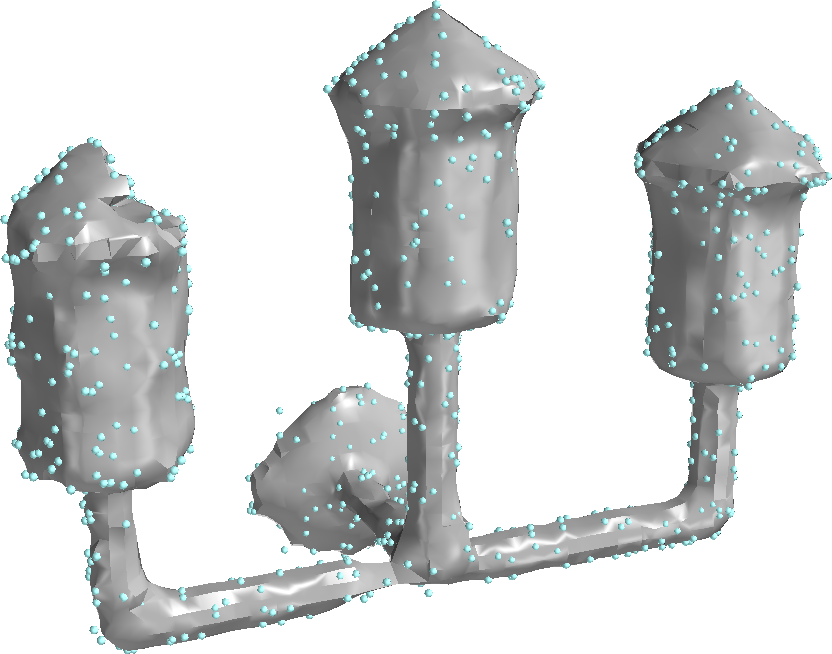}
    \endminipage\hfill
    \minipage{0.1427\linewidth}
    \centering
    \includegraphics[width=0.97\linewidth]{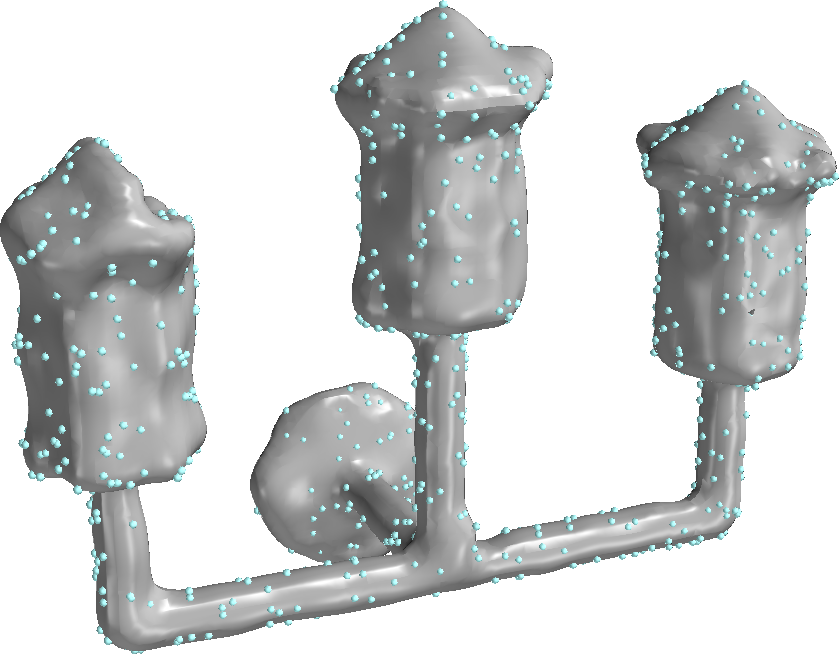}
    \endminipage\hfill
    \minipage{0.1427\linewidth}
    \centering
    \includegraphics[width=0.97\linewidth]{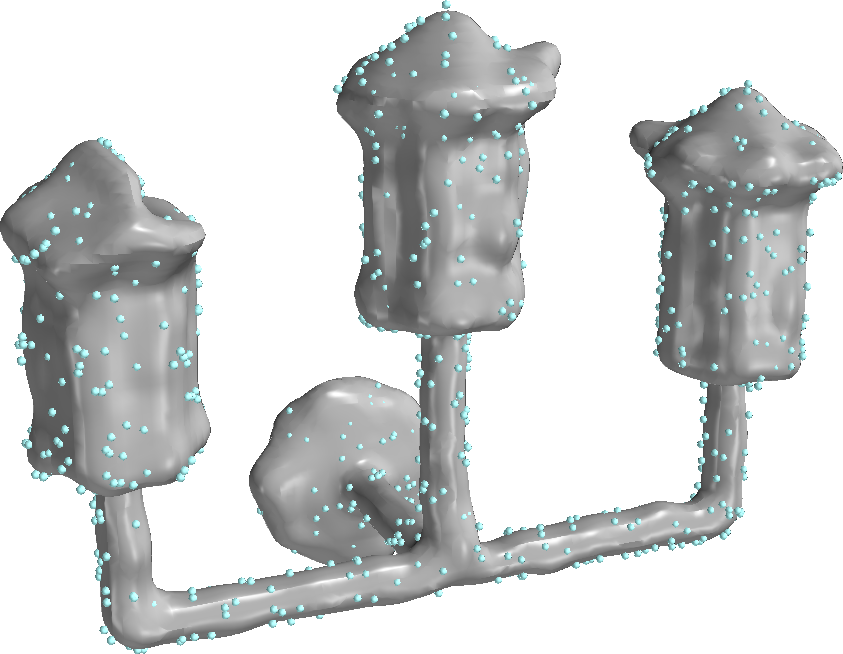}
    \endminipage\hfill
    \vspace{0.5em}
    \hfill
    \minipage{0.1427\linewidth}
    \centering \footnotesize \textbf{Ours}
    \endminipage\hfill
    \minipage{0.1427\linewidth}
    \centering \footnotesize IGR
    \endminipage\hfill
    \minipage{0.1427\linewidth}
    \centering \footnotesize SIREN
    \endminipage\hfill
    \minipage{0.1427\linewidth}
    \centering \footnotesize Fourier Feats
    \endminipage\hfill
    \minipage{0.1427\linewidth}
    \centering \footnotesize Poisson
    \endminipage\hfill
    \minipage{0.1427\linewidth}
    \centering \footnotesize Biharmonic
    \endminipage\hfill
    \minipage{0.1427\linewidth}
    \centering \footnotesize SVR
    \endminipage\hfill
    \vspace{0.5em}
    \caption{Comparisons between reconstruction techniques on Shapenet models. The blue points are the input points to the reconstruction algorithm.}\label{fig:shapenet1}
\end{figure}

\begin{figure}[H]
    \minipage{0.1427\linewidth}
    \centering
    \includegraphics[width=0.97\linewidth]{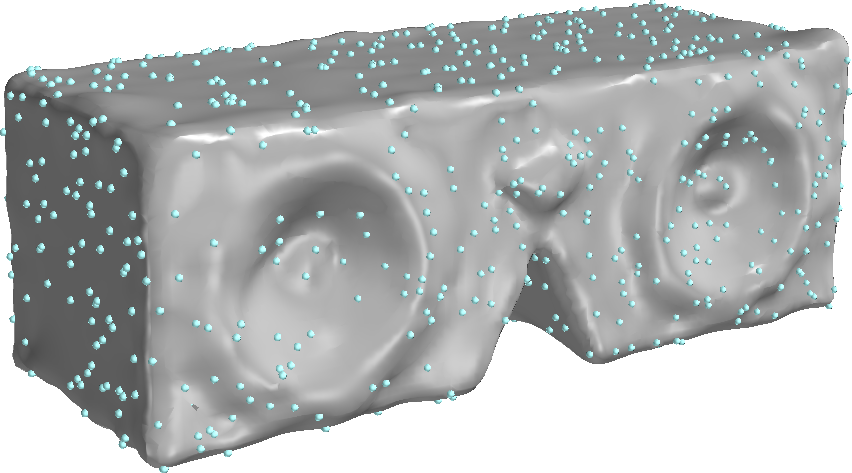}
    \endminipage\hfill
    \minipage{0.1427\linewidth}
    \centering
    \includegraphics[width=0.97\linewidth]{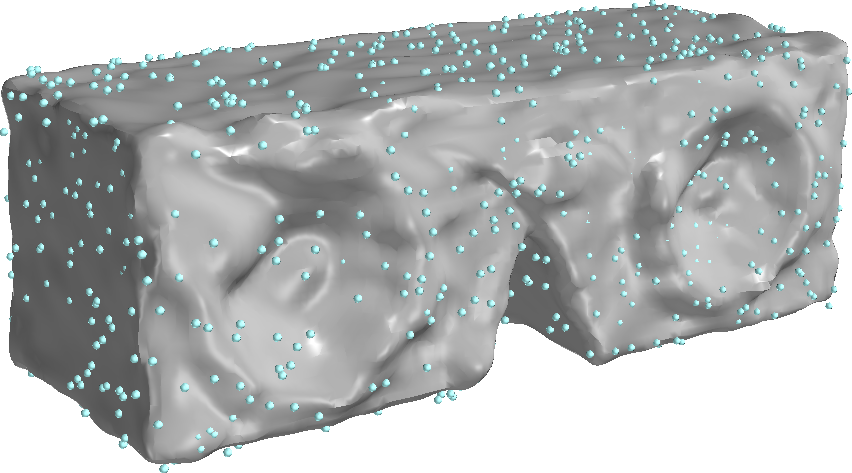}
    \endminipage\hfill
    \minipage{0.1427\linewidth}
    \centering
    \includegraphics[width=0.97\linewidth]{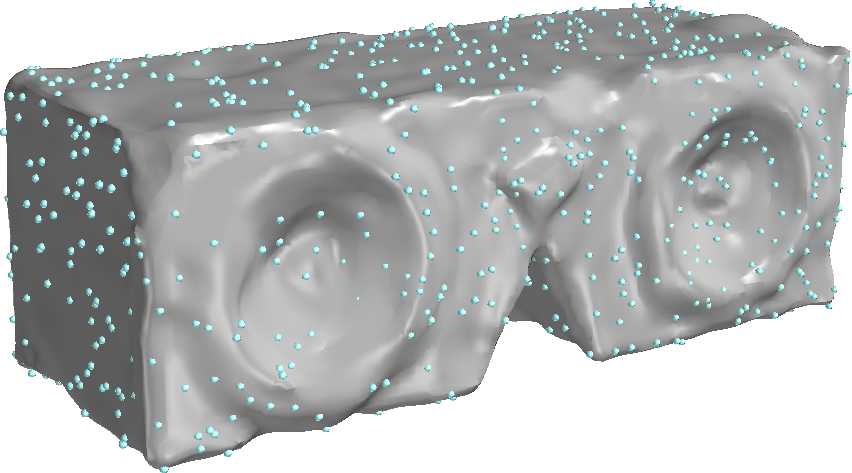}
    \endminipage\hfill
    \minipage{0.1427\linewidth}
    \centering
    \includegraphics[width=0.97\linewidth]{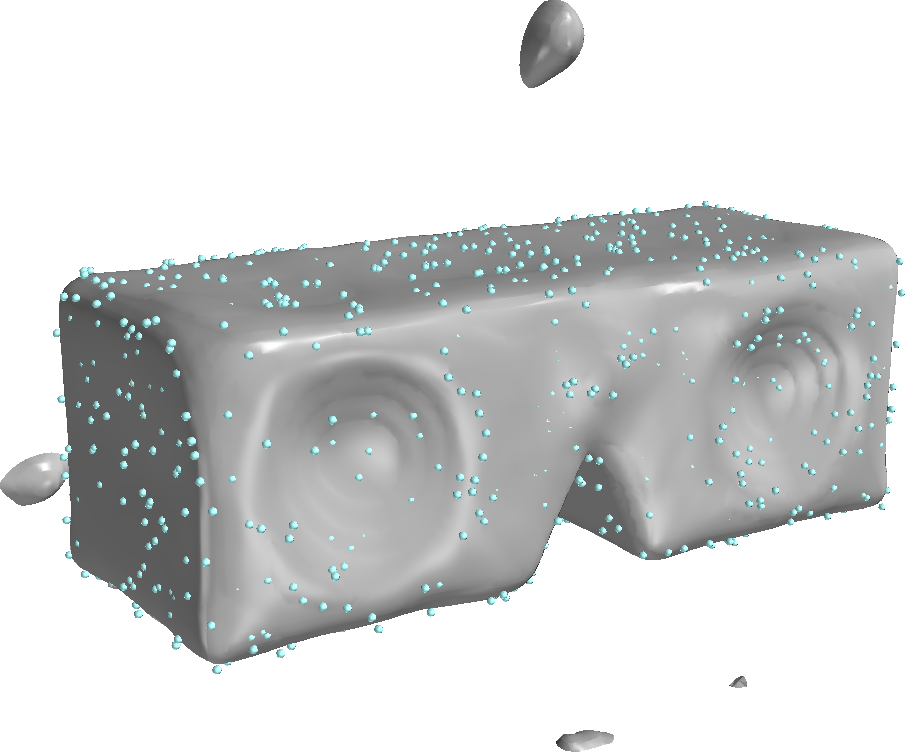}
    \endminipage\hfill
    \minipage{0.1427\linewidth}
    \centering
    \includegraphics[width=0.97\linewidth]{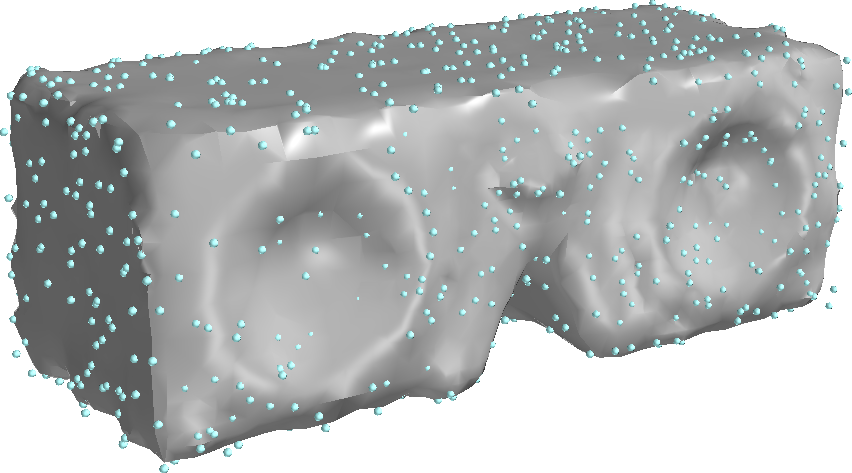}
    \endminipage\hfill
    \minipage{0.1427\linewidth}
    \centering
    \includegraphics[width=0.97\linewidth]{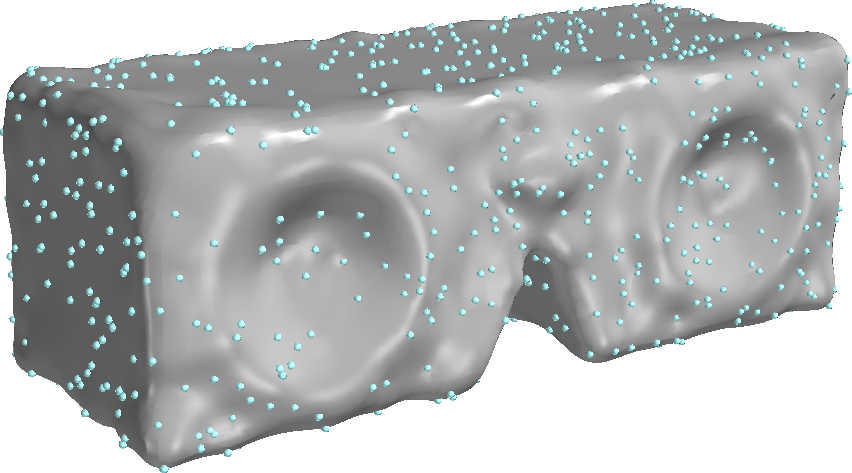}
    \endminipage\hfill
    \minipage{0.1427\linewidth}
    \centering
    \includegraphics[width=0.97\linewidth]{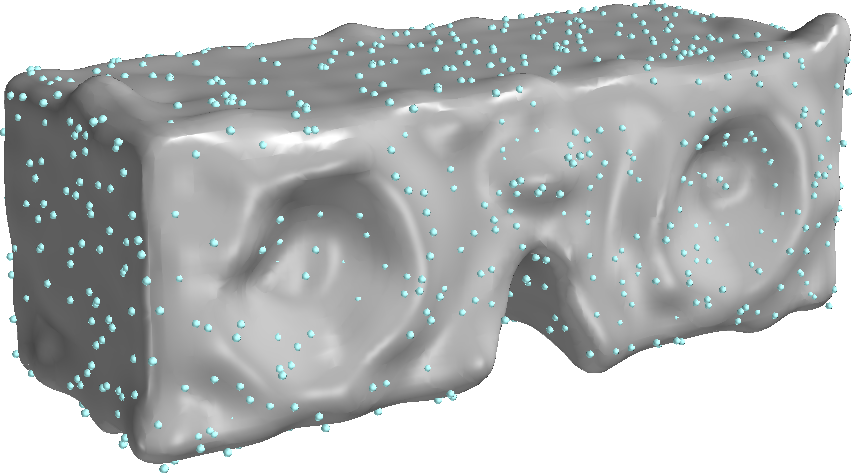}
    \endminipage\hfill
\\
    \minipage{0.1427\linewidth}
    \centering
    \includegraphics[width=0.97\linewidth]{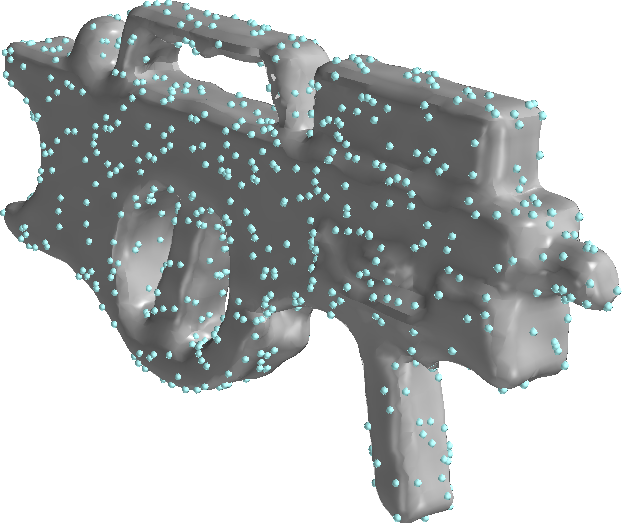}
    \endminipage\hfill
    \minipage{0.1427\linewidth}
    \centering
    \includegraphics[width=0.97\linewidth]{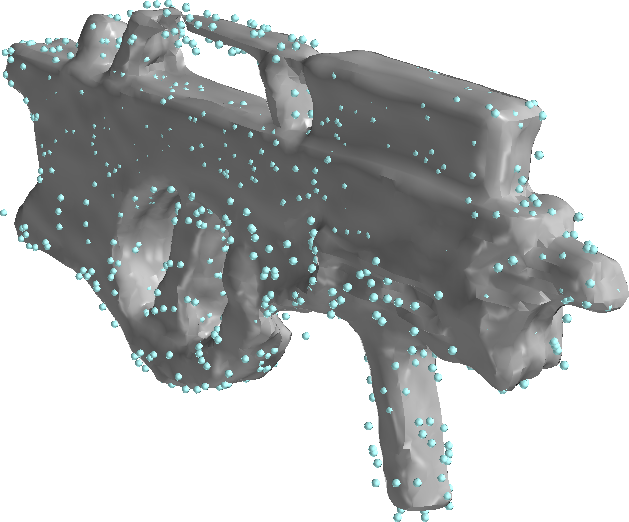}
    \endminipage\hfill
    \minipage{0.1427\linewidth}
    \centering
    \includegraphics[width=0.97\linewidth]{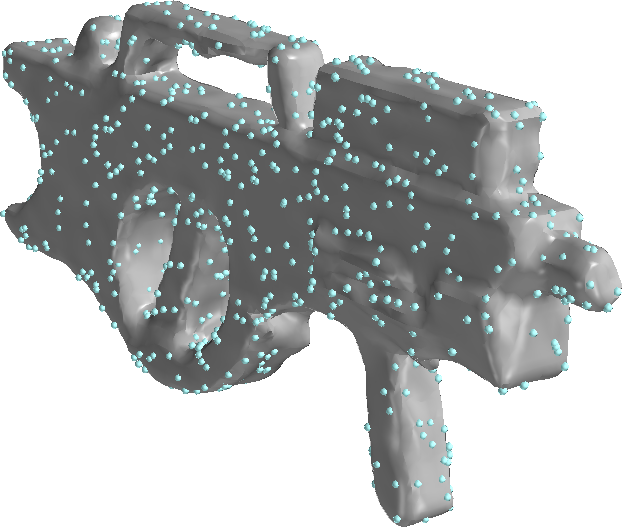}
    \endminipage\hfill
    \minipage{0.1427\linewidth}
    \centering
    \includegraphics[width=0.97\linewidth]{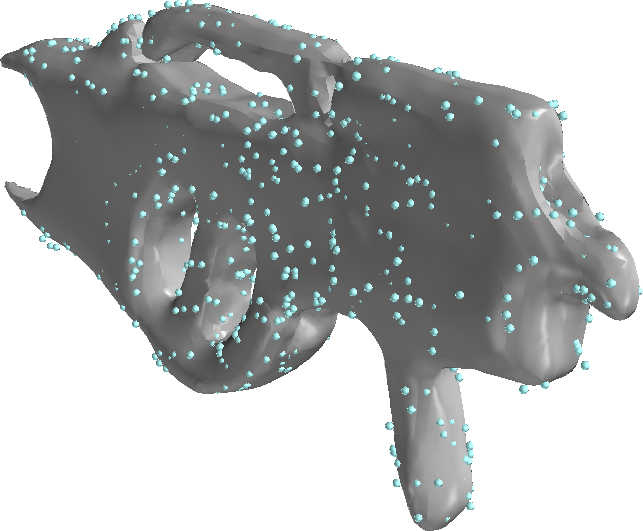}
    \endminipage\hfill
    \minipage{0.1427\linewidth}
    \centering
    \includegraphics[width=0.97\linewidth]{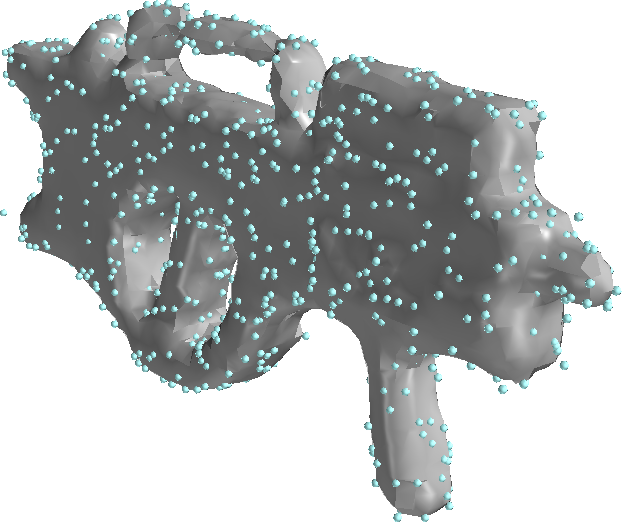}
    \endminipage\hfill
    \minipage{0.1427\linewidth}
    \centering
    \includegraphics[width=0.97\linewidth]{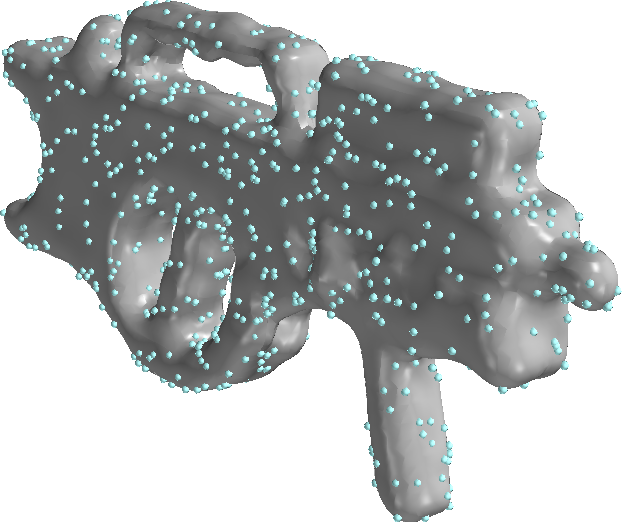}
    \endminipage\hfill
    \minipage{0.1427\linewidth}
    \centering
    \includegraphics[width=0.97\linewidth]{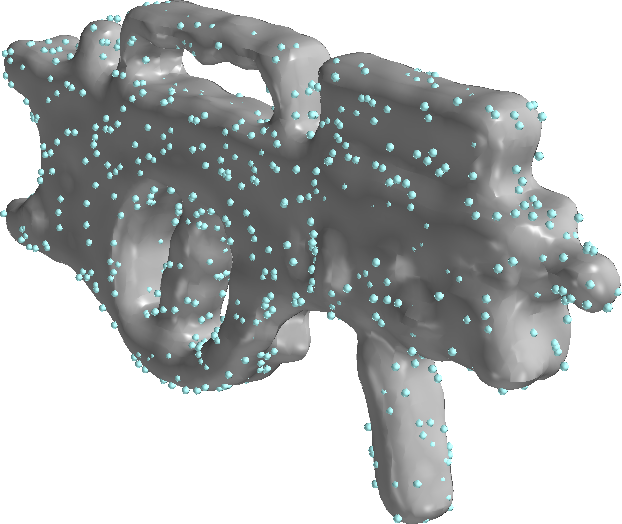}
    \endminipage\hfill
    \vspace{0.5em}
\\
    \minipage{0.1427\linewidth}
    \centering
    \includegraphics[width=0.97\linewidth]{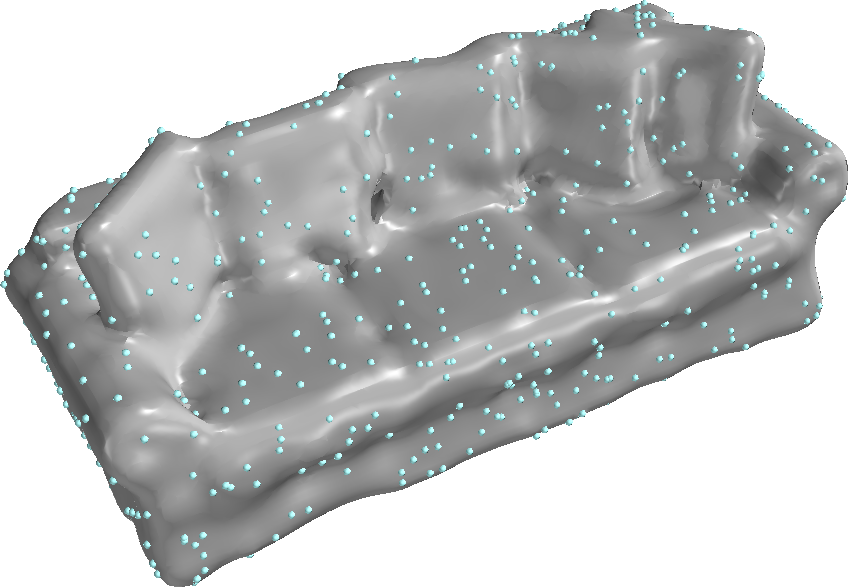}
    \endminipage\hfill
    \minipage{0.1427\linewidth}
    \centering
    \includegraphics[width=0.97\linewidth]{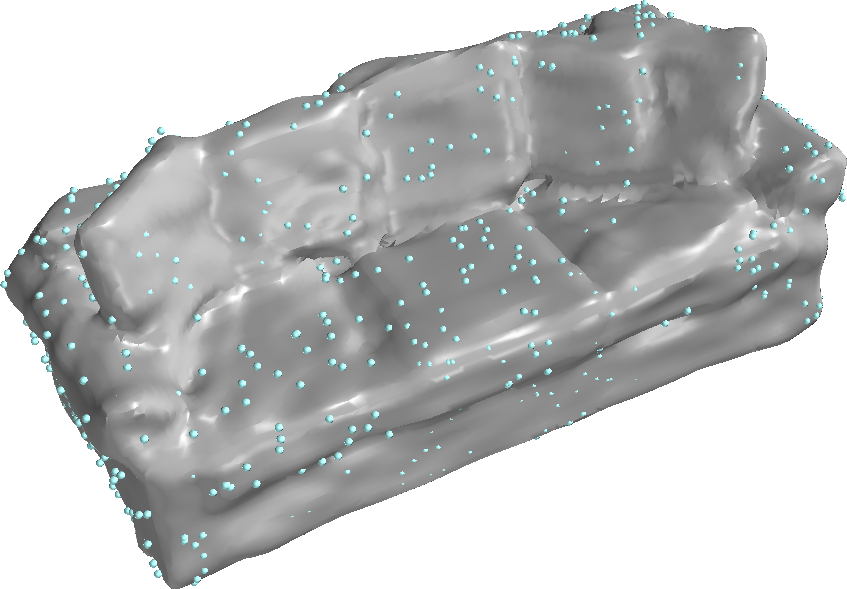}
    \endminipage\hfill
    \minipage{0.1427\linewidth}
    \centering
    \includegraphics[width=0.97\linewidth]{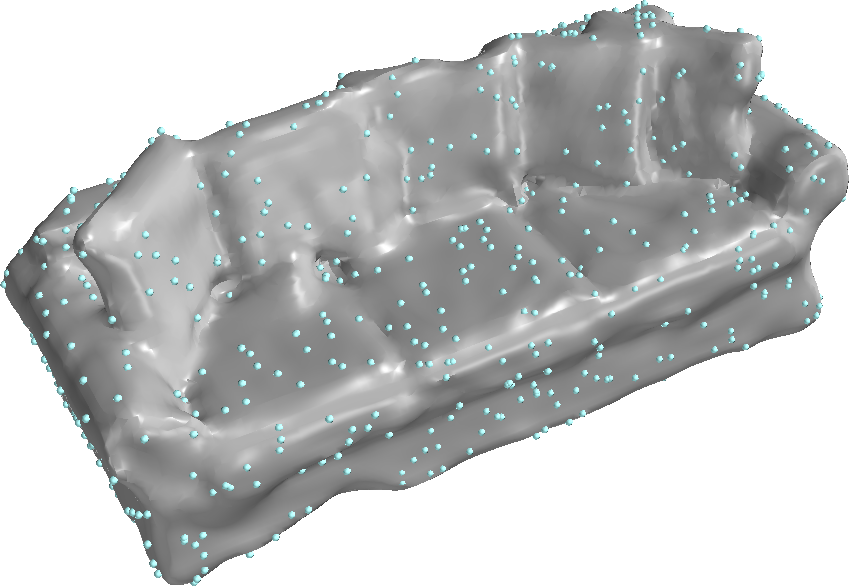}
    \endminipage\hfill
    \minipage{0.1427\linewidth}
    \centering
    \includegraphics[width=0.97\linewidth]{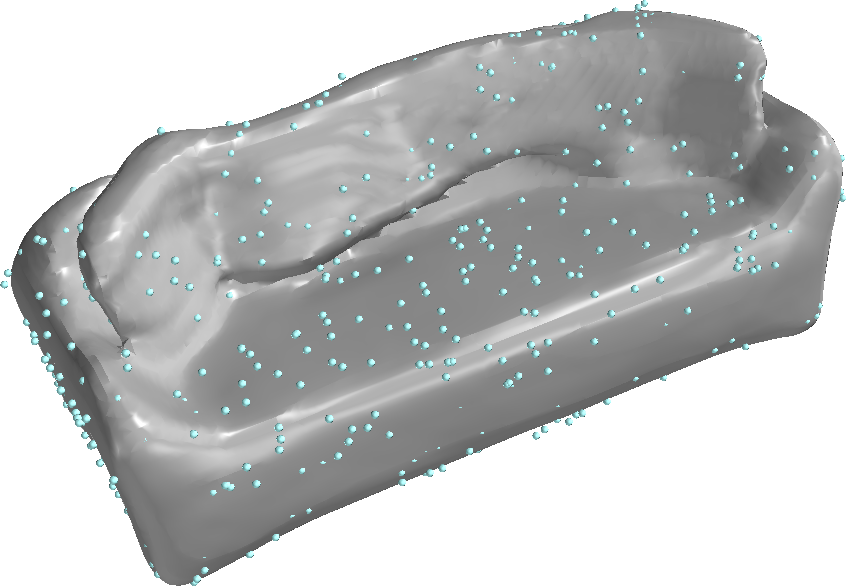}
    \endminipage\hfill
    \minipage{0.1427\linewidth}
    \centering
    \includegraphics[width=0.97\linewidth]{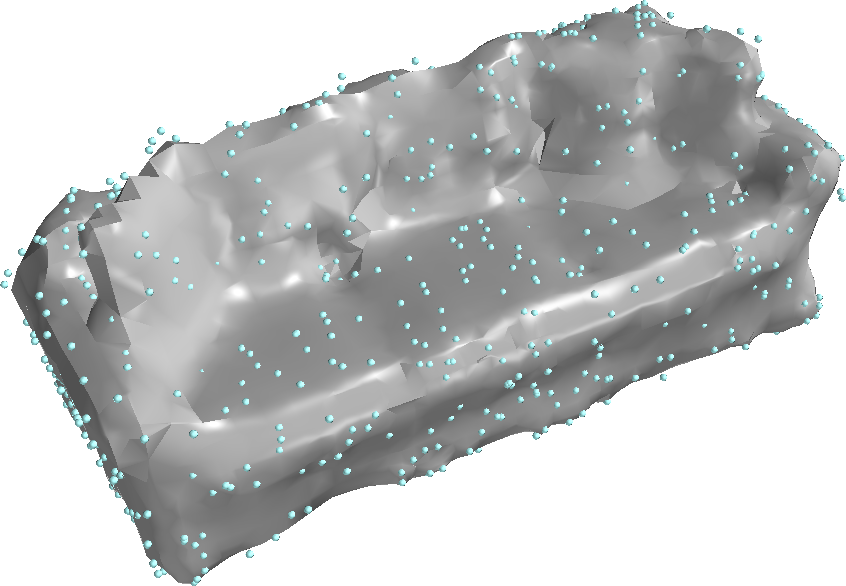}
    \endminipage\hfill
    \minipage{0.1427\linewidth}
    \centering
    \includegraphics[width=0.97\linewidth]{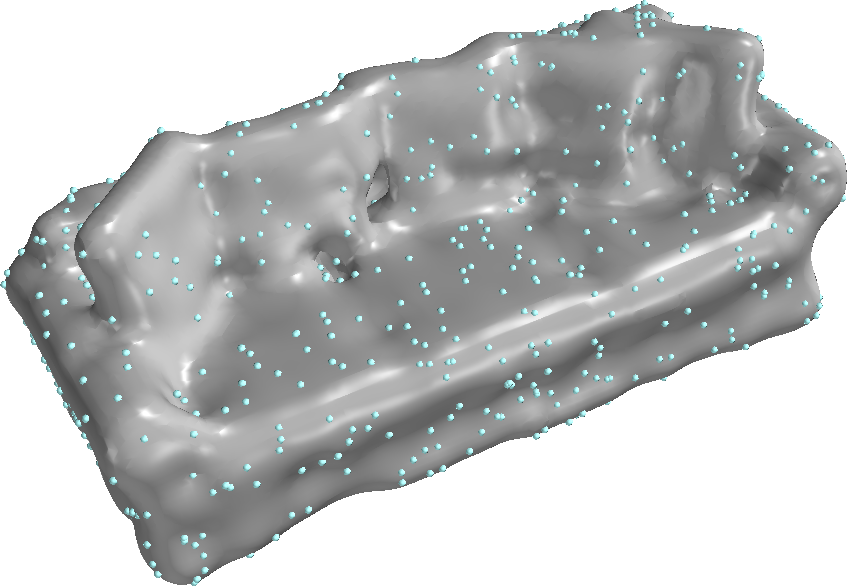}
    \endminipage\hfill
    \minipage{0.1427\linewidth}
    \centering
    \includegraphics[width=0.97\linewidth]{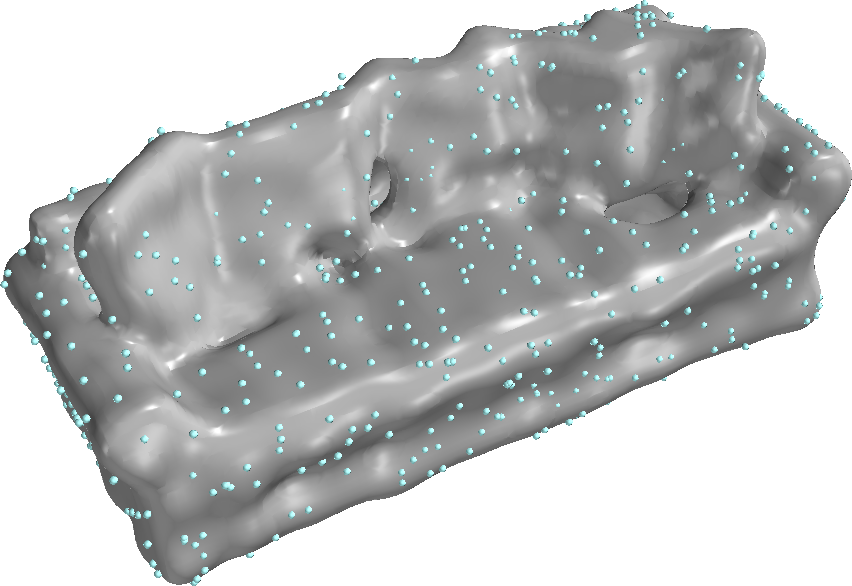}
    \endminipage\hfill
\\
    \minipage{0.1427\linewidth}
    \centering
    \includegraphics[width=0.97\linewidth]{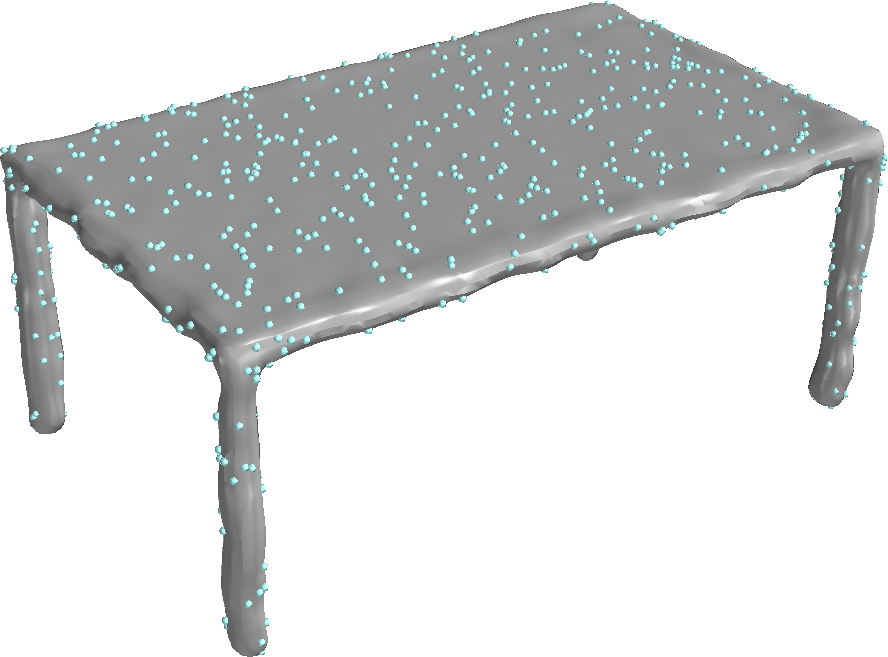}
    \endminipage\hfill
    \minipage{0.1427\linewidth}
    \centering
    \includegraphics[width=0.97\linewidth]{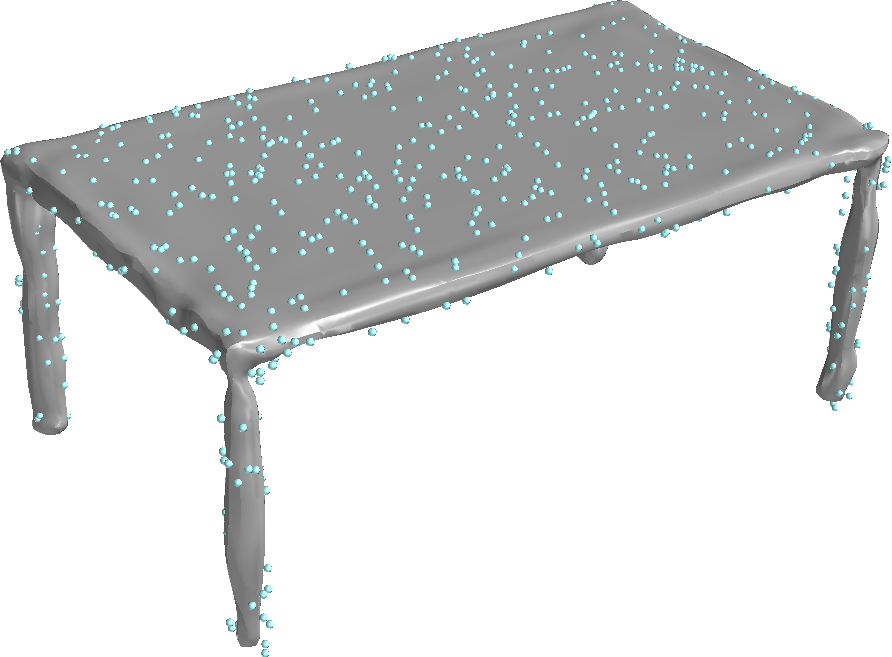}
    \endminipage\hfill
    \minipage{0.1427\linewidth}
    \centering
    \includegraphics[width=0.97\linewidth]{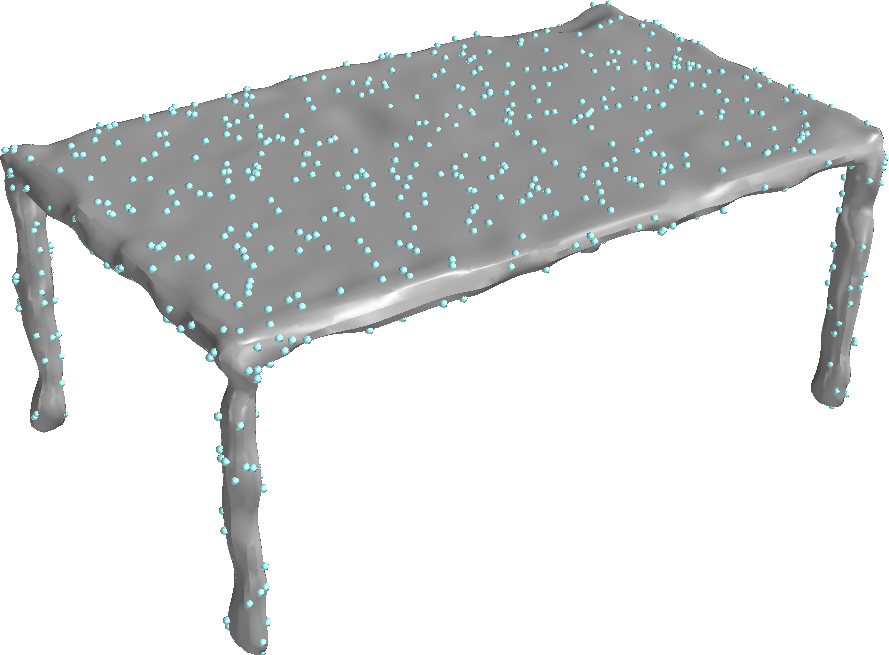}
    \endminipage\hfill
    \minipage{0.1427\linewidth}
    \centering
    \includegraphics[width=0.97\linewidth]{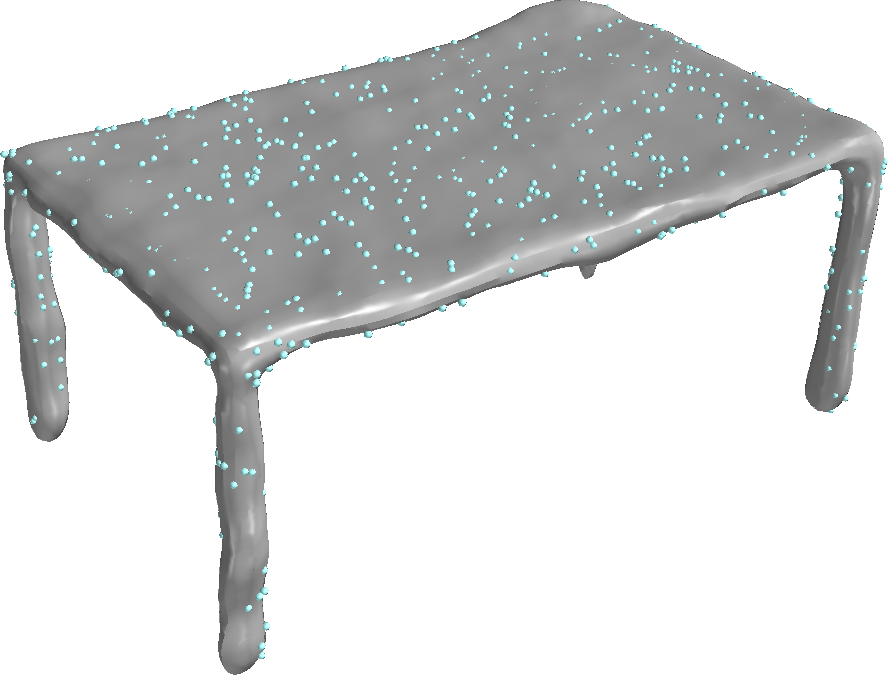}
    \endminipage\hfill
    \minipage{0.1427\linewidth}
    \centering
    \includegraphics[width=0.97\linewidth]{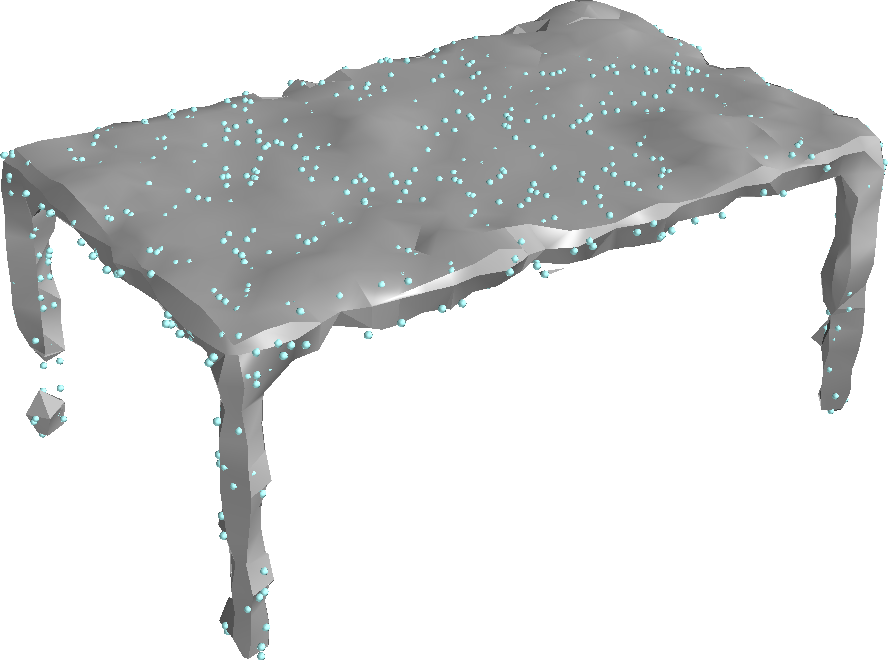}
    \endminipage\hfill
    \minipage{0.1427\linewidth}
    \centering
    \includegraphics[width=0.97\linewidth]{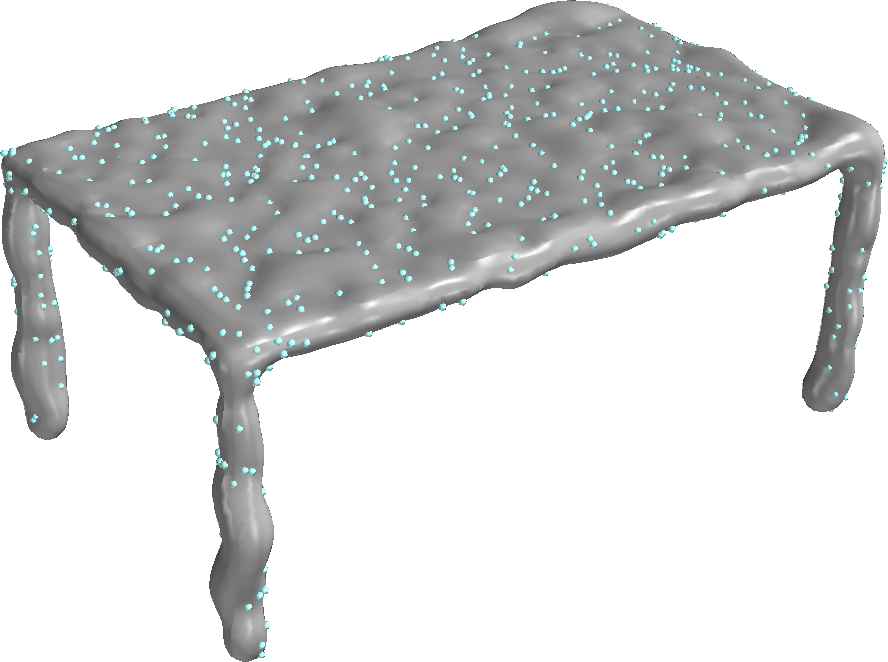}
    \endminipage\hfill
    \minipage{0.1427\linewidth}
    \centering
    \includegraphics[width=0.97\linewidth]{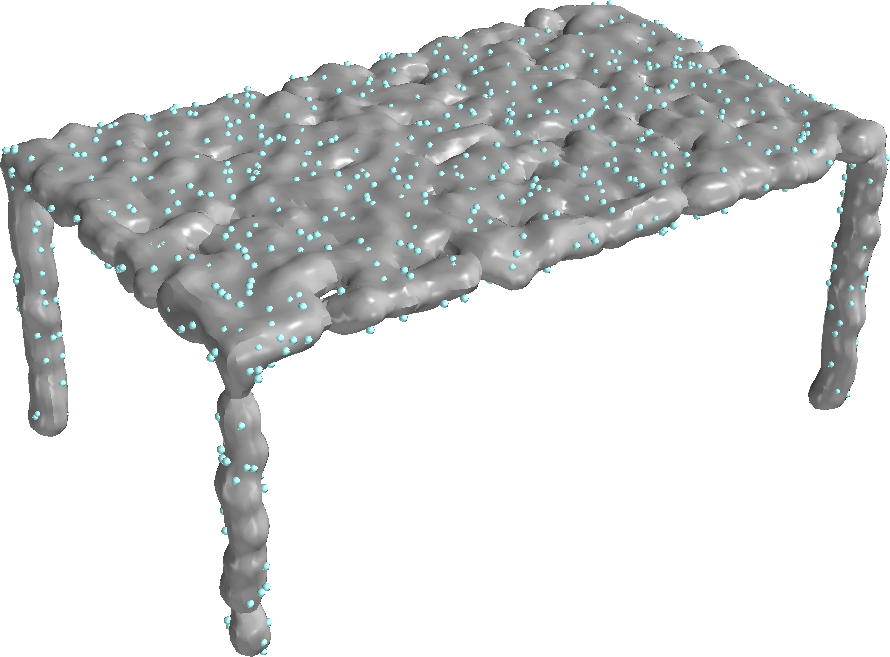}
    \endminipage\hfill
\\
    \minipage{0.1427\linewidth}
    \centering
    \includegraphics[width=0.97\linewidth]{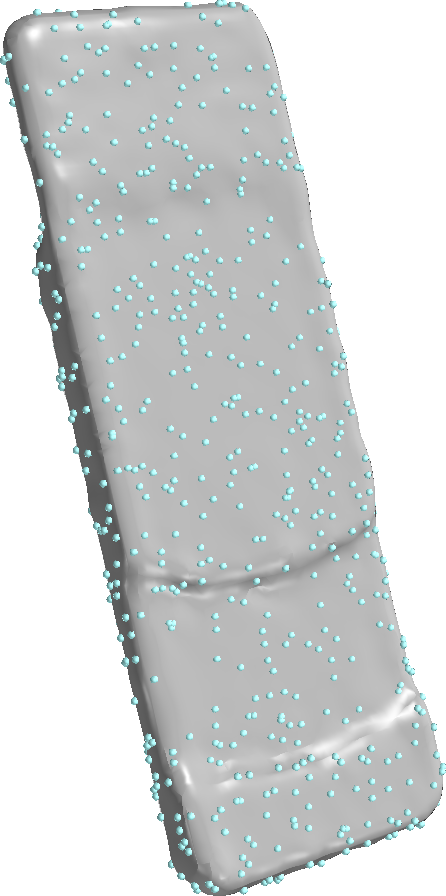}
    \endminipage\hfill
    \minipage{0.1427\linewidth}
    \centering
    \includegraphics[width=0.97\linewidth]{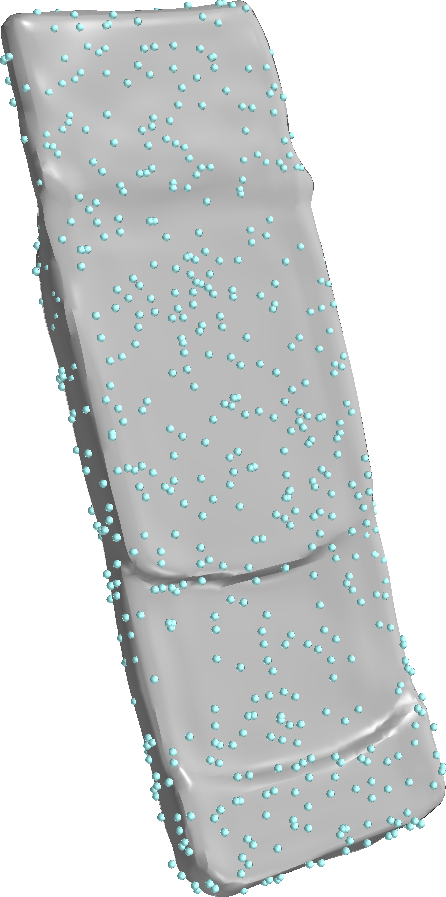}
    \endminipage\hfill
    \minipage{0.1427\linewidth}
    \centering
    \includegraphics[width=0.97\linewidth]{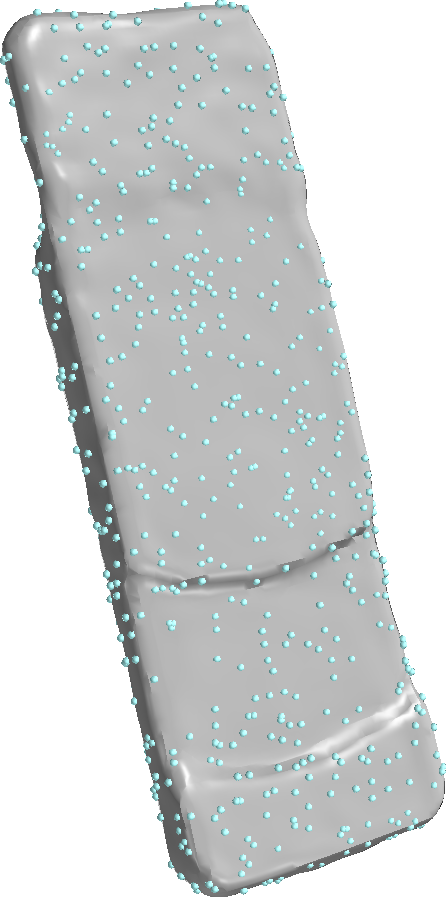}
    \endminipage\hfill
    \minipage{0.1427\linewidth}
    \centering
    \includegraphics[width=0.97\linewidth]{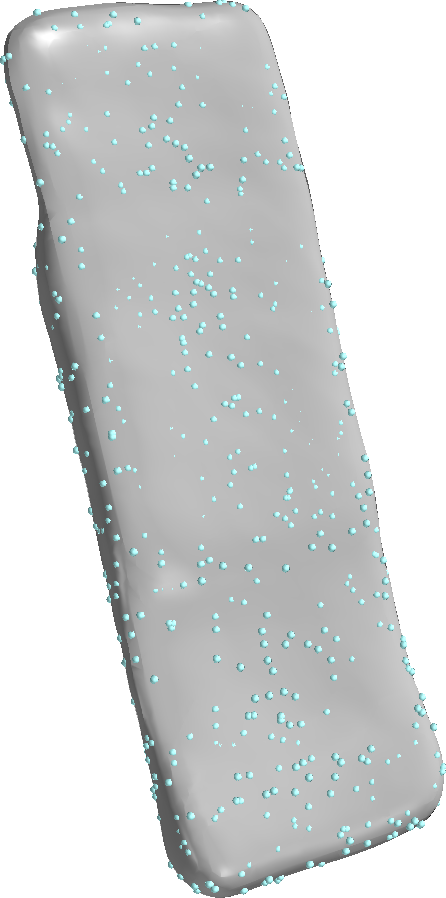}
    \endminipage\hfill
    \minipage{0.1427\linewidth}
    \centering
    \includegraphics[width=0.97\linewidth]{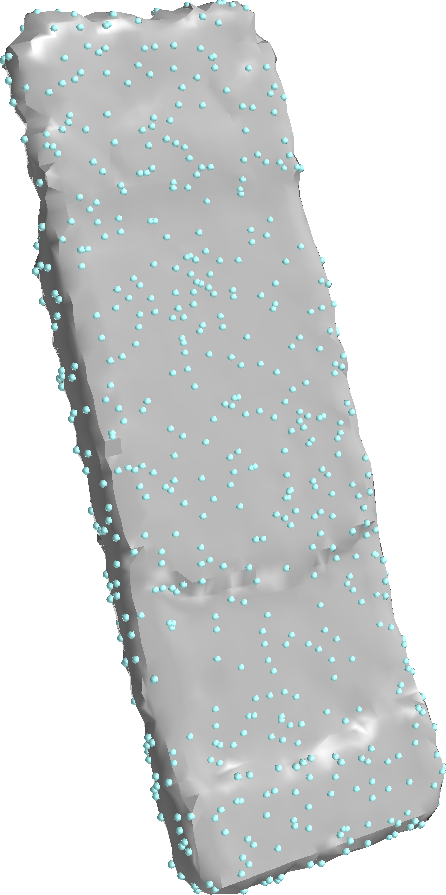}
    \endminipage\hfill
    \minipage{0.1427\linewidth}
    \centering
    \includegraphics[width=0.97\linewidth]{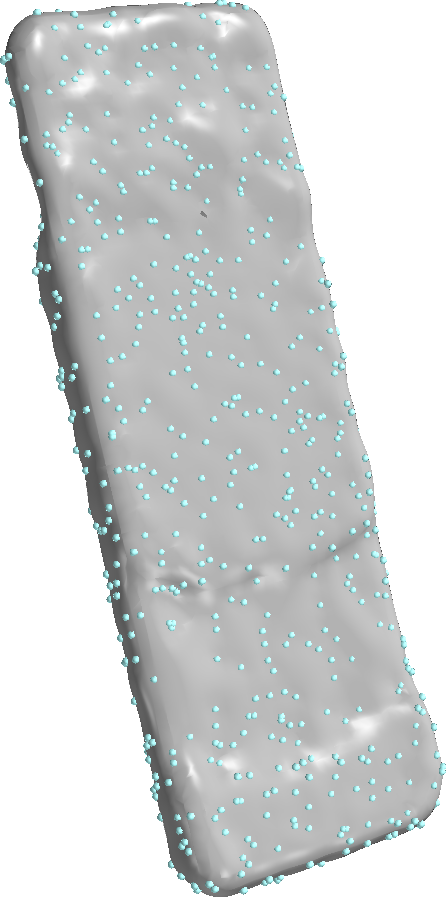}
    \endminipage\hfill
    \minipage{0.1427\linewidth}
    \centering
    \includegraphics[width=0.97\linewidth]{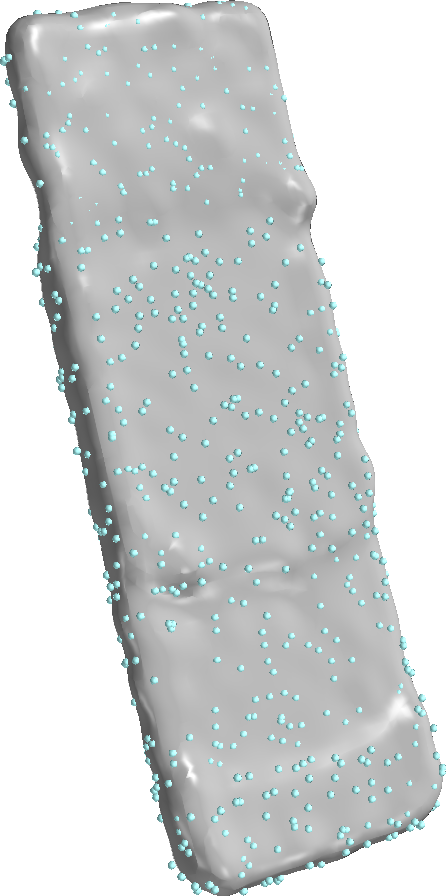}
    \endminipage\hfill
\\
    \minipage{0.1427\linewidth}
    \centering
    \includegraphics[width=0.97\linewidth]{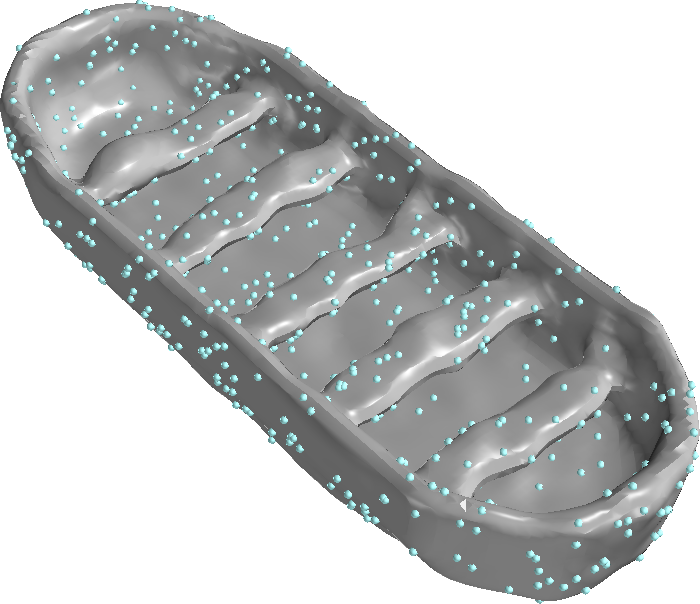}
    \endminipage\hfill
    \minipage{0.1427\linewidth}
    \centering
    \includegraphics[width=0.97\linewidth]{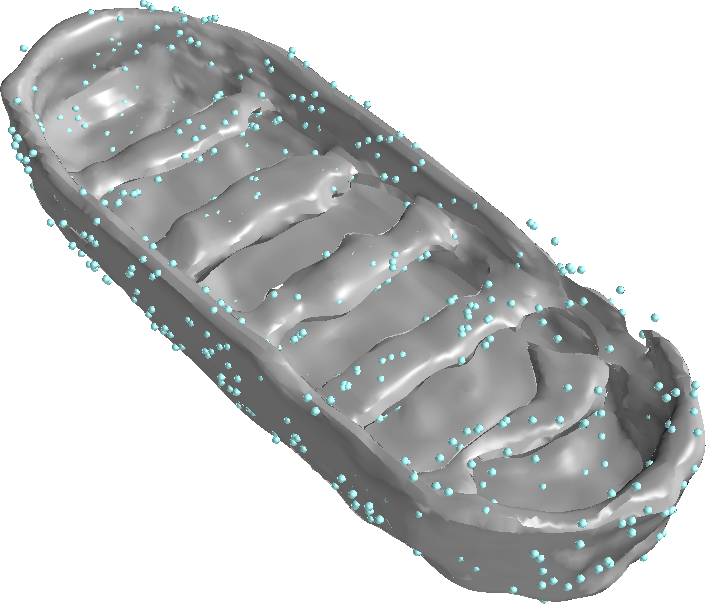}
    \endminipage\hfill
    \minipage{0.1427\linewidth}
    \centering
    \includegraphics[width=0.97\linewidth]{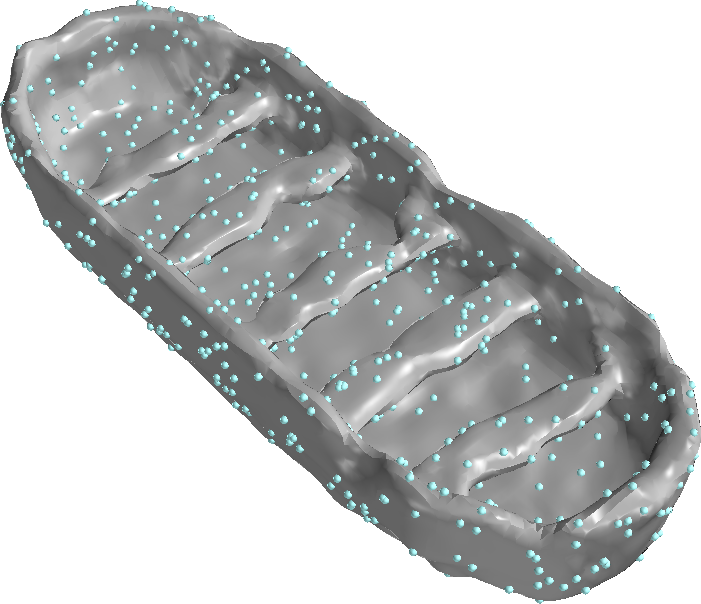}
    \endminipage\hfill
    \minipage{0.1427\linewidth}
    \centering
    \includegraphics[width=0.97\linewidth]{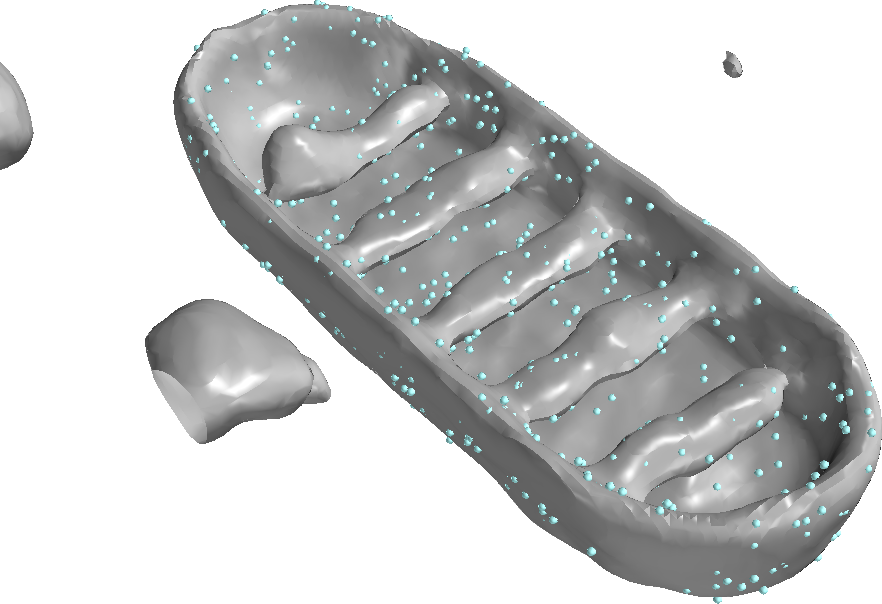}
    \endminipage\hfill
    \minipage{0.1427\linewidth}
    \centering
    \includegraphics[width=0.97\linewidth]{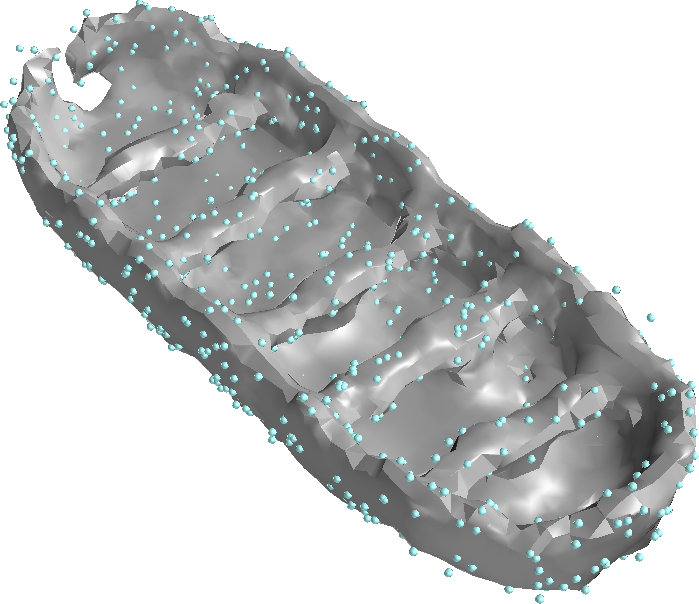}
    \endminipage\hfill
    \minipage{0.1427\linewidth}
    \centering
    \includegraphics[width=0.97\linewidth]{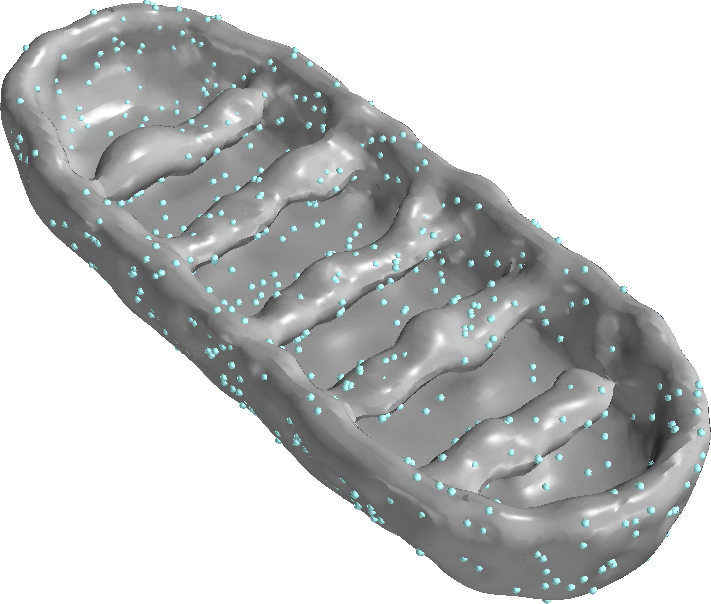}
    \endminipage\hfill
    \minipage{0.1427\linewidth}
    \centering
    \includegraphics[width=0.97\linewidth]{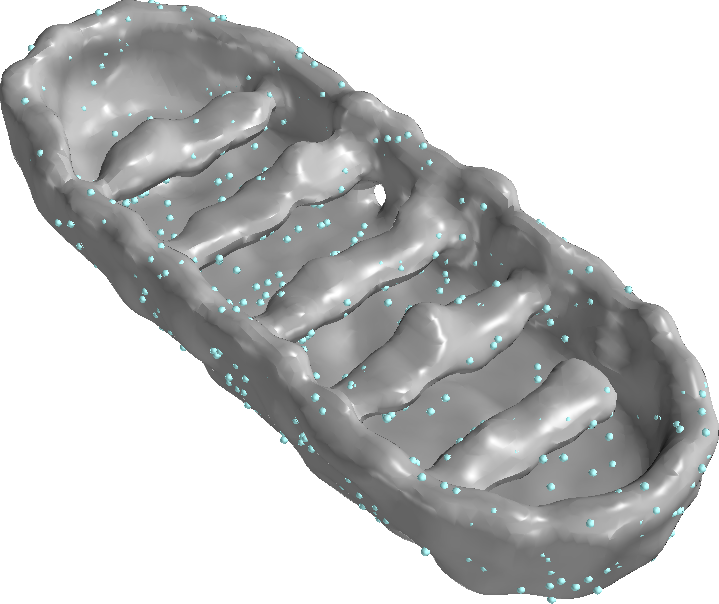}
    \endminipage\hfill
    \vspace{0.5em}
    \hfill
    \minipage{0.1427\linewidth}
    \centering \footnotesize \textbf{Ours}
    \endminipage\hfill
    \minipage{0.1427\linewidth}
    \centering \footnotesize IGR
    \endminipage\hfill
    \minipage{0.1427\linewidth}
    \centering \footnotesize SIREN
    \endminipage\hfill
    \minipage{0.1427\linewidth}
    \centering \footnotesize Fourier Feats
    \endminipage\hfill
    \minipage{0.1427\linewidth}
    \centering \footnotesize Poisson
    \endminipage\hfill
    \minipage{0.1427\linewidth}
    \centering \footnotesize Biharmonic
    \endminipage\hfill
    \minipage{0.1427\linewidth}
    \centering \footnotesize SVR
    \endminipage\hfill
    \vspace{0.5em}
    \caption{Comparisons between reconstruction techniques on Shapenet models. The blue points are the input points to the reconstruction algorithm.}\label{fig:shapenet2}
\end{figure}

\begin{figure}[H]
    \minipage{0.245\linewidth}
    \centering
    \includegraphics[width=0.85\linewidth]{figures/srb_models/anchor/figure_ours_anchor.png}
    \endminipage\hfill
    \minipage{0.245\linewidth}
    \centering
    \includegraphics[width=0.85\linewidth]{figures/srb_models/anchor/figure_igr_anchor.png}
    \endminipage\hfill
    \minipage{0.245\linewidth}
    \centering
    \includegraphics[width=0.85\linewidth]{figures/srb_models/anchor/figure_siren_anchor.png}
    \endminipage\hfill
    \minipage{0.245\linewidth}
    \centering
    \includegraphics[width=0.85\linewidth]{figures/srb_models/anchor/figure_ffk_anchor.png}
    \endminipage\hfill
    \\
    \minipage{0.245\linewidth}
    \centering
    \includegraphics[width=0.85\linewidth]{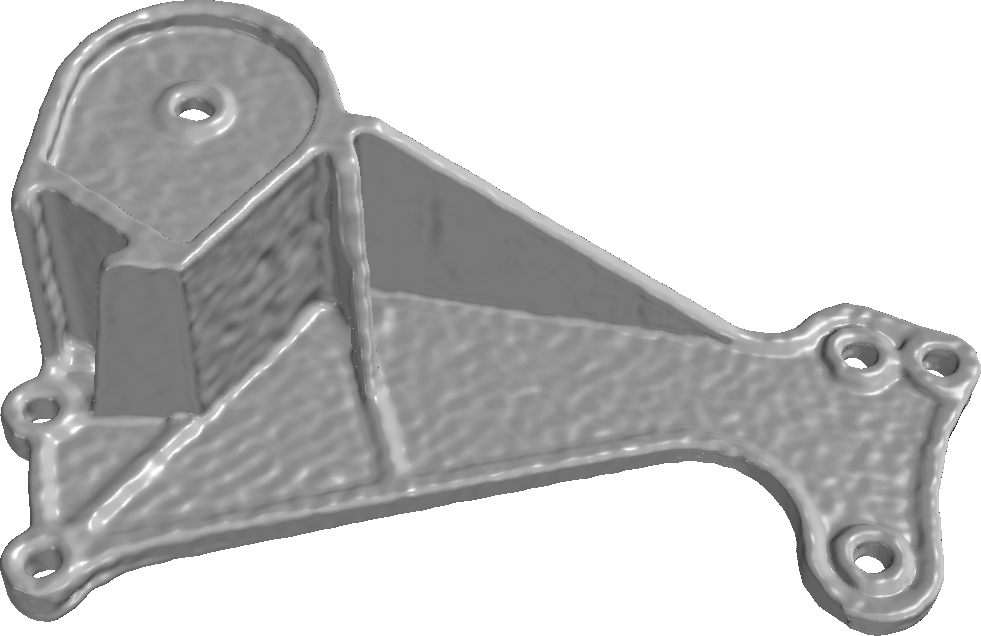}
    \endminipage\hfill
    \minipage{0.245\linewidth}
    \centering
    \includegraphics[width=0.85\linewidth]{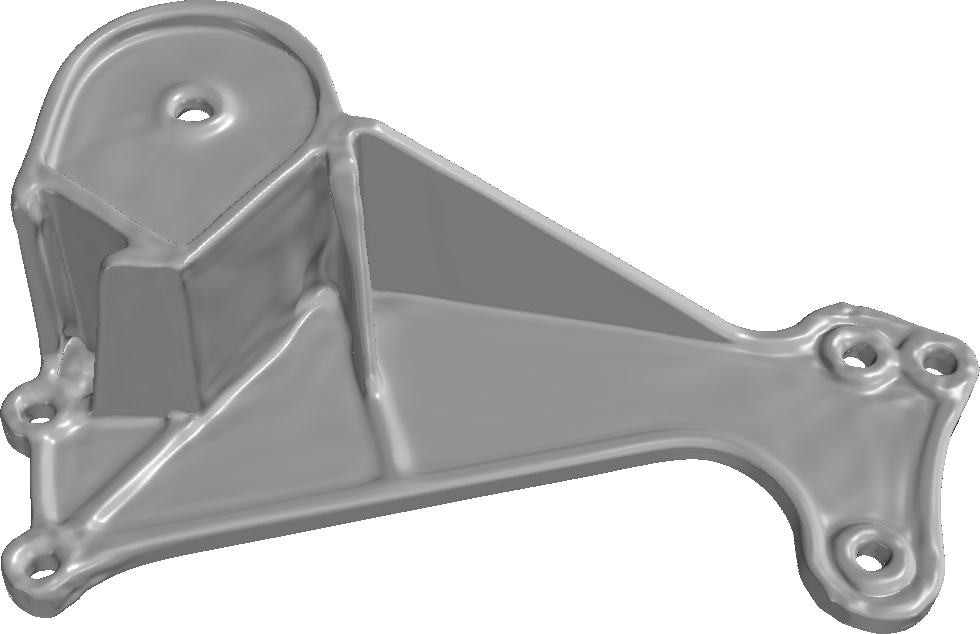}
    \endminipage\hfill
    \minipage{0.245\linewidth}
    \centering
    \includegraphics[width=0.85\linewidth]{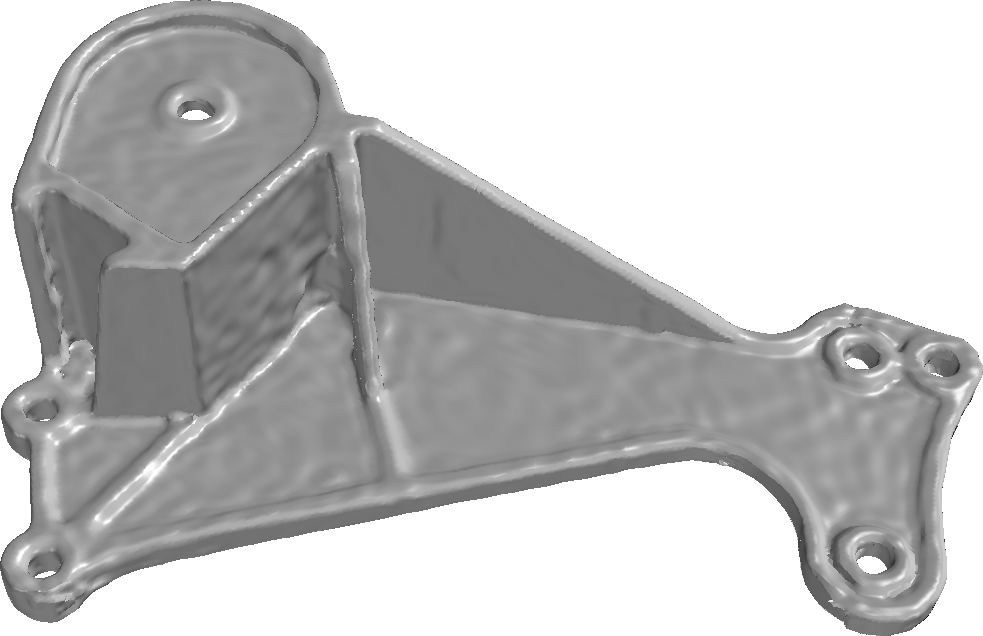}
    \endminipage\hfill
    \minipage{0.245\linewidth}
    \centering
    \includegraphics[width=0.85\linewidth]{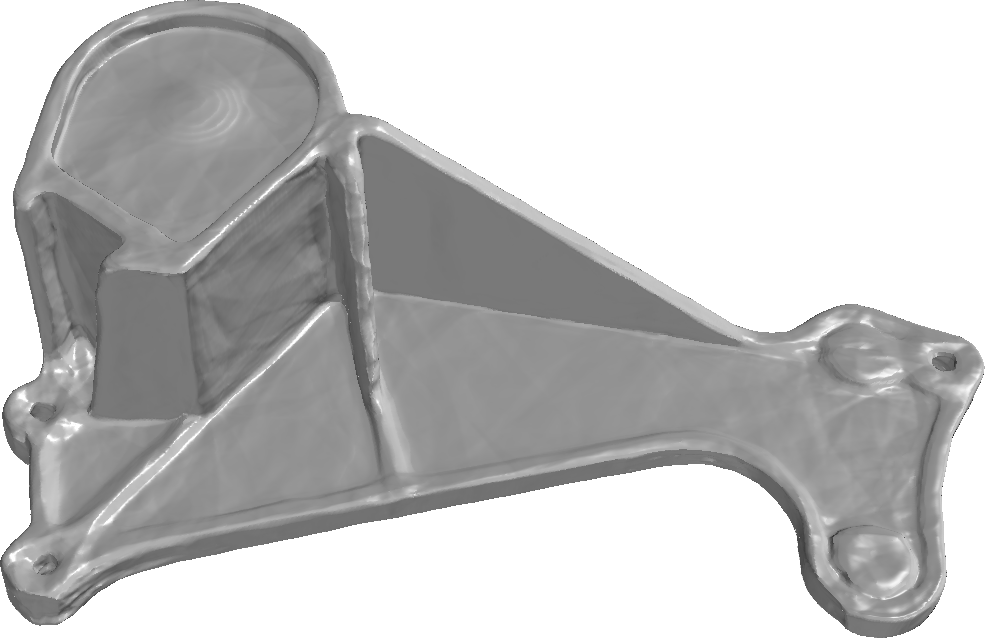}
    \endminipage\hfill
    \\
    \minipage{0.245\linewidth}
    \centering
    \includegraphics[width=0.85\linewidth]{figures/srb_models/dc/figure_ours_dc.png}
    \endminipage\hfill
    \minipage{0.245\linewidth}
    \centering
    \includegraphics[width=0.85\linewidth]{figures/srb_models/dc/figure_igr_dc.png}
    \endminipage\hfill
    \minipage{0.245\linewidth}
    \centering
    \includegraphics[width=0.85\linewidth]{figures/srb_models/dc/figure_siren_dc.png}
    \endminipage\hfill
    \minipage{0.245\linewidth}
    \centering
    \includegraphics[width=0.85\linewidth]{figures/srb_models/dc/figure_ffk_dc.png}
    \endminipage\hfill
    \\
    \minipage{0.245\linewidth}
    \centering
    \includegraphics[width=0.85\linewidth]{figures/srb_models/gargoyle/figure_ours_gargoyle.png}
    \endminipage\hfill
    \minipage{0.245\linewidth}
    \centering
    \includegraphics[width=0.85\linewidth]{figures/srb_models/gargoyle/figure_igr_gargoyle.png}
    \endminipage\hfill
    \minipage{0.245\linewidth}
    \centering
    \includegraphics[width=0.85\linewidth]{figures/srb_models/gargoyle/figure_siren_gargoyle.png}
    \endminipage\hfill
    \minipage{0.245\linewidth}
    \centering
    \includegraphics[width=0.85\linewidth]{figures/srb_models/gargoyle/figure_ffk_gargoyle.png}
    \endminipage\hfill
    \\
    \minipage{0.245\linewidth}
    \centering
    \includegraphics[width=0.65\linewidth]{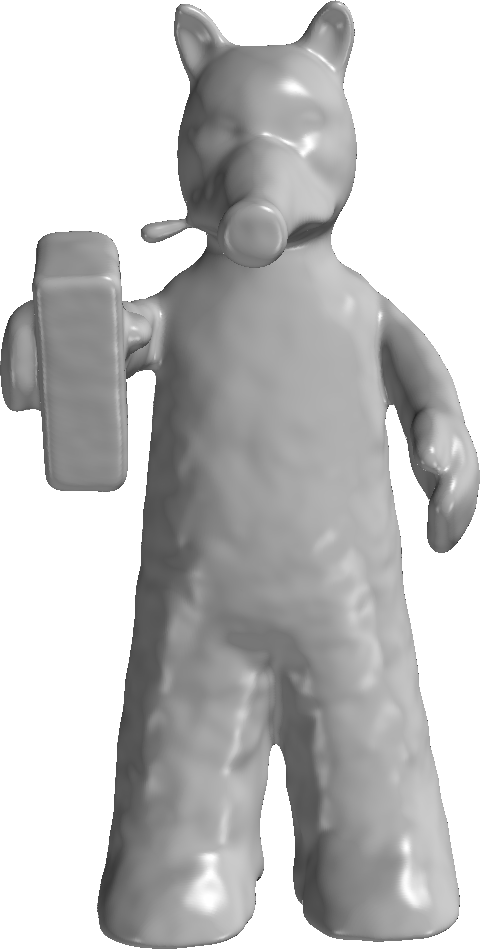}
    \endminipage\hfill
    \minipage{0.245\linewidth}
    \centering
    \includegraphics[width=0.65\linewidth]{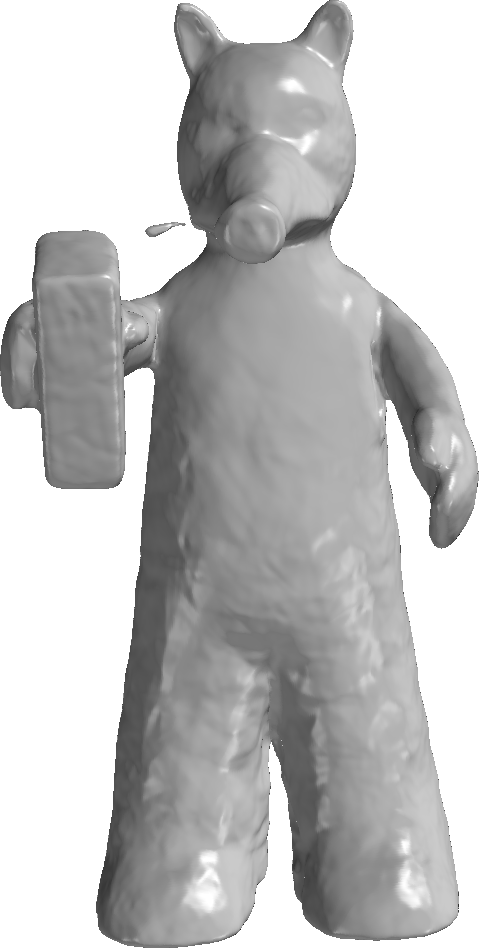}
    \endminipage\hfill
    \minipage{0.245\linewidth}
    \centering
    \includegraphics[width=0.65\linewidth]{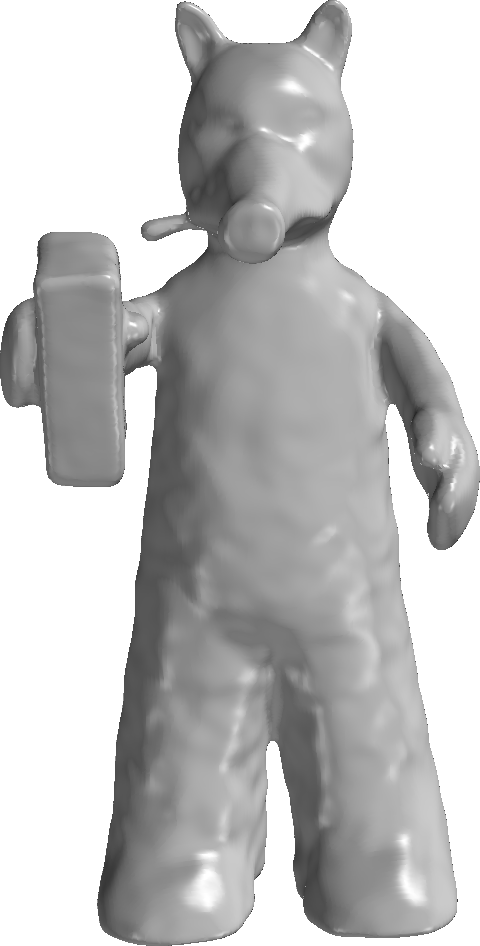}
    \endminipage\hfill
    \minipage{0.245\linewidth}
    \centering
    \includegraphics[width=0.65\linewidth]{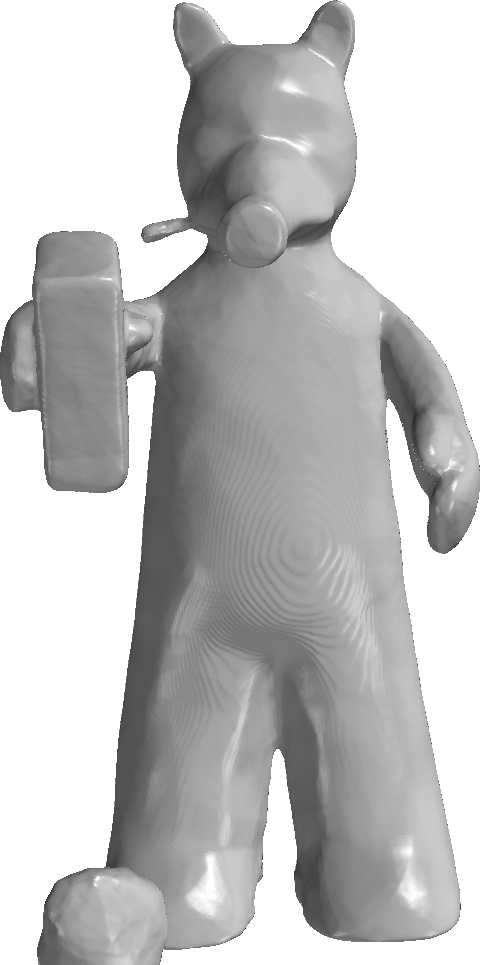}
    \endminipage\hfill
    \vspace{0.75em}
    \hfill
    \minipage{0.245\linewidth}
    \centering \footnotesize \textbf{Ours}
    \endminipage\hfill
    \minipage{0.245\linewidth}
    \centering \footnotesize IGR
    \endminipage\hfill
    \minipage{0.245\linewidth}
    \centering \footnotesize SIREN
    \endminipage\hfill
    \minipage{0.245\linewidth}
    \centering \footnotesize Fourier Feats
    \endminipage\hfill
    \vspace{0.5em}
    \caption{Comparisons between reconstruction techniques on the Surface Reconstruction Benchmark models. For techniques requiring parameter sweeps, we show the result with the lowest Chamfer Distance.}
    \label{fig:allfigs_srb}
\end{figure}

\vspace{-0.35em}
Figures \ref{fig:shapenet1} and \ref{fig:shapenet2} show at least one model reconstructed from each ShepeNet category using our method, Implicit Geometric Regularization (IGR) \cite{gropp2020implicit}, SIREN \cite{sitzmann2020implicit}, Fourier Feature Networks (FFN) \cite{tancik2020fourier}, Screened Poisson Surface Reconstruction \cite{kazhdan2013screened}, Biharmonic RBF \cite{carr2001reconstruction}, and Support Vector Regression (SVR) \cite{NIPS2004_2724}. Figure~\ref{fig:allfigs_srb} shows the reconstructions of all the models from the Surface Reconstruction Benchmark \cite{berger2013benchmark} using our method, IGR \cite{gropp2020implicit}, SIREN \cite{sitzmann2020implicit}, and Fourier Feature Networks \cite{tancik2020fourier}.


\subsection{Quantitative Results Per ShapeNet Class}\label{sec:per_category_stats}
Tables \ref{tbl:iou} show the per ShapeNet category IoU and Chamfer distance statistics for the benchmark described in Section~\ref{sec:benchmark}.
\begin{table}[H]
\begin{center}
\scalebox{.8}{
\begin{tiny}
\begin{tabular}{ |c||c|c|c||c|c|c||c|c|c||c|c|c||c|c|c||c|c|c||c|c|c| }

\multicolumn{19}{c}{\textbf{\normalsize Intersection over Union (IoU)}\vspace{0.5em}}\\
\hline
             & \multicolumn{3}{|c||}{SIREN \cite{sitzmann2020implicit}}  & \multicolumn{3}{|c||}{Fourier Feat. Nets \cite{tancik2020fourier}}   & \multicolumn{3}{|c||}{Biharmonic RBF \cite{carr2001reconstruction}}    & \multicolumn{3}{|c|}{SVR \cite{NIPS2004_2724}}  & \multicolumn{3}{|c||}{Screened Poisson \cite{kazhdan2013screened}}  & \multicolumn{3}{|c||}{IGR \cite{gropp2020implicit}}  & \multicolumn{3}{|c|}{\textbf{Ours}}\\
\hline
Class        & mean   & median & std          & mean   & median & std      & mean   & median & std                  & mean   & median & std              & mean  & median  & std           & mean  & median  & std        & mean            & median          & std\\
\hline
airplane     & 0.8045 & \textbf{0.9080} & 0.2696       & 0.6838 & 0.6926 & 0.0420   & 0.6690 & 0.7064 & 0.0809               & 0.6016 & 0.6257 & 0.0611           & 0.5954 & 0.6139 & 0.0581        & 0.7851 & 0.8193 & 0.0977     & \textbf{0.8165} & 0.8998 & 0.1551\\
bench        & 0.6109 & 0.7442 & 0.3258       & 0.6755 & 0.6583 & 0.1348   & 0.6237 & 0.6592 & 0.1682               & 0.6052 & 0.6282 & 0.1487           & 0.4728 & 0.4384 & 0.1213        & 0.5812 & 0.5923 & 0.2487     & \textbf{0.7872} & \textbf{0.8370} & 0.1236\\
cabinet      & 0.8706 & 0.9263 & 0.1621       & 0.8900 & 0.8984 & 0.0702   & 0.8913 & 0.9206 & 0.0926               & 0.7268 & 0.7142 & 0.0740           & 0.7301 & 0.7764 & 0.1095        & 0.8709 & 0.8857 & 0.0924     & \textbf{0.9274} & \textbf{0.9291} & 0.0422\\
car          & 0.8036 & 0.9241 & 0.2753       & 0.8173 & 0.8346 & 0.0847   & 0.8656 & 0.9020 & 0.1004               & 0.7903 & 0.8065 & 0.0773           & 0.6637 & 0.7192 & 0.1255        & 0.8026 & 0.8664 & 0.1300     & \textbf{0.8954} & \textbf{0.9288} & 0.0740\\
chair        & 0.8721 & 0.8807 & 0.0495       & 0.8250 & 0.8470 & 0.0618   & 0.8428 & 0.8667 & 0.0824               & 0.8127 & 0.8389 & 0.0839           & 0.5880 & 0.6046 & 0.1243        & 0.8049 & 0.8320 & 0.1022     & \textbf{0.8841} & \textbf{0.9034} & 0.0825\\
display      & 0.9014 & 0.9146 & 0.0440       & 0.8663 & 0.8755 & 0.0350   & 0.8470 & 0.8593 & 0.0738               & 0.8307 & 0.8407 & 0.0722           & 0.7027 & 0.7072 & 0.0800        & 0.8741 & 0.8917 & 0.0533     & \textbf{0.9309} & \textbf{0.9316} & 0.0251\\
lamp         & 0.8392 & 0.8995 & 0.2025       & 0.8192 & 0.8394 & 0.0773   & 0.8282 & 0.8706 & 0.1155               & 0.7728 & 0.7907 & 0.0963           & 0.5413 & 0.5786 & 0.1858        & 0.7865 & 0.8259 & 0.1318     & \textbf{0.9037} & \textbf{0.9178} & 0.0512\\
loudspeaker  & 0.8458 & 0.9618 & 0.2404       & 0.8993 & 0.9203 & 0.0709   & 0.9076 & 0.9621 & 0.1067               & 0.7098 & 0.6956 & 0.1193           & 0.7432 & 0.7848 & 0.1309        & 0.8867 & 0.9324 & 0.1017     & \textbf{0.9323} & \textbf{0.9627} & 0.0599\\
rifle        & 0.7329 & 0.9132 & 0.3662       & 0.7650 & 0.7693 & 0.0686   & 0.8350 & 0.8447 & 0.0845               & 0.7764 & 0.8013 & 0.0970           & 0.6803 & 0.6791 & 0.0532        & 0.8279 & 0.8267 & 0.0542     & \textbf{0.9299} & \textbf{0.9313} & 0.0215\\
sofa         & 0.9251 & 0.9411 & 0.0390       & 0.8805 & 0.8988 & 0.0554   & 0.9164 & 0.9377 & 0.0630               & 0.8694 & 0.8791 & 0.0740           & 0.7122 & 0.7326 & 0.0662        & 0.8891 & 0.9139 & 0.0708     & \textbf{0.9387} & \textbf{0.9473} & 0.0264\\
table        & 0.7280 & 0.8058 & 0.2089       & 0.7749 & 0.7786 & 0.0645   & 0.7096 & 0.7350 & 0.1150               & 0.6779 & 0.6596 & 0.1127           & 0.3720 & 0.3565 & 0.1334        & 0.6852 & 0.7260 & 0.2004     & \textbf{0.8414} & \textbf{0.8427} & 0.0534\\
telephone    & 0.9427 & 0.9514 & 0.0310       & 0.9111 & 0.9202 & 0.0483   & 0.9286 & 0.9433 & 0.0662               & 0.9195 & 0.9399 & 0.0638           & 0.7883 & 0.7990 & 0.0552        & 0.9148 & 0.9372 & 0.0639     & \textbf{0.9569} & \textbf{0.9625} & 0.0260\\
watercraft   & 0.8722 & \textbf{0.9279} & 0.1990       & 0.8180 & 0.8176 & 0.0545   & 0.8557 & 0.8756 & 0.0768               & 0.8197 & 0.8379 & 0.0742           & 0.6523 & 0.6793 & 0.0994        & 0.8146 & 0.8445 & 0.0931     & \textbf{0.9207} & 0.9231 & 0.0402\\
\hline
All Classes: & 0.8268 & 0.9097 & 0.2329       & 0.8218 & 0.8396 & 0.0989   & 0.8247 & 0.8642 & 0.1350               & 0.7625 & 0.7819 & 0.1300           & 0.6340 & 0.6728 & 0.1577        & 0.8102 & 0.8480 & 0.1519     & \textbf{0.8973} & \textbf{0.9230} & 0.0871\\
\hline

\multicolumn{19}{c}{\vspace{0.5em}}\\

\multicolumn{19}{c}{\textbf{\normalsize Chamfer Distance (CD)} \vspace{0.5em}}\\
\hline
             & \multicolumn{3}{|c||}{SIREN \cite{sitzmann2020implicit}}  & \multicolumn{3}{|c||}{Fourier Feat. Nets \cite{tancik2020fourier}}  & \multicolumn{3}{|c||}{Biharmonic RBF \cite{carr2001reconstruction}}    & \multicolumn{3}{|c|}{SVR \cite{NIPS2004_2724}}  & \multicolumn{3}{|c||}{Screened Poisson \cite{kazhdan2013screened}}  & \multicolumn{3}{|c||}{IGR \cite{gropp2020implicit}}  & \multicolumn{3}{|c|}{\textbf{Ours}}\\
\hline
Class        & mean   & median & std          & mean   & median & std         & mean   & median & std                    & mean    & median  & std            & mean & median & std             & mean & median & std             & mean             & median           & std\\
\hline
airplane     & 4.15e-5 & 3.87e-5 & 8.57e-6    & 1.04e-4 & 1.04e-4 & 1.02e-5   & 1.41e-4 & 1.31e-4 & 3.07e-5              & 1.61e-4 & 1.66e-4 & 1.01e-5        & 8.37e-5 & 8.76e-5 & 1.77e-5     & 3.04e-4 & 1.74e-4 & 3.47e-4     & \textbf{3.55e-5} & \textbf{3.44e-5} & 2.45e-6\\
bench        & 9.63e-5 & 8.12e-5 & 5.41e-5    & 9.45e-5 & 9.34e-5 & 2.10e-5   & 1.58e-4 & 1.73e-4 & 4.91e-5              & 1.34e-4 & 1.41e-4 & 3.05e-5        & 1.93e-4 & 1.93e-4 & 8.14e-5     & 4.48e-4 & 2.58e-4 & 4.33e-4     & \textbf{5.66e-5} & \textbf{4.82e-5} & 2.09e-5\\
cabinet      & 1.51e-4 & 6.69e-5 & 1.77e-4    & 1.01e-4 & 8.89e-5 & 3.86e-5   & 1.29e-4 & 1.19e-4 & 7.81e-5              & 1.34e-4 & 1.30e-4 & 8.27e-5        & 3.62e-4 & 2.91e-4 & 1.96e-4     & 1.56e-4 & 9.39e-5 & 1.23e-4     & \textbf{6.98e-5} & \textbf{4.69e-5} & 4.34e-5\\
car          & 1.39e-4 & 9.07e-5 & 1.03e-4    & 1.17e-4 & 1.10e-4 & 2.96e-5   & 1.17e-4 & 1.00e-4 & 5.23e-5              & 1.23e-4 & 1.18e-4 & 3.90e-5        & 2.27e-4 & 2.22e-4 & 8.04e-5     & 2.60e-4 & 2.82e-4 & 9.80e-5     & \textbf{8.21e-5} & \textbf{7.18e-5} & 3.60e-5\\
chair        & 1.05e-4 & 6.34e-5 & 1.18e-4    & 1.00e-4 & 8.85e-5 & 4.23e-5   & 1.10e-4 & 9.60e-5 & 6.11e-5              & 1.11e-4 & 1.01e-4 & 5.66e-5        & 2.82e-4 & 2.13e-4 & 1.84e-4     & 9.25e-4 & 9.88e-5 & 3.11e-3     & \textbf{5.62e-5} & \textbf{4.21e-5} & 4.32e-5\\
display      & 6.98e-5 & 5.68e-5 & 3.86e-5    & 7.60e-5 & 7.32e-5 & 2.36e-5   & 9.85e-5 & 9.06e-5 & 3.42e-5              & 9.86e-5 & 1.01e-4 & 3.35e-5        & 2.45e-4 & 2.13e-4 & 1.11e-4     & 9.99e-5 & 7.49e-5 & 8.44e-5     & \textbf{4.36e-5} & \textbf{3.99e-5} & 1.28e-5\\
lamp         & 6.26e-5 & 5.07e-5 & 3.35e-5    & 7.79e-5 & 7.97e-5 & 1.33e-5   & 9.26e-5 & 8.05e-5 & 4.81e-5              & 1.10e-4 & 1.02e-4 & 4.26e-5        & 1.61e-4 & 1.35e-4 & 1.02e-4     & 1.72e-3 & 1.28e-4 & 6.24e-3     & \textbf{4.19e-5} & \textbf{3.91e-5} & 1.00e-5\\
loudspeaker  & 2.77e-4 & 6.88e-5 & 5.54e-4    & 1.07e-4 & 9.22e-5 & 4.66e-5   & 1.36e-4 & 7.94e-5 & 1.16e-4              & 1.36e-4 & 9.91e-5 & 1.02e-4        & 4.29e-4 & 3.54e-4 & 2.68e-4     & 3.77e-3 & 1.15e-4 & 1.49e-2     & \textbf{8.41e-5} & \textbf{4.54e-5} & 7.54e-5\\
rifle        & 3.62e-5 & 3.50e-5 & 4.03e-6    & 7.17e-5 & 7.05e-5 & 1.43e-5   & 5.04e-5 & 5.00e-5 & 9.36e-6              & 7.20e-5 & 7.12e-5 & 1.46e-5        & 4.46e-5 & 3.11e-5 & 2.76e-5     & 9.62e-5 & 5.29e-5 & 1.25e-4     & \textbf{3.26e-5} & \textbf{3.15e-5} & 2.79e-6\\
sofa         & 7.88e-5 & 6.99e-5 & 3.90e-5    & 9.77e-5 & 9.59e-5 & 2.48e-5   & 8.42e-5 & 7.72e-5 & 3.17e-5              & 9.46e-5 & 9.02e-5 & 3.39e-5        & 2.72e-4 & 2.46e-4 & 1.02e-4     & 2.86e-4 & 1.02e-4 & 5.30e-4     & \textbf{5.11e-5} & \textbf{4.80e-5} & 1.24e-5\\
table        & 1.92e-4 & 8.32e-5 & 2.32e-4    & 1.02e-4 & 9.35e-5 & 3.62e-5   & 1.96e-4 & 1.91e-4 & 8.62e-5              & 1.66e-4 & 1.60e-4 & 6.73e-5        & 3.50e-4 & 2.53e-4 & 2.14e-4     & 3.40e-4 & 1.95e-4 & 3.33e-4     & \textbf{6.60e-5} & \textbf{4.88e-5} & 4.17e-5\\
telephone    & 3.88e-5 & 3.58e-5 & 9.64e-6    & 4.91e-5 & 4.63e-5 & 1.28e-5   & 4.90e-5 & 4.04e-5 & 2.93e-5              & 4.77e-5 & 4.06e-5 & 2.19e-5        & 1.28e-4 & 1.17e-4 & 3.36e-5     & 1.03e-4 & 4.43e-5 & 1.54e-4     & \textbf{3.34e-5} & \textbf{3.19e-5} & 3.60e-6\\
watercraft   & 5.57e-5 & 4.21e-5 & 2.95e-5    & 9.05e-5 & 8.46e-5 & 2.18e-5   & 7.81e-5 & 6.73e-5 & 3.89e-5              & 9.47e-5 & 8.56e-5 & 3.07e-5        & 1.07e-4 & 8.82e-5 & 6.12e-5     & 1.47e-4 & 1.12e-4 & 1.23e-4     & \textbf{4.41e-5} & \textbf{3.84e-5} & 1.42e-5\\
\hline
All Classes: & 1.03e-4 & 5.28e-5 & 1.93e-4    & 9.12e-5 & 8.65e-5 & 3.36e-5   & 1.11e-4 & 8.97e-5 & 7.06e-5              & 1.14e-4 & 1.04e-4 & 5.99e-5        & 2.22e-4 & 1.70e-4 & 1.76e-4     & 6.66e-4 & 1.07e-4 & 4.69e-3     & \textbf{5.36e-5} & \textbf{4.06e-5} & 3.64e-5\\
\hline
\end{tabular}
\end{tiny}
}
\end{center}
\caption{Quantitative comparison of the Intersection over Union (IoU) Distance and Chamfer Distance (CD) between SIREN \cite{sitzmann2020implicit}, Biharmonic RBF \cite{carr2001reconstruction}, Implicit Kernel SVR \cite{NIPS2004_2724}, Screened Poisson Surface Reconstructin \cite{kazhdan2013screened}, IGR \cite{gropp2020implicit} and our method over a subset (20 models per class) of the ShapeNet dataset. For each method we did a sweep over a range of parameters choosing the best result for each metric. We did no such tuning for our method.}
\label{tbl:iou}
\end{table}

\subsection{Per Model Performance Numbers}\label{sec:full_perf}
Table~\ref{tab:full_perf} shows the runtime and GPU usage required to reconstruct each model in the Surface Reconstruction Benchmark. For our model, we used 15k Nystr\"{o}m samples and a regularization of 1e-11. We do not report CPU memory usage since it is hard to profile exactly, however we observed that none of the methods used more than 4GiB of CPU memory. All timings were done on a machine with a single NVIDIA-V100 GPU with 16GiB of VRAM, 32GiB of CPU RAM, and and an 8 core Intel Xeon processor.
\begin{table}[H]
    \begin{scriptsize}
    \begin{center}
    \begin{tabular}{c | c | c | c }
                                   & Method                              & Runtime (seconds)  & GPU Memory \\
        \hline
        
        \multirow{5}{*}{Anchor}    & IGR \cite{gropp2020implicit}        & 1726.84            & 5093          \\
                                   & SIREN \cite{sitzmann2020implicit}   & 684.66             & 1855          \\
                                   & FFN \cite{tancik2020fourier}        & 385.90             & 4266          \\
                                   & Poisson \cite{kazhdan2013screened}  & \textbf{2.35}      & N.A.          \\
                                   & \textbf{Ours}                       & 13.22              & 5528          \\
        \hline
        
        \multirow{5}{*}{Daratech}  & IGR \cite{gropp2020implicit}        & 1239.43            & 5093          \\
                                   & SIREN \cite{sitzmann2020implicit}   & 500.42             & 1855          \\
                                   & FFN \cite{tancik2020fourier}        & 281.61             & 4266          \\
                                   & Poisson \cite{kazhdan2013screened}  & \textbf{1.83}      & N.A.          \\
                                   & \textbf{Ours}                       & 10.75              & 5528          \\
        \hline
        \multirow{5}{*}{DC}        & IGR \cite{gropp2020implicit}        & 1443.65            & 5093          \\
                                   & SIREN \cite{sitzmann2020implicit}   & 576.83             & 1545          \\
                                   & FFN \cite{tancik2020fourier}        & 325.10             & 3565          \\
                                   & Poisson \cite{kazhdan2013screened}  & \textbf{1.41}      & N.A.          \\ 
                                   & \textbf{Ours}                       & 12.31              & 5510          \\
        \hline
        \multirow{5}{*}{Gargoyle}  & IGR \cite{gropp2020implicit}        & 1646.44            & 5093          \\
                                   & SIREN \cite{sitzmann2020implicit}   & 766.27             & 2076          \\
                                   & FFN \cite{tancik2020fourier}        & 432.59             & 4778          \\
                                   & Poisson \cite{kazhdan2013screened}  & \textbf{1.44}      & N.A.          \\ 
                                   & \textbf{Ours}                       & 12.87              & 5145          \\
        \hline
        \multirow{5}{*}{Lord Quas} & IGR \cite{gropp2020implicit}        & 1013.33            & 5093          \\
                                   & SIREN \cite{sitzmann2020implicit}   & 466.82             & 1237          \\
                                   & FFN \cite{tancik2020fourier}        & 263.34             & 2870          \\
                                   & Poisson \cite{kazhdan2013screened}  & \textbf{1.21}      & N.A.          \\ 
                                   & \textbf{Ours}                       & 10.41              & 5281          \\
    \end{tabular}
    \end{center}
    \end{scriptsize}
    \caption{Runtime and GPU memory usage of different methods when reconstructing models from the Surface Reconstruction Benchmark \cite{berger2013benchmark}.}
    \label{tab:full_perf}
\end{table}

\subsection{Empirical versus Analytical Kernel}\label{sec:empirical_vs_analytical} 
\begin{figure}[H]
    \minipage{0.5\textwidth}
    \centering
    \includegraphics[width=0.99\textwidth]{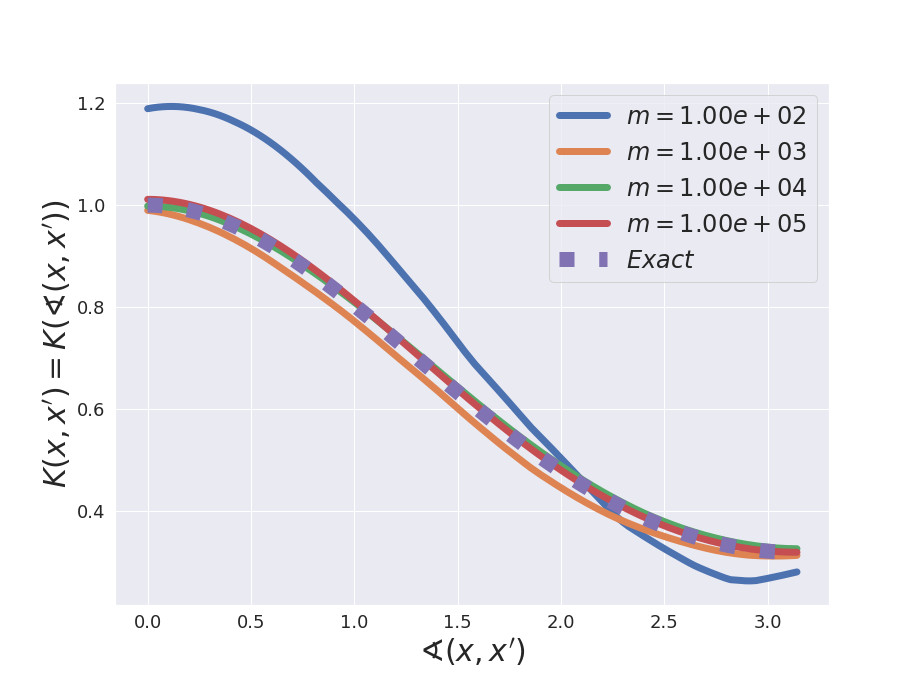}
    \endminipage\hfill
    \minipage{0.5\textwidth}
    \centering
    \includegraphics[width=0.99\textwidth]{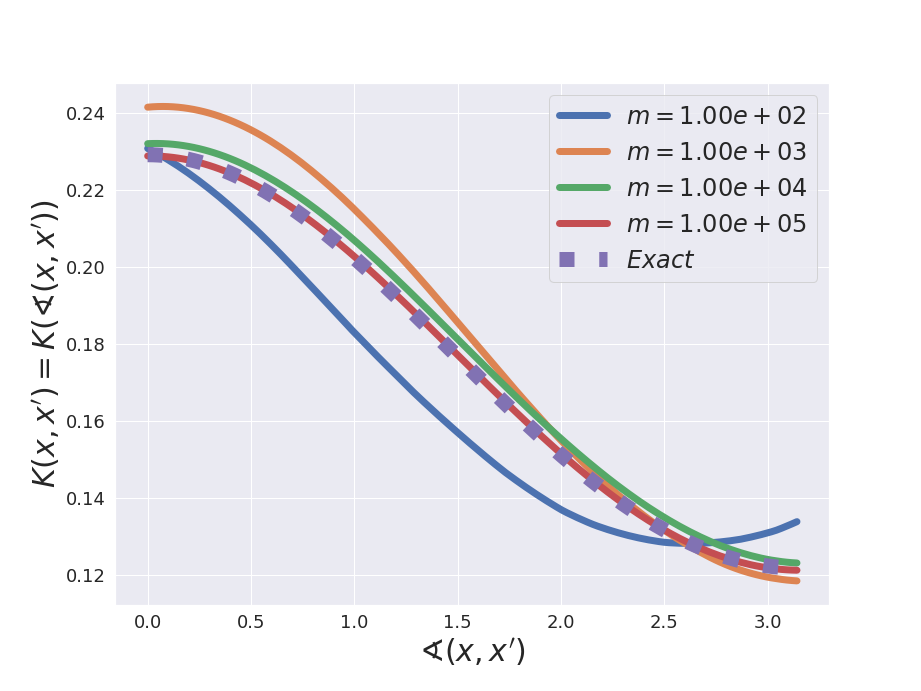}
    \endminipage\hfill
    \vspace{0.25em}
    \parbox{.49\textwidth}{\centering \footnotesize Gaussian Kernel}
    \parbox{.49\textwidth}{\centering \footnotesize Uniform Kernel}
    \vspace{0.5em}
    \caption{Convergence of the empirical kernels to the exact ones. Both the Gaussian (left) and Uniform (right) kernels are rotation invariant and thus depend only on the angle $\sphericalangle(x, x')$ between $x$ and $x'$. This plot shows the value of the kernel as a function of this angle from 0 to $\pi$ (we show here only the scalar term of the kernel $\mathbb E_{(a,b)}[ax+b]_+[ax'+b]_+$).}
    \label{fig:kernel_convergence}
\end{figure}

\begin{figure}[H]
    \minipage{0.25\textwidth}
    \centering
    \includegraphics[width=0.99\textwidth]{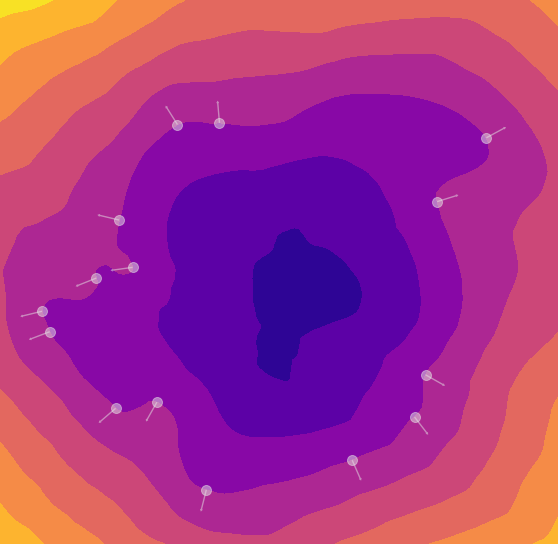}
    \endminipage\hfill
    \minipage{0.25\textwidth}
    \centering
    \includegraphics[width=0.99\textwidth]{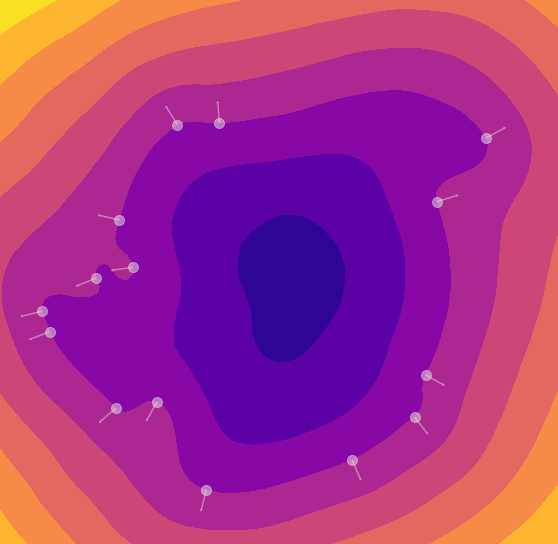}
    \endminipage\hfill
    \minipage{0.25\textwidth}
    \centering
    \includegraphics[width=0.99\textwidth]{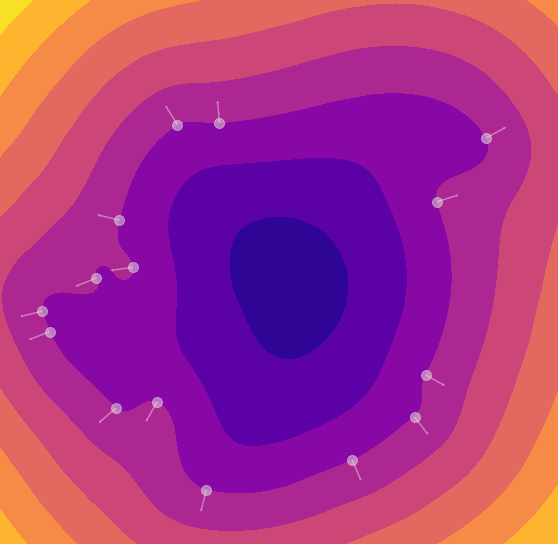}
    \endminipage\hfill
    \minipage{0.25\textwidth}
    \centering
    \includegraphics[width=0.99\textwidth]{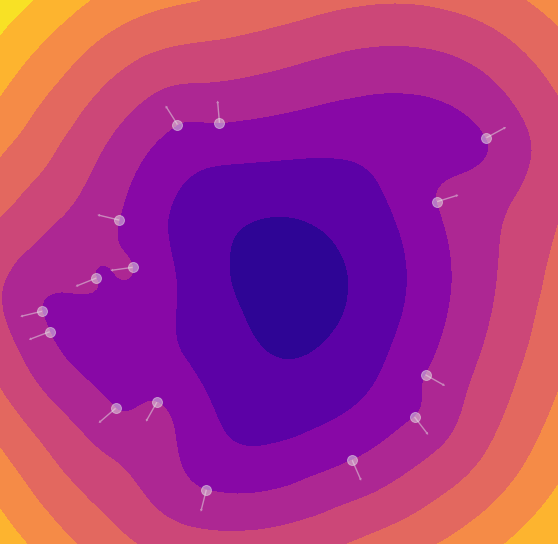}
    \endminipage\hfill
    \vspace{0.5em}
    \parbox{.245\textwidth}{\centering \footnotesize $m$ = 1k}
    \parbox{.245\textwidth}{\centering \footnotesize $m$ = 10k}
    \parbox{.245\textwidth}{\centering \footnotesize $m$ = 50k}
    \parbox{.245\textwidth}{\centering \footnotesize Analytical}
    \vspace{0.5em}
    \caption{The effect of using an approximate kernel with $m$ neurons to do reconstruction. Increasing $m$ makes the approximation closer to the analytical version.}
    \label{fig:empirical_vs_analytical}
\end{figure}

Figure~\ref{fig:empirical_vs_analytical} compares results using the empirical kernel with $m$ neurons and using the analytical kernel. Figure~\ref{fig:kernel_convergence} shows the convergence of the empirical Kernel to the analytic one as the number $m$ of neurons grows.

\subsection{Quantitative Comparison between Gaussian and Uniform Kernels}\label{sec:us_vs_us}
Table \ref{tbl:iou_chamfer:us_vs_us} shows a quantitative comparison between Neural Splines using the Gaussian initialization \eqref{eq:gaussian_initialization} and the Uniform initialization \eqref{eq:uniform_initialization} on the benchmark described in Section~\eqref{sec:benchmark}. The results in both cases are very close to each other in both Chamfer distance and in IoU.
\begin{table}[H]
\begin{center}
\begin{tabular}{ |c||c|c|c||c|c|c| }
\multicolumn{7}{c}{\textbf{\normalsize Intersection over Union (IoU)}\vspace{0.5em}}\\
\hline
 & \multicolumn{3}{|c||}{Neural Spline (Uniform)} & \multicolumn{3}{|c|}{Neural Spline (Gaussian)}\\
\hline
Class & mean & median & std & mean & median & std\\
\hline
car & 0.9082 & 0.9399 & 0.0747 & 0.9084 & 0.9392 & 0.0743\\
chair & 0.9056 & 0.9369 & 0.1062 & 0.9131 & 0.9357 & 0.0852\\
airplane & 0.7773 & 0.8796 & 0.1961 & 0.8388 & 0.9159 & 0.1531\\
display & 0.9533 & 0.9549 & 0.0188 & 0.9523 & 0.9535 & 0.0193\\
table & 0.8968 & 0.9011 & 0.0500 & 0.8952 & 0.8980 & 0.0508\\
rifle & 0.9489 & 0.9491 & 0.0169 & 0.9488 & 0.9498 & 0.0185\\
cabinet & 0.9478 & 0.9467 & 0.0377 & 0.9477 & 0.9475 & 0.0377\\
loudspeaker & 0.9507 & 0.9768 & 0.0518 & 0.9500 & 0.9768 & 0.0531\\
telephone & 0.9746 & 0.9772 & 0.0202 & 0.9741 & 0.9774 & 0.0209\\
bench & 0.8160 & 0.8900 & 0.1365 & 0.8195 & 0.8902 & 0.1296\\
sofa & 0.9565 & 0.9644 & 0.0267 & 0.9584 & 0.9643 & 0.0221\\
watercraft & 0.9340 & 0.9380 & 0.0462 & 0.9360 & 0.9386 & 0.0404\\
lamp & 0.9467 & 0.9470 & 0.0306 & 0.9457 & 0.9459 & 0.0319\\
\hline
All Classes: & 0.9167 & 0.9438 & 0.0985 & 0.9221 & 0.9441 & 0.0834\\
\hline
\multicolumn{7}{c}{\vspace{0.5em}}\\
\multicolumn{7}{c}{\textbf{\normalsize Chamfer Distance (CD)} \vspace{0.5em}}\\
\hline
 & \multicolumn{3}{|c||}{Neural Spline (Uniform)} & \multicolumn{3}{|c|}{Neural Spline (Gaussian)}\\
\hline
Class & mean & median & std & mean & median & std\\
\hline
car & 8.21e-05 & 7.23e-05 & 3.59e-05 & 8.21e-05 & 7.18e-05 & 3.60e-05\\
chair & 5.52e-05 & 4.20e-05 & 4.01e-05 & 5.62e-05 & 4.21e-05 & 4.32e-05\\
airplane & 3.55e-05 & 3.45e-05 & 2.45e-06 & 3.55e-05 & 3.44e-05 & 2.45e-06\\
display & 4.31e-05 & 3.95e-05 & 1.24e-05 & 4.36e-05 & 3.99e-05 & 1.28e-05\\
table & 6.44e-05 & 4.78e-05 & 4.01e-05 & 6.60e-05 & 4.88e-05 & 4.17e-05\\
rifle & 3.27e-05 & 3.16e-05 & 2.81e-06 & 3.26e-05 & 3.15e-05 & 2.79e-06\\
cabinet & 6.93e-05 & 4.67e-05 & 4.30e-05 & 6.98e-05 & 4.69e-05 & 4.34e-05\\
loudspeaker & 8.27e-05 & 4.58e-05 & 7.24e-05 & 8.41e-05 & 4.54e-05 & 7.54e-05\\
telephone & 3.33e-05 & 3.18e-05 & 3.42e-06 & 3.34e-05 & 3.19e-05 & 3.60e-06\\
bench & 5.62e-05 & 4.74e-05 & 2.05e-05 & 5.66e-05 & 4.82e-05 & 2.09e-05\\
sofa & 5.08e-05 & 4.81e-05 & 1.22e-05 & 5.11e-05 & 4.80e-05 & 1.24e-05\\
watercraft & 4.41e-05 & 3.84e-05 & 1.42e-05 & 4.41e-05 & 3.84e-05 & 1.42e-05\\
lamp & 4.15e-05 & 3.89e-05 & 9.61e-06 & 4.19e-05 & 3.91e-05 & 1.00e-05\\
\hline
All Classes: & 5.32e-05 & 4.07e-05 & 3.53e-05 & 5.36e-05 & 4.06e-05 & 3.64e-05\\
\hline
\end{tabular}
\end{center}
\caption{Comparison of IoU and Chamfer Distance between Neural Splines with Gaussian and Uniform kernels on the benchmark described in Section~\ref{sec:benchmark}. We remark that both kernels yield extremely close results.}\label{tbl:iou_chamfer:us_vs_us}
\end{table}

\section{Derivation of the Infinite Width Kernels}\label{sec:analytic_expressions}

\subsection{Uniform Initialization}
We derive an explicit expression for the kernel $K_\infty$ in \eqref{eq:kernel_infinite} in the case of uniform initialization \eqref{eq:uniform_initialization}. We first prove the following Lemma which we will use in our calculations.

\begin{lemma}\label{lemma:homogeneous_integrals} Assume that $\mathcal F: \RR \times \RR \rightarrow \RR$ is such that $\mathcal F(k s,k t) = k^r \mathcal F(s,t)$ for some $r$ and any $k \ge 0$. For any $x,x' \in \RR^d$ we have that
\begin{equation}\label{eq:integrals}
\begin{aligned}
    &\int_{a \in \mathbb S^{d-1}} \mathcal F (a^T x, a^T x') d \Omega = \eta_{d,r} \cdot \frac{r!!}{(d+r-2)!!} \cdot F_1\\
    &\int_{a \in \mathbb S^{d-1}} a \mathcal F (a^T x, a^T x') d \Omega = \eta_{d,r} \cdot \frac{(r+1)!!}{(d+r-1)!!} \cdot Q^T \begin{bmatrix}0_{d-2}\\
    F_{\cos}\\
    F_{\sin}\\
    \end{bmatrix}\\
    &\int_{a \in \mathbb S^{d-1}} aa^T \mathcal F (a^T x, a^T x') d \Omega = \eta_{d,r} \cdot \frac{(r+2)!!}{(d+r)!!} \cdot
    Q^T \begin{bmatrix}
    \frac{1}{r+2} Id_{d-2} F  & 0 & 0 \\
    0 & F_{\cos^2} & F_{\sin\cos} \\
    0 & F_{\sin\cos} & F_{\sin^2}  
    \end{bmatrix} Q
\end{aligned}
\end{equation}
where $F_g=\int_{0}^{2\pi} g(\psi) \mathcal F(\|x\| \cos(\psi),\|x'\| \cos(\psi - \alpha)) d \psi$ (for $g=1,\cos, \cos^2$, \emph{etc.}), $Q \in SO(d)$ is such that $Qx = (0,\ldots,\|x\|,0)^T$,$Qx' = (0,\ldots,\|x'\|\cos(\alpha),\|x'\|\sin(\alpha))^T$  and
\[
\eta_{d,r} = \begin{cases} 2^{\lceil\frac{d-2}{2}\rceil}\pi^{\lfloor\frac{d-2}{2}\rfloor} &\mbox{if } r \mbox{ is even,}\\
2^{\lfloor\frac{d-2}{2}\rfloor}\pi^{\lceil\frac{d-2}{2}\rceil} &\mbox{if } r \mbox{ is odd.}\\
\end{cases}
\]
\end{lemma}

\begin{proof} If $Q \in SO(d)$ then
by change of variables $a = Q^T  \tilde a$ we have that
\[
\begin{aligned}
\int_{a \in \mathbb S^{d-1}} aa^T \mathcal F(a^T x, a^T x') d\Omega &= \int_{\tilde a \in \mathbb S^{d-1}} Q^T\tilde a \tilde a^T Q \mathcal F(\tilde a^T Q x,  \tilde a^T Q x') d\Omega \\
&= Q^T \left(\int_{\tilde a \in \mathbb S^{d-1}} \tilde a \tilde a^T \mathcal F(\tilde a^T \tilde x,  \tilde a^T \tilde x') d\Omega\right) Q,
\end{aligned}
\]
where $\tilde x = Q x$. Without loss of generality we thus assume that $x = (0,\ldots,\|x\|,0)^T$ and $x' = (0,\ldots,\|x'\|\cos(\alpha),\|x'\|\sin(\alpha))^T$ where $\alpha = \arccos(x^T x'/\|x\|\|x'\|) \in [0,\pi]$. We now adopt hyperspherical coordinates $(\theta_1, \ldots \theta_{d-2}, \psi)$ where $\theta_i \in [0,\pi]$ and $\psi \in [0,2\pi]$. The conversion between cartesian and spherical coordinates is given by:
\begin{align*}
    &a_1 = \cos(\theta_1)\\ 
    &a_2 = \sin(\theta_1) \cos(\theta_2)\\
    &\quad \vdots\\
    &a_{d-1} = \sin(\theta_1) \ldots \sin(\theta_{d-2}) \cos(\psi)\\
    &a_{d} = \sin(\theta_1) \ldots \sin(\theta_{d-2}) \sin(\psi)
\end{align*}
We also have that
\begin{equation*}
\begin{aligned}
d\Omega &= \sin^{d-2}(\theta_1) \sin^{d-3}(\theta_2) \ldots \sin(\theta_{d-2}) d\theta_1 d\theta_2 \ldots d\psi\\
a \cdot x  &= a_{d-2}\nonumber\\
               &= ||x||\sin(\theta_1) \ldots \sin(\theta_{d-2}) \cos(\psi)\\
a \cdot x' &= a_{d-2}\nonumber\\
               &= a_{d-2}||x'||\cos(\alpha) + a_{d-1} ||x'||\sin(\alpha)\nonumber\\
               &= ||x'||\sin(\theta_1) \ldots \sin(\theta_{d-2}) (\cos(\psi) \cos(\alpha) + \sin(\psi) \sin(\alpha))\nonumber\\
               &= ||x'||\sin(\theta_1) \ldots \sin(\theta_{d-2}) \cos(\psi - \alpha)
\end{aligned}
\end{equation*}
We now consider the integral $\int  aa^T \mathcal F(a^Tx, a^T x')d\Omega$. For any two indices $i\le j \le d-2$, we have
\begin{equation}\label{eq:integral_aaT}
\begin{small}
\begin{aligned}
&\int_{a \in \mathbb S^{d-1}} a_i a_j \mathcal F(a\cdot x, a \cdot x') d \Omega \\
&=\int_{0}^{2\pi} \ldots \int_0^{\pi}  \sin^{d-2+r}(\theta_1) \ldots \sin^{1+r}(\theta_{d-2}) a_i a_j \mathcal F(\|x\| \cos \psi, \|x'\| \cos(\psi - \alpha)) d \psi  d\theta_1,\ldots d \theta_{d-2} \\
&= \int_0^{\pi} \sin^{d-2+r+2}(\theta_1)d\theta_1 \ldots \int_0^\pi \sin^{(d-1-i)+r+1}(\theta_i) \cos(\theta_i)d\theta_i \ldots \int_{0}^\pi \sin^{(d-1-j)+r}(\theta_j) \cos(\theta_j) \theta_j \\& 
\quad \ldots \int_0^\pi \sin^{(d-1-(j+1))+r} (\theta_{j+1}) d\theta_{j+1} \ldots \int_{0}^{2\pi} \mathcal F(\|x\| \cos \psi, \|x'\| \cos(\psi - \alpha)) d \psi.
\end{aligned}
\end{small}
\end{equation}
This is now a product of $d-1$ one-dimensional integrals. Since $\int_0^\pi \sin^s = 0$ if $s$ is odd, we have that the integral~\eqref{eq:integral_aaT} vanishes if
$i \ne j$. If instead $i=j$, we use the fact that
\[
\int_0^{\pi} \sin^s(t) \cos^2(t) dt = \frac{(s-1)!!}{(s+2)!!}\begin{cases}\pi & \text{if }$s$\text{ is even}\\2 &\text{if }$s$\text{ is odd}\end{cases}
\]
and we deduce that
\[
\begin{aligned}
&\int_{a \in \mathbb S^{d-1}} a_i a_i \mathcal F(a\cdot x, a \cdot x') d \Omega\\
&=\eta_{d,r} \frac{(d-i+r-1)!!}{(d+r)!!} \frac{(d-i+r-2)!!}{(d-i+r+1)!!} \frac{r!!}{(d-i+r-2)!!} \int_{0}^{2\pi} \mathcal F(\|x\| \cos \psi, \|x'\| \cos(\psi - \alpha)) d \psi.\\
&= \eta_{d,r}\frac{r!!}{(d+r)!!}\int_{0}^{2\pi}\mathcal F(\|x\| \cos \psi, \|x'\| \cos(\psi - \alpha)) d \psi.
\end{aligned}
\]
This proves the diagonal part in our expression for $\int  aa^T \mathcal F(a^Tx, a^T x') d\Omega$. All remaining terms as well as the two integrals $\int  a\mathcal F(a^Tx, a^T x')d\Omega$ and $\int   \mathcal F(a^Tx, a^T x')d\Omega$ follow from very similar (and slightly simpler) calculations.
\end{proof}

We now apply Lemma~\ref{lemma:homogeneous_integrals} to compute the kernel $K_\infty$ with the uniform initialization~\eqref{eq:uniform_initialization}. 

\begin{proposition} If $a \sim \mathcal U(\mathbb S^{d-1})$ and $b \sim \mathcal U([-k,k])$ and $\|x\|,\|x'\| < k$ then
\begin{equation}
\begin{aligned}
&2k Vol_{d-1}(\mathbb S^{d-1}) \cdot  \mathbb E_{(a,b)}[ax+b]_+ [ax'+b]_+ &&=&& \eta_{d,0}\frac{1}{(d-2)!!} \frac{2\pi}{3} k^{3} + \eta_{d,2}\frac{2}{d!!} k \|x\|\|x'\|\pi \cos(\alpha)\\
&&&&&+\eta_{d,3}\frac{3}{(d+1)!!} \left( \left[E_1\right]_{\tau}^{\tau+\pi}  + \left[E_2\right]_{\tau-\pi}^{\tau}\right)\\[.2cm]
&2k Vol_{d-1}(\mathbb S^{d-1}) \cdot  \mathbb E_{(a,b)}[ax+b]_+ {\bf 1}[ax'+b]a &&=&& \eta_{d,1} \frac{2}{d!!} Q^T \begin{bmatrix}
0_{d-2}\\
k \pi \|x\|\\
0
\end{bmatrix}
+\eta_{d,2} \frac{3}{(d+1)!!} Q^T
\begin{bmatrix}
0_{d-2}\\
G_1\\
G_2\\
\end{bmatrix}\\[.2cm]
&2k Vol_{d-1}(\mathbb S^{d-1}) \cdot  \mathbb E_{(a,b)}{\bf 1}[ax+b] {\bf 1}[ax'+b]aa^T &&=&& \eta_{d,0} \frac{2\pi k}{d!!} Id_{d} + \eta_{d,1} \frac{3}{(d+1)!!} Q^T \begin{bmatrix}
\frac{1}{3} \delta Id_{d-2} & 0 & 0\\
0 & \alpha & \beta\\
0 & \beta & \gamma\\
\end{bmatrix}
Q
\end{aligned}
\end{equation}
where $\alpha = \arccos\left(\frac{x\cdot x'}{\|x\|\|x'\|}\right)$, $\tau = \arctan \bigg( \frac{||x|| - ||x'|| \cos(\alpha)}{||x'|| \sin(\alpha)} \bigg)$, $Q \in SO(d)$ is such that $Qx = (0,\ldots,\|x\|,0)^T$,$Qx' = (0,\ldots,\|x'\|\cos(\alpha),\|x'\|\sin(\alpha))^T$ and
\[
\eta_{d,r} = \begin{cases} 2^{\lceil\frac{d-2}{2}\rceil}\pi^{\lfloor\frac{d-2}{2}\rfloor} &\mbox{if } r \mbox{ is even,}\\
2^{\lfloor\frac{d-2}{2}\rfloor}\pi^{\lceil\frac{d-2}{2}\rceil} &\mbox{if } r \mbox{ is odd.}\\
\end{cases}
\]
\begin{small}
\begin{equation}
\begin{aligned}
&E_1 = &&\frac{1}{18} \, {\left(\sin\left(\psi\right)^{3} - 3 \, \sin\left(\psi\right)\right)} \|x\|^{3} + \frac{1}{24} \, \|x\|^{2} \|x'\| {\left(3 \, \sin\left(\psi +\alpha \right) + \sin\left(3 \, \psi - \alpha \right) + 6 \, \sin\left(\psi - \alpha\right)\right)} \\
&E_2 = &&\frac{1}{18} \, {\left(\sin\left(\psi - \alpha\right)^{3} - 3 \, \sin\left(\psi - \alpha\right)\right)}\|x'\|^{3} + \frac{1}{24} \, \|x\|\|x'\|^{2} {\left(\sin\left(3\psi - 2\alpha \right) + 3 \, \sin\left(\psi - 2\alpha\right) + 6 \, \sin\left(\psi \right)\right)}\\
&G_1 = &&\frac{1}{3} \, \|x\|^{2} \sin\left(\tau\right)^{3} - \|x\|^{2} \sin\left(\tau\right) + \frac{1}{2} \, \|x\| \|x'\| \sin\left(\alpha + \tau\right) + \frac{1}{6} \, \|x\| \|x'\| \sin\left(-\alpha + 3 \, \tau\right) \\
    &&&+ \|x\| \|x'\| \sin\left(-\alpha + \tau\right) - \frac{1}{12} \, \|x'\|^{2} \sin\left(-2 \, \alpha + 3 \, \tau\right) - \frac{1}{4} \, \|x'\|^{2} \sin\left(-2 \, \alpha + \tau\right) - \frac{1}{2} \, \|x'\|^{2} \sin\left(\tau\right)\\
&G_2 = &&\frac{1}{3} \, \|x\|^{2} \cos\left(\tau\right)^{3} -\frac{1}{2} \, \|x\| \|x'\| \cos\left(\alpha + \tau\right) - \frac{1}{6} \, \|x\| \|x'\| \cos\left(-\alpha + 3 \, \tau\right) + \frac{1}{12} \, \|x'\|^{2} \cos\left(-2 \, \alpha + 3 \, \tau\right) \\
    &&&  - \frac{1}{4} \,  \|x'\|^{2} \cos\left(-2 \, \alpha + \tau\right) + \frac{1}{2} \,  \|x'\|^{2} \cos\left(\tau\right)\\
&\alpha =&& \left[\|x\| \sin\left(\psi\right) -\frac{1}{3} \, \|x\| \sin\left(\psi\right)^{3}\right]_{\tau}^{\tau+\pi} + \left[\frac{1}{4} \, \|x'\| \sin\left(\alpha + \psi\right) + \frac{1}{12} \,\|x'\| \sin\left(3\psi -\alpha\right) + \frac{1}{2} \, \|x'\| \sin\left(\psi -\alpha \right)\right]_{\tau-\pi}^\tau\\
&\beta = &&\left[-\frac{1}{3} \, \|x\| \cos\left(\psi\right)^{3}\right]_{\tau}^{\tau+\pi} + \left[-\frac{1}{4} \, \|x'\| \cos\left(\alpha + \psi\right) - \frac{1}{12} \, \|x'\| \cos\left(3\psi -\alpha \right)\right]_{\tau-\pi}^\tau\\
&\gamma =&& \left[\frac{1}{3} \, \|x\| \sin\left(\psi\right)^{3}\right]_{\tau}^{\tau+\pi} + \left[-\frac{1}{4} \, \|x'\| \sin\left(\alpha + \psi\right) - \frac{1}{12} \, \|x'\| \sin\left(3 \psi - \alpha \right) + \frac{1}{2} \, \|x'\| \sin\left( \psi - \alpha \right)\right]_{\tau-\pi}^\tau\\
&\delta = &&\left[\|x\|\sin(\psi)\right]_{\tau}^{\tau+\pi} + \left[\|x'\| \sin(\psi - \alpha)\right]_{\tau-\pi}^\tau\\
\end{aligned}
\end{equation}
\end{small}
\end{proposition}

\begin{proof} The idea is to compute the integral with respect to the bias term $b$ and then split the result into homogeneous expressions where Lemma~\ref{lemma:homogeneous_integrals} can be applied. In particular, assuming that $-k \le s,t \le k$:
\[
\begin{aligned}
\mathcal I&(s,t) = \int_{-k}^k [s+b]_+ [t+b]_+ db = \int_{\min(s,t)}^k (s+b)(t+b) db\\
&=  \underbrace{\left(\frac{1}{3} \, k^{3} \right)}_{\mathcal I^0} + \underbrace{\left(\frac{1}{2} \, k^{2} {\left(s + t\right)}\right)}_{\mathcal I^1} + \underbrace{\bigg(k s t\bigg)}_{\mathcal I^2} + \underbrace{\left(\frac{1}{3} \, \min(s,t)^{3} - \frac{1}{2} \, \min(s,t)^{2}{\left(s +
t\right)} + \min(s,t) s t\right)}_{\mathcal I^3},
\end{aligned}
\]
where we collected terms that have total degree $0, 1, 2, 3$ in $s,t$.
The computation of $\int_{b \in [-k,k]} \int_{a \in \mathbb S^{d-1}} [a x+b]_+ [ax'+b] db \, d\Omega = \int_{a \in \mathbb S^{d-1}}  I(a\cdot x, a \cdot x') d \Omega$ is now reduced to four one-dimensional integrals of the form $\int_{0}^{2\pi} \mathcal I^j(\|x\| \cos(\psi), \|x'\|\cos(\psi-\alpha)) d \psi$ where $\alpha = \arccos\left(\frac{x \cdot x'}{\|x\|\|x'\|}\right)$. The most tedious case is $j=3$ where we need to compare $\|x\|\cos(\psi)$ and $\|x'\|\cos(\psi+\alpha)$. Assuming $0 \le \alpha \le \pi$, then
$\|x\|\cos(\psi) < \|x'\| \cos(\alpha-\psi)$ holds if and only if $\tau \le
\psi \le \tau + \pi$ where 
\[
\tau = \arctan\left (\frac{\|x\| - \|x'\|\cos \alpha}{\|x'\|\sin(\alpha)}\right).
\]
From this we obtain that 
\[
\begin{aligned}
&\int_0^{2\pi} \mathcal I^3(\|x\|\cos(\psi), \|x'\| \cos(\alpha-\psi) d \psi\\
&= \int_{\tau-\pi}^\tau \left(-\frac 1 6 \|x'\|^3 \cos(\alpha-\psi)^{3} + \frac{1}{2} \, \|x'\|^2 \|x\| \cos(\alpha-\psi)^{2}\cos(\psi)\right) d \psi\\
&+ \int_{\tau}^{\tau+\pi} \left(-\frac 1 6 \|x\|^3 \cos(\psi)^{3} + \frac{1}{2} \, \|x\|^2 \|x'\| \cos(\psi)^{2}\cos(\alpha - \psi)\right) d \psi
\end{aligned}
\]
Expanding this integral we obtain the expressions for $\mathbb E_{(a,b)}[ax+b]_+[ax'+b]_+$ in the statement. The remaining integrals are computed similarly by considering the homogeneous parts of
\[
\begin{aligned}
\mathcal I_t&(s,t) = \int_{-k}^k [s+b]_+ {\bf 1}[s+b] db = \int_{\min(s,t)}^k (s+b) db\\
&=  \frac{1}{2} \, k^{2} + k s + \min(s,t) s - \frac{1}{2}, \, \min(s,t)^{2}\\[.3cm]
\mathcal I_{s,t}&(s,t) = \int_{-k}^k {\bf 1}[s+b] {\bf 1}[s+b] db = \int_{\min(s,t)}^k 1 db
=  k - \min(s,t).
\end{aligned}
\]
\end{proof}

\subsection{Gaussian Initialization}
The Gaussian initialization~\eqref{eq:gaussian_initialization} yields the following simpler formula for $K_\infty$. The first term is well known and is derived in~\cite{cho2009kernel}. The second and third terms are easily derived by taking derivatives of the first term.

\begin{proposition} If $a \sim \mathcal N(0,Id_{d-1})$ and $b \sim \mathcal N(0,1)$, then
\begin{equation}
\begin{aligned}
& 2\pi \cdot \mathbb E_{(a,b)}[ax + b]_+[ax'+b]_+ &&=&&  \|\tilde x\|\|\tilde x'\|(\sin(\tilde \alpha) + (\pi-\tilde \alpha)\cos(\tilde \alpha))\\[.2cm]
& 2\pi \cdot \mathbb E_{(a,b)}[ax + b]_+{\bf 1}[ax'+b]a &&=&& \|\tilde x\|(\sin(\tilde \alpha) + (\pi-\tilde \alpha)\cos(\tilde \alpha)) \frac{x'}{\|\tilde x'\|} + (\pi -\tilde \alpha)\left(Id_{d} - \frac{x'x'^T}{\|\tilde x'\|^2}\right) x\\
& 2\pi \cdot \mathbb E_{(a,b)}{\bf 1}[ax + b]{\bf 1}[ax'+b]aa^T &&=&& (\pi-\tilde \alpha)Id +
\sin(\tilde \alpha)\frac{x'x^T
}{\|\tilde x'\|\|\tilde x\|}\\
&&&&&+ \frac{1}{\sin(\tilde \alpha)}
 \left(Id_d - \frac{x' x'^T}{\|\tilde x'\|^2}\right)\frac{x x'^T}{\|\tilde x\|\|\tilde x'\|}\left(Id_d - \frac{xx^T}{\|\tilde x\|^2}\right),
\end{aligned}
\end{equation}
where $\tilde x = (x,1)$, $\tilde x' = (x',1)$, and $\tilde \alpha = \arccos\left(\frac{\tilde x \cdot \tilde x'}{\|\tilde x\|\|\tilde x'\|}\right)$.
\end{proposition}

\section{Poisson Surface Reconstruction Kernel}\label{sec:poisson_kernel_details}
In its simplest form, Poisson reconstruction of a surface
\cite{kazhdan2013screened}, extracts the level set of a smoothed indicator function determined as the solution of 
\begin{equation*}
-\Delta f  = \nabla \cdot V,
\end{equation*}
where $V$ is a vector field obtained from normals $n_i$ at samples $x_i$, and we use $f$ to denote the (smoothed) indicator function as it plays the same role as  $f$ in \eqref{eq:objective}. 
The equation above is closely related to \eqref{eq:objective}:  specifically, it is the equation for the minimizer of $\int_{\mathbb{R}^3}\|\nabla_x f(x)-V\|^2dx$, i.e., the second term in \eqref{eq:objective}, can be viewed as a approximation of this term by sampling at $x_i$. 
The screened form of Poisson reconstruction effectively adds the first term with $y_i = 0$, as the indicator function at points of interest is supposed to be zero. 
For the Poisson equation, the solution can be explicitly written as an integral 
\begin{equation*}
f(x) = \int_{\mathbb{R}^3} \frac{\nabla_z \cdot V(z) dz}{|x-z|}
\end{equation*}
The vector field $V$ is obtained by interpolating the normals using a fixed-grid spline basis and barycentric coordinates of the sample points with respect to the grid cell containing it. 
This is equivalent to using a non-translation invariant non-symmetric locally-supported kernel $K_B(z,x)$: 
\begin{equation*}
V(z) = \sum_i K_B(z,x_i) n_i.
\end{equation*}
Let $B_{1,3}(x-c_j)$, $x\in \RR^3$ be the trilinear basis
$B_1(x^1-c_j^1)B_1(x^2-c_j^2)B_1(x^3-c_j^3)$
function centered at a regular grid point $c_j$, and  $B_{n,3}(x-c_j)$ be a tensor-product spline basis function of degree $n$ defined in a similar way. \emph{(Note that in Lemma~\ref{lemma:poisson_kernel} in the main document, we slightly abuse notation, denoting $B_1$ as the trilinear basis and $B_n$ as the degree-$n$ spline basis)}. Poisson reconstruction uses $n=1$ or $n=2$ where 
\begin{equation}
    B_2(x) = \begin{cases} 
        0                                      & \text{if } x < -1.5\\
        \frac{1}{2} x^2                        & \text{if } -1.5 \leq x < -0.5\\
        -\frac{1}{2} + x - (x - 1)^2           & \text{if } -0.5 \leq x < 0.5\\
        \frac{5}{2} - x + \frac{1}{2}(x - 2)^2 & \text{if } 0.5 \leq x < 1.5\\
        0                                      & \text{if } 1.5 \leq z
    \end{cases}
\end{equation}
Then 
$ K_B(z,x) = \sum_j B_{1,3}(x-c_j) B_{n,3}(z-c_j)$. 
where  only 8 terms corresponding to the vertices $c_j$ of the grid cube containing $x$ are nonzero. This yields the following expression for the kernel corresponding to Poisson reconstruction, 
\begin{equation*}
K_{\text{PR}}(x,x')_g =  \int_{\mathbb{R}^3} \frac{\nabla_z K_B(z,x')  dz}{|x-z|}
\end{equation*}
\ie, the convolution of the Laplacian kernel $1/|x-z|$ and the gradient of $K_B$.  Using the identity $\nabla (f * g) = (\nabla f * g)$, we can write this as the gradient of $K_{\text{PR}}(x,x')_g$, defined as
\begin{equation}
K_{\text{PR}}(x,x') =  \int_{\mathbb{R}^3} \frac{ K_B(z,x')  dz}{|x-z|}
\end{equation}
To make it easier to understand the qualitative behavior of the kernel, replacing $K_B(z,x)$ with a radial kernel $B^1_n(|z-x|)$, with qualitatively similar behavior (see Figure~\ref{fig:approx_poisson})
yields a translation-invariant radial approximation $K_{\text{PR}}^{\text{approx}}$ of the kernel $K_{\text{PR}}$, as the convolution of two radial kernels is a radial function.  

\begin{equation}
K^{\text{approx}}(x,x') = \int_{\R^3} \frac{ B^1_n(|z-x'|)  dz}{|x-z|}
\end{equation}
As both $B^1_n$ and the Laplace kernel are radial functions, their convolution is also radial. 
It can be expressed in a more explicit form using the relation between Fourier and Hankel transforms for radial functions.  For $n=3$, the Hankel transform is related to Fourier transform by
\cite{poularikas2018transforms}
\[s^{1/2}\F[g](s) = (2\pi)^{3/2} \Hk[g](s)
\]
The Hankel transform is an involution, so the relationship for the inverse Fourier transform is similar. 
Writing $g*h = \F^{-1}[\F[g]\F[h]]$, we  obtain 
the expression for the radial convolution in terms of
one-dimensional integrals,

\begin{equation*}
K^{\text{approx}}(x) =  \Hk_s[s^{-3/2}\Hk_r[B^1_n]](|x|),
\end{equation*}
where we use $\Hk_r[1/r] = 1/s$.
and $K^{\text{approx}}_g(x)$ is just the gradient of this, i.e., a derivative times $|x|/x$.

\emph{The RKHS norm} for the space corresponding to this kernel is given by 
\begin{equation*}
\|f\|_{\mathcal H} = \int \frac{|\F[f]|^2}{\F[K^{\text{approx}}]}d\omega
\end{equation*}
with $\F[K^{\text{approx}}]$ obtained 
using the Hankel transforms as above. 

\section{RKHS Norm of the Neural Spline Kernel}\label{sec:rkhs_norm}
We now discuss how $c(a, b)$ in \eqref{eq:RKHS} is related to the Laplacian of the fucntion. If we make the mild assumption that our functions contain a linear and bias term (Lemma~\ref{lemma:even_measures}), then $c(a, b)$ is the Radon Transform of the laplacian of the function. Thus, the least norm minimizers of the least squares problem \eqref{eq:objective} are related to the laplacian of the function and the RKHS norm corresponds to the integral of the laplacian over hyperplanes in the domain. In our experiments, we added an option to include the linear and bias terms to the solution. They appear to have no effect on the final reconstruction. The derivation below is borrowed from \cite{ongie2019function}.

\begin{lemma}\label{lemma:even_measures}
Let $f_\text{lim}(x): \RR^d \rightarrow \RR$ be an infinite-width, one hidden layer neural whose weights $a, b$ are distributed according to the measure $c(a, b) : \mathbb{S}^{d-1} \times \mathbb{R} \rightarrow \mathbb{R}$
\begin{equation}\label{eq:limfun}
   f_\text{lim}(x; c) = \int_{\mathbb{S}^{d-1} \times \mathbb{R}} [a^Tx - b]_+ c(a, b) da db + v^Tx + d~.
\end{equation}
Then, $f_\text{lim}(x)$ can always be rewritten as 
\begin{equation}
    \int_{\SpS^{d-1} \times \RR} [a^Tx - b]_+ c^+(a, b)dadb + v'^Tx + d'
\end{equation}
where $c^+(a, b)$ is an even measure on $\mathbb{S}^{d-1} \times \mathbb{R}$, $v \in \RR^d$, and $d \in \RR$.
\end{lemma}
\begin{proof}
We can split the integral in $f_\text{lim}$ into even and odd parts:
\begin{align*}
    f_\text{lim}(x) &= \frac{1}{2}\int_{\SpS^{d-1} \times \RR} ([a^Tx - b]_+ + [a^Tx - b]_+) c^+(a, b)dadb\\
                    &+ \frac{1}{2}\int_{\SpS^{d-1} \times \RR} ([a^Tx - b]_+ - [a^Tx - b]_+) c^-(a, b)dadb\\
                    &+ v^Tx + d
\end{align*}
where $c^+$ and $c^-$ are the even and odd parts of $c$ respectively. Observing that $[t]_+ + [-t]_+ = |t|$ and $[t]_+ - [-t]_+ = t$, we have that
\begin{align}
        f_\text{lim}(x) &= \frac{1}{2}\int_{\SpS^{d-1} \times \RR} \big(|a^Tx - b| c^+(a, b) + (a^Tx - b) c^-(a, b)\big)dadb + v^Tx + d\\
                        &= \frac{1}{2}\int_{\SpS^{d-1} \times \RR} |a^Tx - b| c^+(a, b)dadb + v'^Tx + d'\\
                        &= \int_{\SpS^{d-1} \times \RR} [a^Tx - b]_+ c^+(a, b)dadb + v'^Tx + d'
\end{align} where $v' = v + \int_{\SpS^{d-1} \times \RR} a c^-(a, b) dadb$, $d' = d + \int_{\SpS^{d-1} \times \RR} b c^-(a, b) dadb$, and the last step holds because $c^+$ is even.
\end{proof}

Using Lemma~\ref{lemma:even_measures}, we will consider without loss of generality, neural networks of the form \eqref{eq:limfun} with even measures $c(a, b)$.
We now give a few useful definitions and lemmas.

\begin{definition} Let $f : \RR^d \rightarrow \RR$. The \emph{Radon Transform} of $f$ is
\begin{equation}
    \Radon{f}(a, b) \coloneqq \int_{a^Tx = b} f(x) ds(x)
\end{equation} 
where $ds(x)$ is a measure on the $(d-1)$-hyperplane $a^Tx = b$. Intuitively the Radon transform represents a function in terms of its integrals along all possible hyperplanes.
\end{definition}
\begin{remark}
Since the hyperplane $a^Tx = b$ is the same as the hyperplane $-a^Tx = -b$, the Radon transform is an \emph{even} function. i.e. $\mathcal{R}\{f\}(a, b) = \mathcal{R}\{f\}(-a, -b)$.
\end{remark}

\begin{definition} Let $\varphi : \SpS^{d-1} \times \RR \rightarrow \RR$. The \emph{Dual Radon Transform} of $\varphi$ is the adjoint of the Radon Transform $\mathcal{R}$
\begin{equation}
    \RadonT{\varphi}(x) \coloneqq \int_{\SpS^{d-1}} \varphi(a, x^Ta) d\Omega
\end{equation} 
where $d\Omega$ is a measure on the $(d-1)$-hypersphere $\SpS^{d-1}$. Intuitively the Dual Radon transform represents a function at $x$ in terms of its integrals on all hyperplanes through $x$.
\end{definition}

The Radon Transform satisfies the \emph{intertwining property}. i.e. for any positive integer $s$
\begin{equation}
 \Radon{(-\Delta)^{\frac{s}{2}} f} = (-\partial_b^2)^{\frac{s}{2}} \Radon{f}
\end{equation}

\begin{lemma} \label{lemma:inverse_dual_radon} (From \cite{solmon1987asymptotic}) If $\varphi(a, b) = \varphi(-a, -b)$ is an even function mapping $\SpS^{d-1} \times \RR$ to $\RR$ which is $\mathcal{C}^\infty$ smooth and whose partial derivatives decrease at a rate faster than $\mathcal{O}(|b|^{-N})$ as $|b| \rightarrow \infty$ for any $N \geq 0$, then the Dual Radon Transform can be inverted using
\begin{equation}
    \frac{1}{2(2\pi)^{d-1}} \Radon{(-\Delta)^{\frac{d-1}{2}} \RadonT{\varphi}} = \varphi
\end{equation}
\end{lemma}

\begin{lemma} (From \cite{ongie2019function})
Let $f_\text{lim}(x): \RR^d \rightarrow \RR$ be an infinite-width, one hidden layer neural whose weights $a, b$ are distributed according to the even measure $c(a, b)$. Then $c(a, b)$ can be expressed as
\begin{equation*}
    c(a, b) = \gamma_d \mathcal{R} \{ (-\Delta)^\frac{d+1}{2} f_\text{lim}(x) \},
\end{equation*} 

where $\mathcal{R}\{f\}(a, b)$ is the Radon Transform of $f$. In particular, for $d=3$, 
\begin{equation*}
    \gamma_d \mathcal{R} \{ \Delta f_\text{lim}(x) \} = c(a, b)
\end{equation*}
\end{lemma}

\begin{proof}
The Laplacian of $f_\text{lim}$ in is \eqref{eq:limfun}
\begin{equation}
    \Delta f_\text{lim}(x; c) = \int_{\SpS^{d-1} \times \RR} \delta(a^Tx = b) c(a, b) da db = \int_{\SpS^{d-1}} c(a, a^Tx) d\Omega
\end{equation}
which is precisely the Dual Radon Transform of $c(a, b)$. Since $c$ is even, and assuming it decays rapidly with $b$, we can invert it using Lemma~\ref{lemma:inverse_dual_radon} yielding
\begin{equation*}
    c(a, b) = \gamma_d \mathcal{R} \{ (-\Delta)^\frac{d+1}{2} f_\text{lim}(x) \}.
\end{equation*}
\end{proof}
\begin{corollary}
The RKHS norm of the function $f_\text{lim}$ is
\begin{equation}
    ||f_\text{lim}||_{\mathcal H} = ||c(a, b)||_2 + ||v||_2 + |d| = \bigg(\int_{\SpS^{d-1} \times [-k, k]} c(a, b)^2 d\Omega db\bigg)^{\frac{1}{2}} + ||v||_2 + |d|
\end{equation}
\end{corollary}

\end{document}